\documentclass[9pt,twocolumn,twoside,lineno]{pnas-new}

\templatetype{pnasinvited} 

\usepackage{amssymb}            
\usepackage{mathtools}          
\usepackage{mathrsfs}           
\mathtoolsset{showonlyrefs}     
\usepackage{graphicx}           
\usepackage{subcaption}         
\usepackage[space]{grffile}     
\usepackage{url}                
\usepackage{pdflscape}
\usepackage{longtable} 

\usepackage{leftidx}
\usepackage{microtype}
\usepackage{graphicx}
\usepackage{booktabs} 
\usepackage[scr=boondox]{mathalfa}
\usepackage{array}
\usepackage{wrapfig}
\usepackage{mathrsfs}
\usepackage{tabu}
\usepackage{mathtools}
\usepackage{csquotes}
\usepackage{amsmath}
\usepackage{amsthm}
\usepackage{bm}
\usepackage{amssymb}
\usepackage{cancel}
\usepackage{bbm,dsfont}
\usepackage{soul}
\usepackage{enumitem}
\newtheorem{theorem}{Theorem}[section]
\newtheorem{corollary}{Corollary}[section]

\newtheorem{remark}{Remark}

\definecolor{darkred}{rgb}{0.55, 0.0, 0.0}
\definecolor{tradrb}{rgb}{0.0, 0.14, 0.4}
\definecolor{royalazure}{rgb}{0.0, 0.20, 0.57}
\definecolor{crimsonglory}{rgb}{0.75, 0.0, 0.2}
\definecolor{darkcyan}{rgb}{0.0, 0.55, 0.55}
\definecolor{darkcerulean}{rgb}{0.03, 0.27, 0.49}
\newcommand\crule[3][blue]{\textcolor{#1}{\rule{#2}{#3}}}

\usepackage{multicol, blindtext}
\usepackage[algo2e,linesnumbered,ruled,lined]{algorithm2e}

\usepackage[utf8]{inputenc}     
\usepackage[english]{babel}    
\usepackage{tikz}
\usetikzlibrary{cd}
\usepackage{graphicx}
\newcolumntype{L}{>{$}l<{$}}
\newcommand{\dg}{\textsl g}
\newcommand{\xp}{x_f}
\newcommand{\xm}{x_f}
\newcommand{\tp}{t_f}
\newcommand{\tm}{t_f}
\newcommand{\mc}{\mathcal}

\newcommand{\mpol}{\mu}
\newcommand{\pfun}{\rho}
\newcommand{\fff}{F}
\newcommand{\bz}{\mathbf{z}}
\newcommand{\bos}{\mathbf{s}}

\newcommand{\bod}{\mathbf{d}}

\newcommand{\ba}{\boldsymbol{\alpha}}
\newcommand{\boa}{\mathbf{a}}

\newcommand{\bs}{\boldsymbol{\sigma}}

\newcommand{\bpsi}{\boldsymbol{\psi}}
\DeclareMathOperator*{\argmax}{argmax}
\DeclareMathOperator*{\argmin}{argmin}
\DeclareMathOperator*{\expec}{\mathop{\mathbb{E}}}
\newcommand\mydots{\hbox to 1em{.\hss.\hss.\hss}}
\allowdisplaybreaks

\DeclareMathSymbol{\shortminus}{\mathbin}{AMSa}{"39}
\newcommand{\medminus}{\scalebox{0.6}[0.7]{\(-\)}}
\newcommand{\minus}{\mathchoice{-}{-}{\medminus}{\shortminus}}

\makeatletter
\newcommand{\pushright}[1]{\ifmeasuring@#1\else\omit\hfill$\displaystyle#1$\fi\ignorespaces}
\newcommand{\pushleft}[1]{\ifmeasuring@#1\else\omit$\displaystyle#1$\hfill\fi\ignorespaces}
\makeatother

\newtheorem{definition}{Definition}[section]

\begin{document}




\title{A Unified Theory of Compositionality, Modularity, and Interpretability in Markov Decision Processes}


\author[1]{Thomas J. Ringstrom}
\author[a]{Paul R. Schrater}

\affil[a]{University of Minnesota}

\leadauthor{Ringstrom}

\authorcontributions{Please provide details of author contributions here.}
\correspondingauthor{\textsuperscript{1}To whom correspondence should be addressed. E-mail: rings034@gmail.com}


\begin{abstract}
We introduce \textit{Option Kernel Bellman Equations} (OKBEs) for a new reward-free Markov Decision Process. Rather than a value function, OKBEs directly construct and optimize a predictive map called a \textit{state-time option kernel} (STOK) to maximize the probability of completing a goal while avoiding constraint violations. STOKs are compositional, modular, and interpretable initiation-to-termination transition kernels for policies in the Options Framework of Reinforcement Learning. This means: 1) STOKs can be composed using Chapman-Kolmogorov equations to make spatiotemporal predictions for multiple policies over long horizons, 2) high-dimensional STOKs can be represented and computed efficiently in a factorized and reconfigurable form, and 3) STOKs record the probabilities of semantically interpretable goal-success and constraint-violation events, needed for \textit{formal verification}. Given a high-dimensional state-transition model for an intractable planning problem, we can decompose it with local STOKs and goal-conditioned policies that are aggregated into a factorized \textit{goal kernel}, making it possible to forward-plan at the level of goals in high-dimensions to solve the problem. These properties lead to highly flexible agents that can rapidly synthesize meta-policies, reuse planning representations across many tasks, and justify goals using \textit{empowerment}, an intrinsic motivation function. We argue that reward-maximization is in conflict with the properties of compositionality, modularity, and interpretability. Alternatively, OKBEs facilitate these properties to support verifiable long-horizon planning and intrinsic motivation that scales to dynamic high-dimensional world-models.
\end{abstract}


\maketitle
\thispagestyle{firststyle}
\ifthenelse{\boolean{shortarticle}}{\ifthenelse{\boolean{singlecolumn}}{\abscontentformatted}{\abscontent}}{}

Replicating intelligent behavior is a central goal of artificial intelligence; as humans, our cognitive abilities have allowed us to thrive, as we have drawn strength from our remarkable flexibility in adapting, planning, and problem solving at multiple levels of abstraction. The world as understood by an agent is dynamic, re-configurable, and high-dimensional, consisting of many coupled systems. An agent may need to drink water from a lake to regulate hydration levels to avoid dying \cite{keramati2014homeostatic, laurenccon2021continuous, laurencon2024continuous}, solve complex history-dependent tasks \cite{gaon2020reinforcement, icarte2018using}, create and use abstractions \cite{abel2016near, konidaris2019necessity, ho2019value}, modify the environment, or repeat action sequences to complete a task \cite{pickett2002policyblocks, topin2015portable}. The disciplines of artificial intelligence and computational neuroscience are both confronted with a core outstanding question: what principles underpin flexible and interpretable goal-directed action in high-dimensional worlds \cite{lake2017building}? Let's consider three.

In artificial intelligence, \textbf{compositionality} is the ability to compose primitives into composite structures, crucial for developing algorithms with reduced sample complexity \cite{du2021unsupervised, du2024compositional}. 
Furthermore, \textbf{modularity} enables system decomposition and representation remapping. Different architectures could allow agents to plan with the flexibly of animals \cite{lecun2022path, tsividis2021human}, but they may need to be modular. Deep-reinforcement learning (DRL, RL) is often used in an attempt to \textit{discover} architectures that generalize structure across tasks \cite{team2021open}. Lastly, safety researchers advocate for scaling verification methods to entire world-models to ensure deployed systems satisfy task constraints \cite{dalrymple2024towards, leike2017ai, seshia2022toward}. This requires \textbf{interpretability}: planning representations should \textit{say what events an agent might cause}. 

In computational neuroscience, researchers have emphasized the importance of reasoning with goals \cite{molinaro2023goal} and the sufficiency of reward as a general theory of motivation has been called into question \cite{juechems2019does}. Interest in predictive representations like the successor representation (SR) \cite{dayan1993improving, momennejad2017successor, stachenfeld2017hippocampus, gershman2018successor, brunec2022predictive, carvalho2024predictive}, the concept of default dynamics \cite{piray2021linear, piray2024reconciling}, and the Options Framework in RL \cite{sutton1999between, xia2021temporal} have been motivated out of a desire to identify principles of representation reuse, compositionality, and temporal abstractions for planning; and, compositionality itself is of great interest to cognitive scientists \cite{reverberi2012compositionality, lake2015human, frankland2020concepts, driscoll2022flexible, davidson2022creativity, zhou2024compositional}. Researchers have also begun to identify the mechanisms of goal and task-structure remapping, demonstrating the interaction between cortical task representations and hippocampal planning representations \cite{mark2023flexible, el2023cellular, samborska2022complementary}, as well as the generation of compositional sequences in the entorhinal cortex \cite{kurth2023replay}. Others have highlighted benefits of sequence chunking for fast compositional planning \cite{eltetHo2023habits}, and state-space composition \cite{sharma2021map} could allow an animal to consider yet-to-be experienced states of the world \cite{bakermans2025constructing}. Experimentalists have also discovered `lap cells' which record, as a transferable abstract state, the number of cycles around a hallway loop an animal has completed to get food (a problem we model in Fig. \ref{fig:3level}) \cite{sun2020hippocampal}. All of these findings point to a need to derive general mathematical principles for rapidly reasoning about goal-feasibility across hierarchical spaces and the flexible reuse of task structure and planning representations.

We argue that a theory of compositional and modular planning \textit{architecture}, and a theory interpretable and verifiable planning \textit{algorithms}, are problems with the same solution, but it is not found within the standard reward-maximization Bellman-foundations common to computational neuroscience and artificial intelligence. Instead we argue that these properties, sought after in DRL, require the formalization of new Bellman equations for \textit{program synthesis} in Markov Decision Processes (MDPs) \cite{bellman1957markovian, puterman2014markov, manna1971toward, bastani2018verifiable, inala2020synthesizing, trivedi2021learning, qiu2022programmatic, silver2020few, cui2024reward}. Computer scientists will appreciate we have created new Bellman equations for high-dimensional compositional and verifiable planning as program synthesis for problems traditionally formalized with sparse-rewards.
Neuroscientists will appreciate that we have combined the themes of 1) compositional state-spaces and predictive representations, 2) goal-conditioned options, 3) representation reuse, 4) hierarchical  planning, and 5) compositional sequence optimization into one unified framework, with new ideas for studying animal intelligence. 

\begin{figure}[h]
    \centering \includegraphics[width=\linewidth]{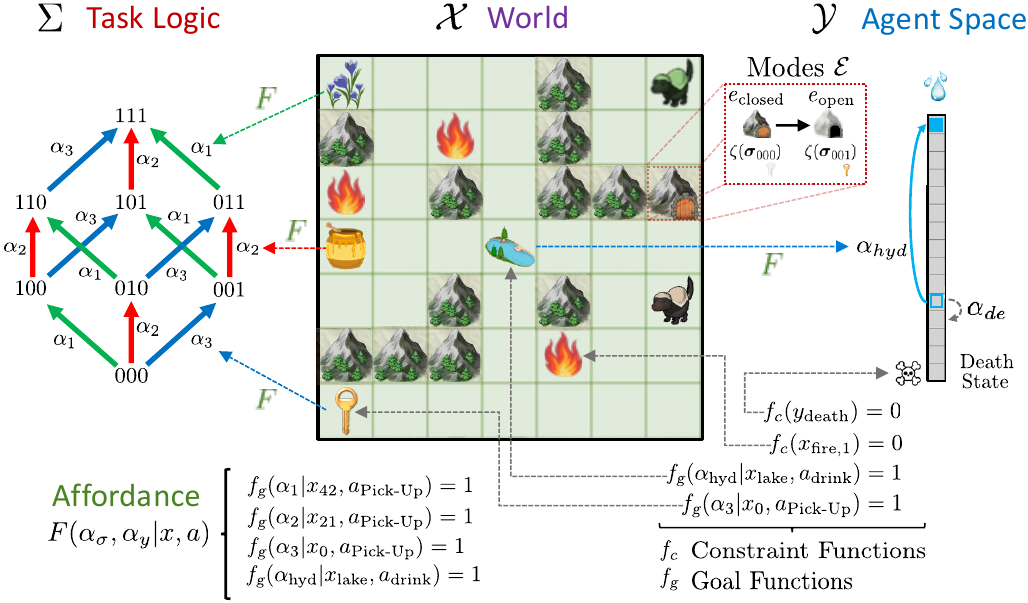}
    \caption{The agent must bring honey and flowers to a friend while avoiding death by regulating internal states. A key also must be obtained to unlock the mountain door. Sub-goals are encoded in $f_{\dg}$ and derived from $F$, and constraints are encoded in $f_c$.
    }
    \label{fig:goal-constraint}
\end{figure}

\subsection{A motivating example}

Consider a honey badger agent in Fig. \ref{fig:goal-constraint}, minimally constituted as a set of coupled transition kernels: a base-level (BL) grid-world space $P_x(x'|x,a,e)$, a hydration space $P_y(y'|y,\alpha_y)$ with high-level (HL) ``internal-action" variables $\alpha_y \in \mc A_{y}=\{\alpha_{\text{hyd}},\alpha_{\text{de}}\}$ which hydrates or dehydrate the agent, respectively; and a logical space $P_{\bs}(\bs'|\bs, \alpha_{\bs})$, where $\bs$ is a binary vector, and $\alpha_{\bs}\in \mc A_{\sigma}=\{\alpha_0,\alpha_1,\alpha_2,\alpha_3\}$ is a variable that flips bits ($\alpha_{2}$ flips bit $2$, $(1,0,0)\xrightarrow{\alpha_{2}} (1,1,0)$, $\alpha_0$ flips no bits). The full Cartesian product-space dynamics has a factorization:
\begin{align}
    &\quad\quad P_{\bos}(\bs',y',x'|\bs,y,x,a)=\label{eq:joint_kernel}\\
    &\sum_{\alpha_{\bs},\alpha_y}P_{\bs}(\bs'|\bs, \alpha_{\bs})P_y(y'|y,\alpha_y)F(\alpha_{y},\alpha_{\bs}|x,a)P_x(x'|x,a,\zeta(\bs)).\\[-1cm]
\end{align}
We couple the transition kernels with an \textit{affordance function} $F$:
\begin{definition}[Affordance Function] \label{def:afford}
    An affordance function $F:(\mc X \times \mc A) \times \mc A_{\bz} \rightarrow [0,1]$ is a factorized conditional joint distribution, $F(\ba_{\bz}|x,a)=F(\alpha_{z_1},...,\alpha_{z_n}|x,a)=\prod_kF_k(\alpha_{z_k}|x,a),$ on HL action vectors $\mc A_{\bz} \subseteq \mc A_{z_1} \times...\times \mc A_{z_n}$, where $\ba_{\bz} = (\alpha_{z_1},...,\alpha_{z_n})\in \mc A_{\bz}$, and $\mc A_{z_1}, ..., \mc A_{z_n}$ are action sets for state-spaces $\mc Z_1,...,\mc Z_n$. 
\end{definition}
This function communicates the probability that a state-action of one transition system induces transformations on other transition systems, e.g. an agent can drive the $y$-dynamics by taking action $a_{\text{drink}}$ at $x_{\text{lake}}$ to induce an HL action $\alpha_{\text{hyd}}$. A vector of HL actions $\ba=(\alpha_{z_1},...,\alpha_{z_n})$ are determined by $F$ at any $(x,a)$. \textit{High-level} refers to any non-BL state, action, or space, (e.g. hydration or logical); only BL actions are \textit{free variables} that can be directly optimized. A mode-function $\zeta:\Sigma\rightarrow \mc E$ can change the \textit{dynamics mode}, $e\in \mc E$, of the grid-world transition kernel from $e_{\text{closed}} \rightarrow e_{\text{open}}$ when a key is obtained and registered as a one in the third bit $\bs(3)=1$, allowing the agent to traverse through the mountain pass door under $P_x(x'|x,a,e_{\text{open}})$. The agent's task is to bring honey and flowers to the friend in the northeast corner of the map.

When optimizing a policy with $P_{\bos}$, reward functions are a problem because they are difficult to define in high-dimensions and they link an agent's representations to a fixed normative quantity, e.g. value functions on a product-space are brittle if the reward or world-model changes. However, reality is dynamic, new systems can become known for an agent to control, and different goals and constraints may become relevant. States in an internal \textit{need space} $\mc Y$ could make Boolean logic task states in $\Sigma$ and BL goals on $\mc X$ salient. Agents need to flexibly reason about solutions to new complex goals and constraints (i.e. \textit{tasks}) and the feasibility that they can be satisfied within high-dimensional world-models. We believe that the key to achieving this is to propagate information about HL state-predictions and local sub-goal \textit{feasibility} to rapidly plan over both BL and HL state-spaces. Here, feasibility means probabilistic goal reachability while avoiding constraints---reachability analysis with Bellman equations has been developed and studied in stochastic hybrid systems control \cite{lygeros2004reachability, amin2006reachability, abate2008probabilistic, tkachev2013quantitative,  haesaert2018temporal}.

Given recent discussion about the generality of reward-maximization to subserve a theory of general intelligence and express notions of goal and purpose \citep{sutton1998introduction,silver2021reward,vamplew2022scalar,skalse2023limitations,bowling2023settling, ringstrom2022reward, abel2021expressivity, abel2024three}, it is important to critically investigate whether reward-maximization actually facilitates the properties of compositionality, modularity, and interpretability characteristic of general intelligence and essential for verification. Modern policy learning algorithms in RL are formalized on a foundation of reward maximization, but many important problems do not have a known, scalable model-based Bellman formalization. For example, within complex video-games like \textit{The Legend of Zelda} the problems are non-stationary (a function of time), non-Markovian (a function of history), and high-dimensional with numerous variables to track––complexity often offloaded to recurrent neural networks. However, the \textit{Reward Hypothesis}, i.e. `that goals and purposes can be well thought of as maximization of the expected value of the cumulative sum of a received scalar signal' \cite{sutton1998introduction, bowling2023settling}, suggests that a reward-maximization Bellman equation should admit solutions to these complex problems if we had an \textit{ideal} latent-space model, from pixels, of how a game's logic and dynamics are structured. While there are some decomposition results for flat and hierarchical planning problems \cite{russell2003q, dietterich2000hierarchical, todorov2009efficient, horowitz2014compositional, jonsson2016hierarchical, saxe2017hierarchy, ringstrom2020jump, infante2022globally}, these results make narrow restrictions on the reward or cost functions that limit their expressiveness. There are no known hierarchical decompositions of value functions from Bellman equations that solve a perfectly formulated sparse-reward problem at the scale of \textit{Zelda}; we believe that the reward hypothesis is a barrier to progress in this domain. 

To develop scalable Bellman-foundations, we argue for eliminating the reward function because reward maximization destroys interpretability and restricts the space of possible solutions. Instead, we let goals, constraints, and the world-model dictate the optimization of composable predictive planning representations and policies, cast into the Options Framework of RL \citep{sutton1999between}.

Researchers have investigated planning with a task-automaton which tracks the progress of non-Markovian \cite{icarte2018using}, temporal logic \citep{hasanbeig2020deep, icarte2022reward, camacho2019ltl, haesaert2018temporal, hasanbeig2019reinforcement, hasanbeig2023certified, li2017reinforcement, littman2017environment, tasseskill}, or boolean logic tasks  \cite{ringstrom2020jump, araki2021logical}. While expressive, task automata can be restrictive because they have \textit{static} dynamics, unlike a naturally evolving physiological state that changes over time. We generalize these ideas by creating solution methods over many coupled static and dynamic systems, without the required directed acyclic form of factored MDPs \cite{boutilier2000stochastic}. Skill chaining, policy-stitching, and compositional RL also share themes with our theory \cite{jothimurugan2021compositional, neary2022verifiable}. 
Our work fits into \textit{Option Models} \cite{sutton1995td, precup1998theoretical, silver2012compositional, ciosek2015value, ringstrom2020jump} where optimization involves jumping an agent from initial to final states with Bellman equations that compose options, and is similar to the Option Keyboard which uses successor features for composing options \cite{carvalho2024combining, carvalho2023composing, machado2023temporal,barreto2017successor, barreto2018transfer, barreto2019option}. Our theory can be thought of as high-dimensional option models with feasibility Bellman equations that both directly \textit{optimize and produce} an option's predictive map \textit{and} compose them to compute solutions for affordance-aware planning \cite{khetarpal2020can, khetarpal2021temporally, xu2021deep}. 

In sec. \ref{sec:TMDP} we develop a new MDP \cite{puterman2014markov} for complex goals and constraints, and define feasibility Bellman equations called \textit{Option Kernel Bellman Equations} (OKBE) that optimizes an option's composable initiation-to-termination \textit{transition kernel}. In sec. \ref{sec:CTMDP} we show how OKBE transition kernels have a critical factorization that can be reused across tasks, and they can propagate information about an option's goal and constraint satisfaction events across a high-dimensional world-model for verifiable planning. 
In sec. \ref{sec:connections} we discuss connections between OKBEs and other optimizations, including the intrinsic motivation of empowerment for goal selection \cite{salge2014empowerment}. We build on work by Ringstrom \cite{ringstrom2023reward} by formalizing a more general stationary theory for options with task constraints.




\section{The Task Markov Decision Process}\label{sec:TMDP}

We begin by defining a Task Markov Decision Process (TMDP):

\begin{definition}[Task MDP]
    A TMDP is a 5-tuple $\mathscr{M}=\langle\mc X, \mc A, P, f_{\dg},f_c\rangle$, containing a set of discrete states state $\mc X$, a set of discrete action $\mc A$, a transition kernel $P: (\mc X \times \mc A) \times \mc X \rightarrow [0,1]$, a goal function $f_{\dg}: \mc X \times \mc A \rightarrow [0,1]$, and a constraint function $f_{c}: \mc X \times \mc A \rightarrow [0,1]$. \textit{Tasks} are defined as goals and constraints.\label{TMDP_def}
\end{definition}







\subsection{Goal Function} \label{sec:goal-avail-func}
The \textit{goal function} $f_{\dg}(x,a)$ outputs a number in $[0,1]$ which represents the probability a goal is satisfied at a given state-action, and $1-f_{\dg}(x,a)$ is the probability it is not satisfied. Just like we could solve a family of regular MDPs for $N$ reward functions $\{R_{\dg_1},...,R_{\dg_N}\}$, we can also solve a family of $N$ individual TMDPs $\{f_{\dg_1},...,f_{\dg_N}\}$, indexed by goals $\mc G=\{\dg_1,...,\dg_N\}$, which group satisfaction conditions, where each solution defines an option (to be discussed in sec. \ref*{sec:CTMDP}\ref{sec:options}). 
In sec. \ref{sec:CTMDP}, goal functions will be derived from the affordance function $F$, and represent the probability an agent induces a transformation on an HL space, seen in Fig.\ref{fig:goal-constraint}. In high-dimensions, we call goal functions separable if $f_{\dg}(w,\alpha_w,...,x,a) = 1-\big(1-f_{\dg,w}(w,\alpha_w)\big)\times...\times\big(1-f_{\dg,x}(x,a)\big)$.

\subsection{Constraint Function}\label{sec:obstacles}
We also defined a constraint function $f_c$, where a constraint is violated with probability $1-f_c(x,a)$. Therefore, $f_c(x,a)=0$ encodes deterministic constraints, and $f_c(x,a)=1$ encodes free-space (see Fig.\ref{fig:goal-constraint}). 
In high dimensions, $f_c$ is separable if $f_c(w,\alpha_w,...,x,a) = f_{c,w}(w,\alpha_w)\times ...\times f_{c,x}(x,a)$, so zero-encodings mean if a constraint is violated in one function, it is violated in the entire product-space.




\subsection{State-Time Event Function}\label{sec:traj-STFF}
We now discuss \textit{state-time event functions} (STEF) for a \textit{single} goal and a set of constraints before addressing multiple dependent goals that arise later in the paper. Let $\mathbf{xt}=((x_{t_0},t_0),...,(x_{T_f},T_f))$ be a state-time trajectory over $\mc X$. We can calculate the \textit{feasibility} that a sub-trajectory of $\mathbf{xt}$ satisfies the task by choosing a starting state and time $(x_s,\tau_s)\in \mathbf{xt}$ and final state-time $(x_f,\tau_f)\in \mathbf{xt}$, where the total time is $t_f = \tau_f-\tau_s$. Since goal-completion is an event, we can use event logic for a given sub-trajectory of $\mathbf{xt}$. Consider an indicator of a goal-success event Boolean R.V. $S$, where $P(S^+)=f_{\dg}(x)$ and $P(S^-)=1-f_{\dg}(x)$, and constraint-violation event R.V. $V$, where $P(V^+)=1-f_c(x)$ and $P(V^-)=f_c(x)$ (we use capitalized realizations to avoid notational conflict). A task is completed if $(S^+,V^-)$, and is uncompleted (but not failed) if $(S^-,V^-)$. There are two varieties of STEFs that represent feasibility and infeasibility. We define a goal state-time feasibility function (STFF) $\eta^+:(\mc X) \times (\mc X \times\mc T)\rightarrow[0,1]$ that outputs the probability that the first goal-success event is at $(x_f)$ without a preceding failure event $(S^-,V^+)$, starting from $(x_{\tau_s})$ and taking $t_f$ time-steps:
\begin{align}
    \resizebox{1\hsize}{!}{$ \eta^+_{\mathbf{xt}}(x_f,t_f|x_{t_s}) = 
P((S_{\tau_s}^-,V_{\tau_s}^-),
(S_{\tau_s+1}^-, V_{\tau_s+1}^-),...,
(S_{\tau_f}^+, V_{\tau_f}^-)). $}\label{eq:logic-goal-eta}
\end{align}\\[-0.65cm]
We can express this with $f_{\dg}$ and $f_c$, which leads to a recursive form in equation \eqref{eq:eta-recursive}. Let $f_1=f_{\dg}f_c$ and $f_2=(1-f_{\dg})f_c$, we have:\\[-0.5cm]
\begin{align}
    &\resizebox{1\hsize}{!}{$\eta^+_{\mathbf{xt}}(x_f,t_f|x_{\tau_s})=\left(\mathlarger\prod_{t=0}^{t_f-1}(1-f_{\dg}(x_{\tau_s+t}))f_c(x_{\tau_s+t})\right)f_{\dg}(x_{\tau_f})f_c(x_{\tau_f}),$}\label{eq:basic-eta-1}\\[-0.5cm]
    &\implies \eta^+_{\mathbf{xt}}(x_f,t_f|x_{\tau_s}) =f_2(x_{\tau_s})\left(\prod_{t=1}^{t_f-1}f_2(x_{\tau_s+t})\right)f_1(x_{\tau_f}),\label{eq:basic-eta-2}\\
    &\implies \eta^+_{\mathbf{xt}}(x_f,t_f|x_{\tau_s}) = f_2(x_{\tau_s})\eta^+_{\mathbf{xt}}(x_f,t_f-1|x_{\tau_s+1}),\label{eq:eta-recursive}
\end{align}
with the $t_0$ condition $\eta^+_{\mathbf{xt}}(x_{j},t_0|x_{i})=f_1(x_{i})\delta_{ij}$ ($\delta$ is a Kronecker delta). A state-time infeasibility function (STIF) $\eta^{-}:(\mc X) \times (\mc X \times\mc T)\rightarrow[0,1]$ returns the probability of the first infeasibility event:
\begin{align}
    &\eta^{-}_{\mathbf{xt}}(x_f,t_f|x_{t_s}) = 
P((S_{\tau_s}^-,V_{\tau_s}^-),
(S_{\tau_s+1}^-, V_{\tau_s+1}^-),...,
(V_{\tau_f}^+))\\
    &\implies \eta^-_{\mathbf{xt}}(x_f,t_f|x_{\tau_s}) = f_2(x_{\tau_s})\eta^-_{\mathbf{xt}}(x_f,t_f-1|x_{\tau_s+1}),\label{eq:eta-fail-recursive} 
\end{align}
where $\eta^-_{\mathbf{xt}}(x_{j},t_0|x_{i})=(1-f_c(x_{i}))\delta_{ij}$ is a boundary condition.
\subsection{Achievement and Continuation Functions}
We combined the goal and constraint functions into two different functions, 
\begin{align}
    &\text{Achievement Function:}&&f_1(x,a) = f_{\dg}(x,a)f_c(x,a),\\
    &\text{Continuation Function:}&&f_2(x,a) = (1-f_{\dg}(x,a))f_c(x,a),
\end{align}
where we include actions for generality. The achievement function captures both goal-success termination events and the absence of a constraint-violation termination event, whereas the state-action dependent continuation function \cite{white2017unifying} represents the absence of both goal-success and constraint-violation events (i.e. the agent can \textit{continue} the task). It is important to understand that while goal-success and constraint-violation events will terminate the use of a control policy, so too will the event of the agent entering a state from which the goal is infeasible, which will depend on the feasibility of a goal under a policy (represented by $\kappa$, defined in the next subsection). This \textit{third} policy termination condition, not represented by $\eta_{\mathbf{xt}}$, will be used in the OKBEs (Eq. \eqref{eq:failure-at-termination}).

\subsection{Cumulative Feasibility Function}\label{sec:CFF}
Summing over $x_f$ and $t_f$ in the STFF gives us the total cumulative probability of achieving the goal while not violating the constraints. This is represented by $\kappa:\mc X \rightarrow [0,1]$, the cumulative feasibility function (CFF) \eqref{eq:basic-kappa-1}. We can substitute the R.H.S. of \eqref{eq:basic-eta-2} into \eqref{eq:basic-kappa-1}, and with some additional manipulations (not shown) we obtain a recursion for $\kappa$ in \eqref{eq:basic-kappa-rec}:
\begin{align}
    &\kappa_{\mathbf{xt}}(x_{t_s}) = \sum_{x_{f}}\sum_{t_f} \eta^+_{\mathbf{xt}}(x_{f},t_f|x_{t_s})\label{eq:basic-kappa-1}\\
    \implies &\kappa_{\mathbf{xt}}(x_{t_s})= f_1(x_{t_s})+f_2(x_{t_s})\kappa_{\mathbf{xt}}(x_{t_s+1}),\label{eq:basic-kappa-rec}
\end{align}
Here, $\kappa_{\mathbf{xt}}$ summarizes all possible events that could complete the goal. By having a recursive form of $\kappa$ and $\eta$, we can introduce a transition kernel $P_x$ to Eq. \eqref{eq:basic-kappa-rec} to define a Bellman equation. Above, we computed $\kappa$ and $\eta$ on a single trajectory $\mathbf{xt}$, but for Bellman equations we will optimize a policy $\pi$, so $\kappa$ and $\eta$ will summarize the feasibility over distributions of trajectories induced by a policy. For a sequence of policies (options), this will allows us to stitch trajectory distributions together by kernel composition (see \ref{sec:compositional}).

\subsection{Options}
The Options Framework in RL is a form of \textit{semi-Markov} planning \citep{sutton1999between}. An option $o \!= \!\langle\pi_o, \beta_o\rangle$ is a policy $\pi_o:\mc X \rightarrow \mc A$ and termination function $\beta_o: \mc X \!\rightarrow \![0,1]$. Semi-Markov means an agent follows Markovian policy dynamics $~P(x'|x,\pi_o(x))~$ until a  termination event determined by the probability $\beta_o(x)$; then, a new option's policy dynamics are initiated by an open- or closed-loop meta-policy $\mu$ and followed. Options and meta-policies are instructions for sets of policies and their switching dynamics.



\subsection{Option Kernel Bellman Equations}\label{sec:flat-OKBE}
We now introduce new Bellman Equations for optimizing a feasibility-maximizing option. The four Option Kernel Bellman Equations (OKBEs) are
defined with $\mathscr{M}=\langle\mc X, \mc A, P, f_{\dg},f_c\rangle$ and $f_1=f_{\dg}f_c$ and $f_2=(1-f_{\dg})f_c$:
\newline
\begin{align}
    &\nonumber\textbf{Policy Optimization}\text{: Optimize task success and minimize time,}\\
    &\kappa^*_{\dg}(x) = \max_{a} \left[f_1(x,a) + f_2(x,a) 
    \sum_{x'} P(x'|x,a)
    \kappa^*_{\dg}(x') \right], \label{eq:cumu_feas_2}\\
    &\pi^{**}_{\dg}(x) = \argmin_{a\in \mc A^*_{x}}\Bigg[f_2(x,a)\!\expec_{x'\sim P_a}\sum_{\mathclap{x_f,t_f}} (t_f+1) \eta^{+}_{\pi_\dg}(x_f,t_f|x') \Bigg],\label{eq:time-min-pol}\\
    &\nonumber\textbf{State-time Event Functions}\text{: Record of success and failures,} \\
    &\eta_{\pi_\dg}^{+}(x_f,\tp|x) = f_2(x,a^{\pi}_{x})
    ~~~\expec_{\mathclap{x'\sim P_{\pi}}}~~~\eta_{\pi_\dg}^{+}(\xp,\tp - 1|x'),\label{eq:asdf} \\
    &\eta_{\pi_\dg}^{-}(\xm,\tm|x) = 
    f_2(x,a^{\pi}_{x})
    ~~~\expec_{\mathclap{x'\sim P_{\pi}}}~~~\eta_{\pi_\dg}^{-}(\xm,\tm-1|x'),\label{eq:failure-eta}
\end{align}
where $a^{\pi}_{x}=\pi^{**}_{\dg}(x)$. 
Like equations \eqref{eq:eta-recursive}, \eqref{eq:eta-fail-recursive}, the above $\eta$-OKBEs are defined for $\tp,t_{\minus} > t_0$,
and if $\tp,t_{\minus}=t_0$ STEFs are defined: 
\begin{align}
    &\!\!\!\!\!\!\!\!\!\!\eta_{\pi_\dg}^{+}(x_j,t_0|x_i)=f_1(x_i,a_{x}^{\pi})\delta_{ij},\label{eq:terminal-success}\\
    &\!\!\!\!\!\!\!\!\!\!\eta_{\pi_\dg}^{-}(x_j,t_0|x_i) =\mathbbm{1}_\kappa(x_i)(1-f_c(x_i,a_{x}^{\pi})) \delta_{ij}+\bar{\mathbbm{1}}_\kappa(x_i)\delta_{ij},\label{eq:failure-at-termination}
\end{align}
where $\mathbbm{1}_\kappa(x)=\{1 ~\text{if: } \kappa^*(x)>0; ~0 ~\text{if: }\kappa^*(x)=0\}$ is the feasibility indicator function for the third termination condition, outputting $1$ if the task is feasible from $x$, $0$ if not, $\bar{\mathbbm{1}}_\kappa(x) = 1-\mathbbm{1}_\kappa(x)$ is the infeasibility indicator function,
outputting $1$ if the task is infeasible at $x$ and $0$ if not, and $\delta_{ij}$ is a Kronecker delta.
All times $t\in \mathbbm{N}_0$ express a state-relative time-to-event, and the $+1$ in Eq. \eqref{eq:time-min-pol} is for the time added by the one-step expectation.  
In the $\kappa$-OKBE \eqref{eq:cumu_feas_2}, $\kappa(x)$ has the logical interpretation: the probability of completing the goal \texttt{AND} \texttt{NOT} violating the constraint now, \texttt{OR} (+), \texttt{NOT} completing the goal \texttt{AND} \texttt{NOT} violating the constraint now, \texttt{AND} completing the goal while \texttt{NOT} violating the constraint in the future under policy $\pi^{**}_{\dg}$; this can be interpreted as \textit{planning-as-inference} \cite{kalman1960new,attias2003planning,toussaint2006probabilistic,todorov2009efficient,rawlik2013stochastic,levine2018reinforcement,ringstrom2020jump,saxe2017hierarchy,jonsson2016hierarchical}. The action $a^{\pi}_{x} \!= \!\pi_{\dg}^{**}(x)$ and two stars $**$ indicate optimal cumulative feasibility \textit{and} expected time minimization. The action set $\mc A^*_{x}$ in the $\pi$-OKBE \eqref{eq:time-min-pol} is the set of maximizing-arguments of the $\kappa$-OKBE \eqref{eq:cumu_feas_2} at $x$, so Eq. \eqref{eq:time-min-pol} minimizes the expectation over final success times conditioned on optimal cumulative feasibility. 

The $\eta$-OKBEs \eqref{eq:asdf} and \eqref{eq:failure-eta}, preserve \textit{event} probabilities for verification by back-propagating probability mass through $P_x$; the STFF, $\eta^+_{\pi_\dg}$, records when and where goals are satisfied, and the STIF $\eta^-_{\pi_\dg}$ records when and where failure occurs. For compactness, in some equations STEFs will be combined with the affordance function and policy (shown as a deterministic distribution) to include terminal actions if they are variables in $f_\dg$ and $f_c$:
\begin{align}\label{eq:term-actions}
    \eta_{\pi}(\alpha',a',x',t'|x) := F(\alpha'|x',a')\pi(a'|x')\eta_{\pi}(x',t'|x).
\end{align}
OKBEs are solved with feasibility iteration, which is a dynamic programming (DP) value iteration algorithm \cite{bellman1957markovian, bertsekas2012dynamic} (Alg. \ref{alg:stationary_feasibility_iteration}).

\subsubsection{Defining an OKBE Solution as an Option}\label{sec:options} 
OKBE solutions can be cast as options. In equations \eqref{eq:terminal-success} and \eqref{eq:failure-at-termination}, when $\kappa^*_{\dg}(x) > 0$, the task-success termination event probability is $f_1(x,a)$ and the constraint-violation termination event probability is $1 - f_c(x,a)$; and, failure occurs when $\kappa_{\dg}^*(x) = 0$, so the termination function,
\begin{align}
    \beta_{\kappa_{\dg},f_{\dg},f_c}(x,a) =
        \mathbbm{1}_{\kappa_{\dg}}(x)(f_1(x,a)+(1-f_c(x,a))) + \bar{\mathbbm{1}}_{\kappa_{\dg}}(x),
\end{align}
ensures all trajectories generated by $\pi$ contribute probability mass to an event.
At initiation, infeasible options immediately terminate and can be culled. For a task with index $\dg_i$, an option $o_{\dg_i}$ is defined:
\begin{align}
    o_{\dg_i} = \langle\pi_{\dg_i}^{**}, \beta_{\dg_i}\rangle\leftarrow \texttt{option}(\pi_{\dg_i}^{**},\kappa_{\dg_i}^{*},f_{\dg_i},f_c),\label{eq:option}
\end{align}
Options can be considered programs where optimizing option sequences is program synthesis. OKBEs optimize and construct an option's initiation-to-termination kernel $\eta^{**}_{o_\dg}$, discussed next.
\begin{figure}
    \includegraphics[width=\linewidth]{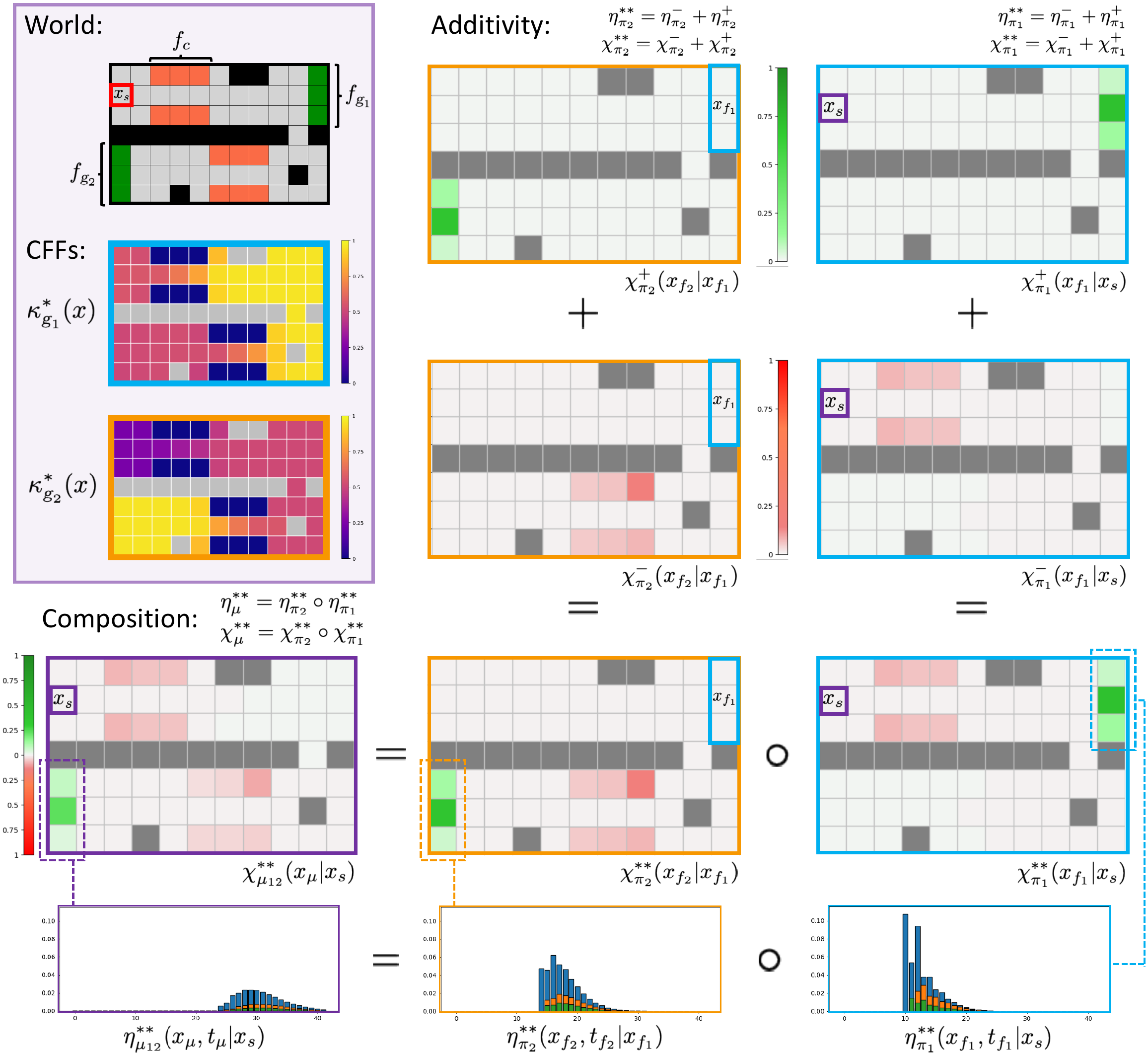}
    \caption{Compositional Predictive Maps for Verification. In the gridworld (top left), orange squares represent constraints encoded into one constraint function $f_c$. Green squares represent goal states where two sets of three goal-states are encoded into $f_{\dg_1}$ and $f_{\dg_2}$. The agent has a noisy controller which transitions the intended direction 80\% of the time, and one of the adjacent or center directions 6.67\% of the time. The CFF maps show the feasibility of each goal from a given state (yellow = 1, blue = 0). The column maps show the addition of the partial SEFs for each goal, equaling the SOK $\chi_{\pi}^{**}$ (Eq. \eqref{eq:stff-add} with time marginalized); red is constraint violation probability in $\chi_{\pi}^{-}$, green is goal completion probability in $\chi_{\pi}^{+}$. These maps are conditioned on state $x_s$ (purple square outline) and $x_{f_1}$, (blue square outline). The bottom row of panels shows the composition of two SOKs into $\chi_{\mpol_{12}}$ (Eq. \eqref{eq:SFF-compo}). The bar plots show the probability for final goal-state times of the STOKs $\eta_{\pi_1}^{**},\eta_{\pi_2}^{**},$ and $\eta_{\mpol_{12}}^{**}$ via Eq. \eqref{eq:stff-compo}.}
    \label{fig:feas-maps}
\end{figure}

\subsubsection{State-Time Option Kernel}\label{sec:kappa_eta}
The CFF and the STEFs (i.e. the STFF and STIF) are simply related through summation (Appx.\ref{appx:eta-kappa}):
\begin{align}
    \kappa_{\dg}^*(x) &= \sum_{\xp}\sum_{\tp}\eta_{\pi_\dg}^{+}(\xp,\tp|x),\label{eta-kappa-relationship-1}\\
    1-\kappa_{\dg}^*(x) &= \sum_{\xm}\sum_{\tm}\eta_{\pi_\dg}^{-}(\xm,\tm|x).\label{eta-kappa-relationship-2}
\end{align}
By addition, STEFs form a \textit{state-time option kernel} (STOK), $\eta_{o_{\dg}}^{**}$,
\begin{align}
    &\!\!\!\!\!\!\eta_{o_{\dg}}^{**}(x_f,t_f|x)=\eta_{\pi_\dg}^{+}(x_f,t_f|x)+\eta_{\pi_\dg}^{-}(x_f,t_f|x), \label{eq:stff-add}
\end{align}
which sums to $1$, $\sum_{x_f,t_f}\eta_{o_\dg}^{**}(x_f,t_f|x)\!=\!1$, following Eqs. \eqref{eta-kappa-relationship-1} and \eqref{eta-kappa-relationship-2}, making it a \textit{transition kernel} with one action, $o_\dg$.
The subscript $f$ tags \textit{final} variables, which can be $+$ or $-$. We can also marginalize time to create a State Option Kernel (SOK): $\chi_{o}^{**}(x_f|x) = \sum_{t_f}\eta_{o}^{**}(x_f,t_f|x),$
which we visualize in figure \ref{fig:feas-maps} and has additive partial State Event Functions (SEFs), $\chi_{o_\dg}^- +\chi_{o_\dg}^+=\chi_{o_{\dg}}^{**}$. SOKs are useful if the objective is not a function of time, and if the HL states of a high-dimensional problem (discussed in sec. \ref{sec:CTMDP}) do not naturally change over time (e.g. Boolean logic).

\subsubsection{Compositionality of STOKs and SOKs}\label{sec:compositional}
S(T)OKs express a distribution over interpretable termination events, so they can compose into a new S(T)OK for an \textit{abstract action} $\mpol$, an open-loop meta-policy. Let $\mpol = (o_1, o_2)$. The composition equations for option kernels, 
$\eta_{\mpol}=\eta_{o_2}\circ\eta_{o_1}$ and $\chi_{\mpol}=\chi_{o_2}\circ\chi_{o_1}$, are:
\begin{align}
    \!\!\!\!\!\!\!\!\!\!\eta_{\mpol}(x_{\mpol},t_\mpol|x)&=\sum_{x_{f_1}}\sum_{\mathclap{t_{f_1}}}\eta_{o_2}(x_{\mpol},t_{\mpol}-t_{f_1}|x_{f_1})\eta_{o_1}(x_{f_1},t_{f_1}|x),\label{eq:stff-compo}\\
    \!\!\!\!\!\!\!\!\!\chi_{\mpol}(x_{\mpol}|x)&=\sum_{x_{f_1}}\chi_{o_2}(x_{\mpol}|x_{f_1})\chi_{o_1}(x_{f_1}|x)\label{eq:SFF-compo},
\end{align}
which are Chapman-Kolmogorov equations. STOKs compose by averaging time-convolutions over intermediate states $x_{f_1}$ and are compositional predictive maps of the termination states $x_{\mu}$ and times $t_{\mu}$ of following $\mu$ (cf. SRs \cite{dayan1993improving}, which are not compositional). Composition of STOKs with terminal actions (Eq. \eqref{eq:term-actions}) requires an intermediate one-step update (Eq. \eqref{eq:GoalOp})
and STOKs will be used to define goal kernels in sec. \ref*{sec:CTMDP}.\ref{sec:goal-op} for planning from goal to goal.

\section{Compositional Task Markov Decision Processes}\label{sec:CTMDP}
We focused on TMDPs with independent goals, but problems can inherit goal-dependency structure from higher-level spaces (which may \textit{appear} non-Markovian). We now formalize the composition of a modular product-space transition kernel, and then use it to define a Compositional TMDP and OKBEs, allowing us to solve factorized STOKs with an independent sub-goal decomposition. 
\subsection{Notation}\label{sec:notation}
 For notation, we will sometimes use many \textit{specific} state-spaces: $\Sigma, \mc W, \mc X, \mc Y$. The space $\mathscr{Z} = \mc Z_1 \times ... \times \mc Z_n$ will be a product-space of \textit{generic} state-spaces (e.g. $\mc Z_3 = \mc Y$), where vectors $\bz = (z_1,...,z_n)\in \mathscr{Z}$, with variables $\{z_{k,1},...,z_{k,m}\}=\mc Z_k$. The set $\mathscr{S} = \mc X \times \mathscr{Z}$ is the full product-space, with $\mathbf{s}=(\bz,x)\in \mathscr{S}$. Generic transition kernels $P_{z,1},...,P_{z,n}$ are written $P_{z_k}(z_k'|z_k,\alpha_{z_k})$, and $P_{\bz}(\bz'|\bz,\ba_{\bz})$, $\ba_{\bz}=(\alpha_{z_1},...,\alpha_{z_n})$. 
 We will use generic notation in proofs and definitions, but not examples.


\subsection{Composition Function}
\begin{figure*}
    \centering
    \includegraphics[width=\linewidth]{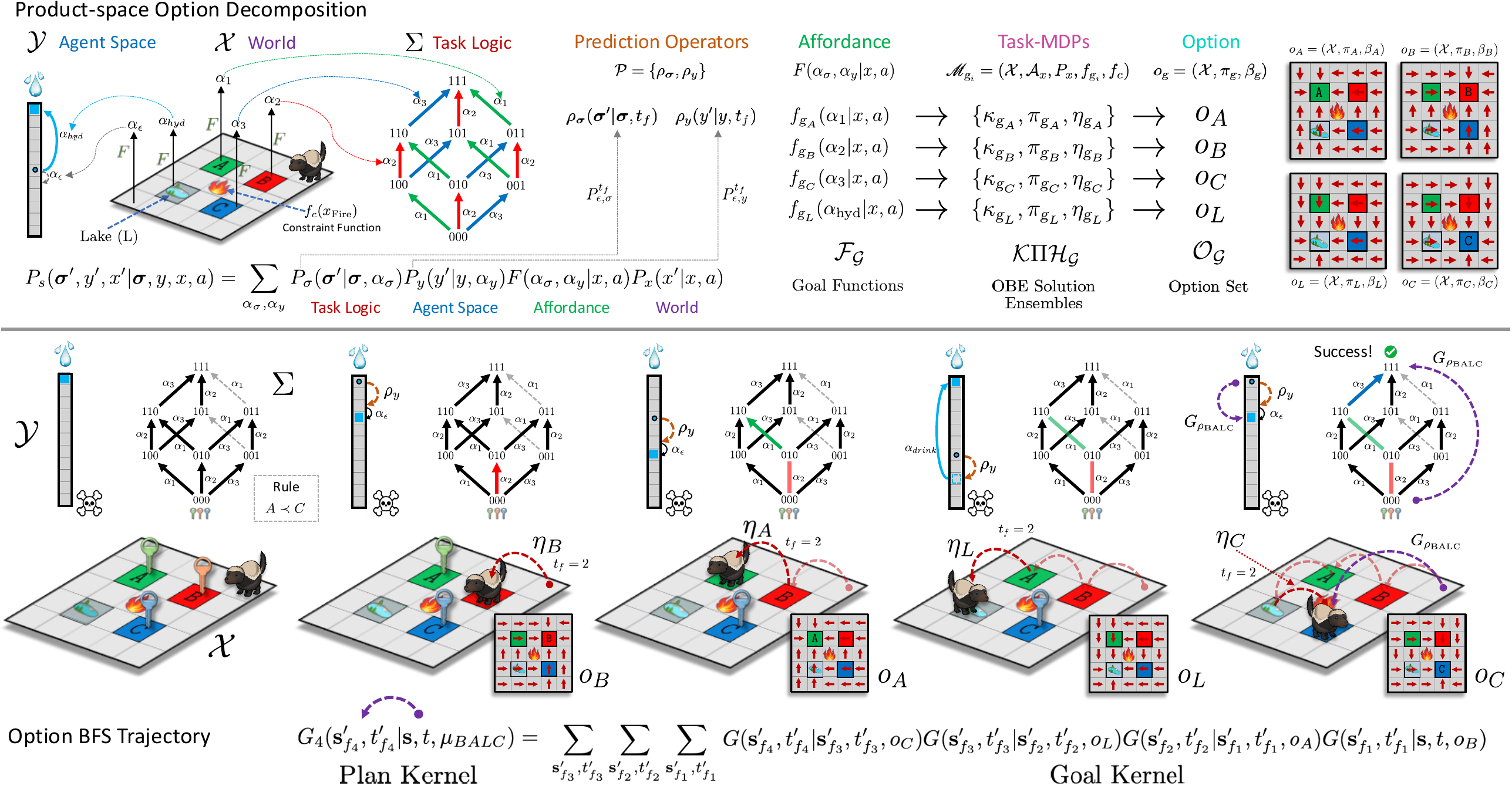}
    \caption{The affordance function $F$ links the base-space transition kernel $P_x$ with the logical task-space $P_{\bs}$ and hydration space $P_y$. The default-action $\alpha_{\ell}$ variable causes the agent to become thirstier over time, whereas $\alpha_{\text{hyd}}$ makes the agent fully hydrated by drinking at the lake (L). We can decompose $F$ into a set of goal functions $f_{\dg}$. These functions can then be used to solve OKBEs and create feasibility functions and policies defining a set of goal-conditioned options: $\mc O_{\mathcal{G}}$. The STOKs and prediction kernels are combined to create the factorization for $G$ (Eq. \eqref{eq:GoalOp}). Tree-search can solve for the optimal sequences of options using the factorization summarized by the Plan Kernel.}
    \label{fig:big-fig}
\end{figure*}

We can use the affordance function $F$ from Eq. \eqref{def:afford} to couple component transition kernels to create a composite kernel by using a composition function, $\lambda$:

\begin{definition}[Composition Functions]
    A composition function, $\lambda$, takes two transition kernels along with an affordance function $F$ and a mode function $\zeta$ (or another affordance function $\widetilde{F}$) to produce a product-space kernel (shown below for $P_{\bos}$),
\begin{align}
    P_{\bos}(\bos'|\bos,a)=&~P_{\bos}(\bz',x'|\bz,x,a)=\lambda(P_{\bz},F_x^\bz,\zeta,P_x) \label{eq:joint_kernel-compo}\\
    \nonumber:=&\sum_{\ba_{\bz},e}P_{\bz}(\bz'|\bz,\ba_{\bz})F_x^\bz(\ba_{\bz}|x,a)P_x(x'|x,a,e)\zeta(e|\bz).
\end{align}
\end{definition}
We can see an illustration of the composition function in Fig.\ref{fig:big-fig}, for the full factorization of $P_{\bos}$. The function $F_x^\bz$ has directionality ``$x$ to $\bz$". The mode function $\zeta:\mathscr{Z}\rightarrow \mc E$ is a deterministic affordance function ($F_\bz^x(e|\bz)$) that directly sets the mode variable $e$ (if it exist in $P_x$) to let the HL dynamics on $\mathscr{Z}$ condition the BL dynamics on $\mc X$.
Affordance functions can also be between HL state-spaces, $P_{\bz}=\lambda(P_{z_2},F_{z_2}^{z_1},F_{z_1}^{z_2},P_{z_1})$, and while our theorems are general enough for these cases, we do not provide examples of this kind. Also, the middle two arguments are optional. For example, $P_{\bz}(\bz'|\bz,\ba_{\bz})=\lambda(P_{z_2},P_{z_1})=P_{z_2}(z_2'|z_2,\alpha_{z_2})P_{z_1}(z_1'|z_1,\alpha_{z_1})$. 
Composite kernels can also be defined in a nested fashion: $P_{\bos}(\bos'|\bos,a)=\lambda(\lambda(P_{z_2},F_{z_2}^{z_1},F_{z_1}^{z_2},P_{z_1}),F_x^\bz,P_x) = \lambda(P_{\bz},F_x^\bz,\zeta,P_x).$

\subsection{Compositional Task MDP Definition}
We can now define a Compositional TMDP (CTMDP) using the product-space kernel:

\begin{definition}[Compositional Task MDP]
    A CTMDP $\widetilde{\mathscr{M}}=\langle\mathscr{S}, \mc A_{x}, P_{\bos}, \bar{f}_{\dg},\bar{f}_c\rangle$ is a TMDP where $P_{\bos}=\lambda(P_{\bz},F,\zeta,P_x)$ is the product-space kernel on $\mathscr{S} = \mc X \times \mathscr{Z} $.
    \label{def:uni-TMDP}
\end{definition}

The problem with the CTMDP is that it is of size $|\mathscr{S}| = |\mc X \times \mathscr{Z}|$, so dynamic programming is not practical for high-dimensional OKBEs
(omitting the $\pi$-OKBE and $\eta^-$-OKBE for brevity):  
\begin{align}
    &\widetilde{\kappa}^*_{\dg}(\bos) = \max_{a} \bigg[\bar{f}_{1}(\bos) + \bar{f}_{2}(\bos) \sum_{\bos'} \overbrace{P_{\bos}(\bos'|\bos,a)}^{\text{Eq. } \eqref{eq:joint_kernel-compo}}\widetilde{\kappa}^*_{\dg}(\bos') \bigg], \label{eq:high-dim-kappa}\\
    &\widetilde{\eta}_{\pi_\dg}^{+}(\bos_+,\tp|\bos) = \bar{f}_{2}(\bos)\expec_{\bos'\sim P_{\bos}^\pi} 
    \widetilde{\eta}_{\pi_\dg}^{+}(\bos_+,\tp \!\minus \!1|\bos').\label{eq:high-dim-goal-eta} 
\end{align}
However, obtaining $\widetilde{\kappa}_{\dg}^{*}(\bos)$, $\bar{\pi}^{**}_{\dg}(\bos)$, and $\widetilde{\eta}_{\pi_\dg}^{**}(\bos_f,t_f|\bos)$ 
would be of immense value for planning in high-dimensions, and in sec. \ref{sec:default}, \ref{sec:prediction-op}, and \ref{sec:TCV} we introduce key theory to be used for a STOK decomposition theorem in sec. \ref{sec:STFF-factorization} that will make this possible.

\subsection{Regions and Default Variables}\label{sec:default}
In high-dimensions, regions $\mc R_{\ell}\subseteq\mc X \times \mathscr{Z}$ will have a consistent \textit{default dynamics} induced by \textit{default variables}. Default variables, $\ba_{\mathbf{s}a}, e_{\mathbf{s}a}$, are relative to region state-actions ${\bos}a_i^{\ell}\in \mc R_{\ell}$ that condition $\fff(\alpha_{\ell}|\bos^i, a^i)$ and $\zeta(e_{\ell}|\bz_i)$. All state-actions $(\bos,a)^{\ell}_i$ for default variables $(\ba,e)^{\ell}$ defines one of $m$ disjoint \textit{regions} $\mc R_{\ell}\in \mathscr{R}=\{\mc R_1,...,\mc R_m\}$, $\bigcup_{\ell}\mc R_{\ell}=\mc S \times \mc A$, $$\mc R_{\ell} = \{(\bos,a) : \fff(\ba_{\ell}|\bos,a)\!=\!1 ~\land~ \zeta(e_{\ell}|\bz_{\bos})\!=\!1\},$$ of state-actions that induce \textit{consistent} HL dynamics, needed for decomposing CTMDPs. For example, the default variable of a logic-space $\Sigma$ common to most $(\bos,a)$ pairs in $\mc S\times \mc A$, is often the flip-no-bits action $\alpha_0$, meaning most state-actions in $\mc S \times \mc A$ will have no effect in the logic space. Or, a region $\mc R_{\text{Cold}}$ could induce a default variable $\alpha_{\text{cool}}$ on temperature space where the agent becomes colder (see Fig. \ref{fig:temp}). Default variables are tagged with a region index $\ell$ (e.g. $\ba_{\ell}$). We introduce regions for generality, but many problems have only one contiguous non-singleton region; for example, Fig. \ref{fig:goal-constraint} has one non-singleton region, Fig. \ref{fig:temp} has two.

\subsection{State Prediction Kernel}\label{sec:prediction-op}
It will be necessary to define a state-prediction kernel (SPK) that predicts the final state of a space when the agent executes a policy on $\mc X$. If we know that a trajectory on $\mc X$ only induces the same default variables over a time-span, we can create a default absorbing Markov chain (e.g.) $P_{z\ell c}(z'|z) = P_z(z'|z,\alpha_{\ell})$ with absorbing states corresponding to constraints in $f_c$. We define the SPK $\pfun$ to predict the final state $z_f$ when the default dynamics evolves under $P_{z\ell c}$ from $z_i$ for $t_f$ time-steps,
\begin{align}
    \pfun_{z}^{\ell}(z_f|z_i,t_f) = P_{z\ell c}^{t_f}(i,f)
    \label{eq:omega}
\end{align}
illustrated with the brown arcs in Figs. \ref{fig:big-fig}, \ref{fig:stff-decomp}.
If a space has \textit{static} dynamics (e.g. a bit-vector kernel $P_{\bs}$), then $\pfun^{\bs}_{\alpha_\ell}(\bs_f|\bs_i,t_f)=\delta_{if}$. 
A BL SPK is defined $\pfun_{\pi}^{\ell}(x_f|x_i,t_f) =P_{\pi,\ell}^{t_f}(i,f)$.

\subsection{High-level STEFs, CEFs, and TEFs}\label{sec:TCV}
A necessary piece of information to plan with HL constraints is the time until a goal, constraint, or region-exiting event occurs in another space. For instance, our agent might die of dehydration if it tries an ambitions journey. An agent can compute an HL-STEF, the probability of an HL first-event, by using the default Markov chain $\bar{P}_{z,\ell}$ obtained from clamping the action of $P_{z}$ to $\alpha_{\ell}$. A pair $(z,\alpha_z)$ will induce an HL event with probability $1-f_{3}^{\ell}(z,\alpha)$ (defined below). The HL-STEF is defined by the $\eta$-OKBE for $t_f > t_0$ and $t_f=t_0$:
\begin{align}
    &\eta_{z,\ell}(z_f,t_f|z) = 
    f_{3,z}^{\ell}(z,\alpha_{z,\ell})
    ~~~~\expec_{\mathclap{z'\sim P_{z,\ell}}}~~~~\eta_{z,\ell}(z_f,t_f-1|z'),\\[-0cm]
    &\eta_{z,\ell}(z_j,t_0|z_i)=(1-f_{3,z}^{\ell}(z_i,\alpha))\delta_{ij}, ~~~ f_{3,z}^{\ell} := (1-f_{\dg,z})f_{c,z}f_{\ell,z},\\[-0.9cm]
\end{align}
where $P_{z,\ell}\!$ is the default dynamics of $\mc Z$ for region $\mc R_{\ell}$, and $f_{\ell}(z,\alpha_z)=\{1, (z,\alpha_z)\in \mc R_{\ell}^z; ~0, ~\text{o.w.}\}$ where $\mc R_{\ell}^z$ are $z$-elements of $\mc R_{\ell}$. An HL cumulative event function (CEF) $\kappa_{z,\ell}$ returns the probability a first-event will occur in $[t_0:t_f]$ (generalizing Eq. \eqref{eq:basic-kappa-1}):
\\[-0.5cm]
\begin{align}
    \kappa_{z,\ell}(z,t_f)=\sum_{\tau_f=0}^{t_f}\sum_{z_f}\eta_{z,\ell}(z_f,\tau_f|z).
    \\[-0.6cm]
\end{align}
This means if an option is called from $(x,z_i)$ and a BL STOK duration is $t_f$ from $x$, then the option will fail (terminate early) with probability $\kappa_{z,\ell}(z,t_f)$. On the space $\mathscr{Z}$, the \textit{compliment} CEF $\bar{\kappa}_{\mathbf{z},\ell}$,
\begin{align}
    \bar{\kappa}_{\mathbf{z},\ell}(\bz,t_f)=\prod_{k}\big(1-\kappa_{k,\ell}(z^k,t_f)\big),
\end{align}
outputs the probability an HL event does not occur in region $\mc R_\ell$ of the full product-space $\mathscr{Z}$ starting from $\bz$ after $t_f$ time-steps. This will allow us to cull invalid options if $\bar{\kappa}_{\mathbf{z},\ell}(z,t_f)=0$, seen in Fig. \ref{fig:stff-decomp}. The red region is states an option cannot reach without violating an HL constraint. Blue and green indicate a feasible final state for a given space ($\mc W$ or $\mc Y$), where the most restrictive HL space (hydration) determines the validity of an option through $\bar{\kappa}_{\mathbf{z},\ell}$.

Defined with the CEF, $\kappa_{\bz}=1-\bar{\kappa}_{\bz}$, a temporal event function (TEF) $\xi_{\bz}$ returns the probability that when starting from $\bz$, no events occur in $\mathscr{Z}$ prior to $t_f$ and one or more HL events occur at $t_f$, 
\begin{align}\label{eq:TEF}
    \xi_{\bz}^{\ell}(t_f|\bz)=\kappa_{\bz,\ell}(\bz,t_f)-\kappa_{\bz,\ell}(\bz,t_f-1).
\end{align}
This TEF will be used in the STOK factorization, discussed next.
\subsection{STOK Factorization}\label{sec:STFF-factorization} A key property of the OKBE is that it entails a decomposition for a product-space STOK, which allows us to break it down into factors that we can solve for individually, thereby avoiding the prohibitive complexity of DP on $\mc X \times \mathscr{Z}$. 

Let us call the tuple $(\fff,\zeta,f_{c,\ell}^{\dg_i})$ \textit{homogeneous} if $f_{c,\ell}^{\dg_i}$ encodes as constraints all states which induce non-default variables $(\ba ,e)_{k} \neq (\ba,e)_\ell$ under $\fff$ and $\zeta$, except for the default variables associated with the goal $\dg_i$. Homogeneity implies that \textit{all} policies computed using $(\fff,\zeta,f_{c,\ell}^{\dg_i})$ are going to have a guaranteed \textit{equivalent} effect on all HL state dynamics in region $\mc R_{\ell}$. If a goal $\dg_i$ is not being achieved, only the default dynamics will be induced by the policy for $t_f$ steps, and we can make predictions of HL state-dynamics (with $\pfun$) conditionally independent of the dynamics over $\mc X$. 

We now state the STOK decomposition theorem: 

\begin{theorem}[STOK Decomposition]\label{simple-stff-corollary}
If $\bar{\mathscr{M}}=\langle\mathscr{Z},\mc X, \mc A_x, P_{\bos},f_{\dg},f_{c,\ell}\rangle$ where ($f_\dg$,$f_{c,\ell}^{\dg}$) are separable, $P_{\bos} =$ $\lambda(P_{\bz},\fff,\zeta,P_x)$, $\mc P = \{\pfun^{z_1}_{\alpha_1},...,\pfun^{z_n}_{\alpha_m}\}$, $\eta^{**}_{\pi_{\dg},\ell}$ is the STOK of TMDP $\mathscr{M}_{\dg,\ell}=\langle\mc X,\mc A,P_x,f_{\dg_i},f_{c,\ell,x}^{g}\rangle$, $(F_x^{\bz},\zeta,f_{c,\ell,x}^{g})$ is homogeneous,
$f_{\dg}$ is not a function of HL states, 
then 
the product-space STOK $\widetilde{\eta}^{**}_{\pi}$ is:
\begin{align}
     &\!\!\!\!\!\!\overbrace{\widetilde{\eta}^{**}_{\pi_i}\!(\bz_f,x_f,t_f|(\bz,x)^{\ell})}^{\text{High-dimensional STOK}}=\overbrace{\xi_{\bos}^{\ell}(t_f|\bz,x)}^{\text{TEF}}\overbrace{\pfun_{\pi_{\dg}}^{\ell}(x_f|x,t_f)}^{\text{BL SPK}}\overbrace{\pfun_{\bz}^{\ell}(\bz_f|\bz,t_f)}^{\text{Joint HL SPK}},\label{eq:stff-factor-full}\\
     &\!\!\!\!\!\!(\text{If: } 
     \bar{\kappa}_{\mathbf{z},\ell}(\mathbf{z},t_f)= 1)~~~~=\overbrace{\eta^{**}_{\pi_i,\ell}(x_f,t_f|x)}^{\text{BL STOK}}\prod_{\mathclap{k}}\overbrace{\pfun_{k}^{\ell}(z_{f}^k|z^k,t_f)}^{\text{HL SPKs}}.\label{eq:stff-factor} 
     \\[-0.65cm]
\end{align}
where
$\pfun_{\bz}^{\ell}(\bz_f|\bz,t_f)=\pfun_{z_1}(z_{1,f}|z_{1},t_f)...\pfun_{z_n}(z_{n,f}|z_{n},t_f)$ and $\xi_{\bos}^{\ell}$ is a TEF \eqref{eq:TEF} defined on $\mc X \times \mathscr{Z}$ using $\eta_{\pi,\ell}$ along with $\eta_{z,\ell}$ STEFs.
\label{uni-decomp-thm}
\end{theorem}

\begin{figure}
    \includegraphics[width=\linewidth]{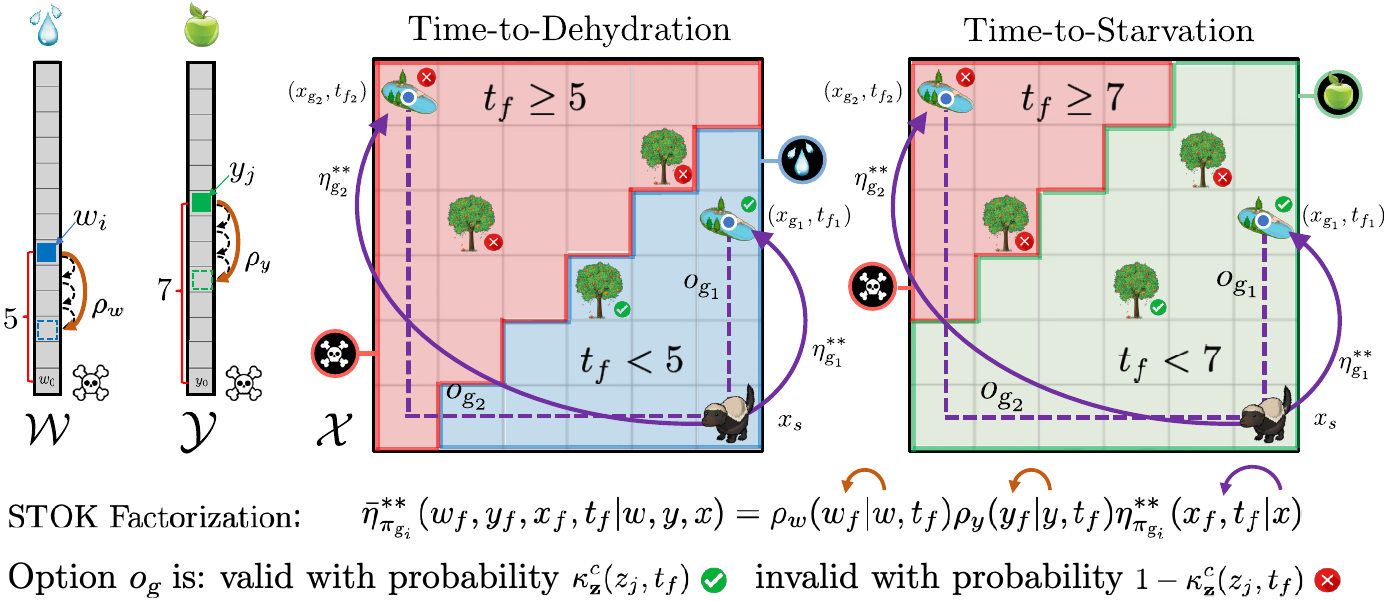}
    \caption{The STOK decomposition: The agent can plan with state-time jumps under $\eta_{\pi}$ and forecast other systems in the product-space with $\pfun$ (the final one-step update after $\alpha_{f}$ given in $\eqref{eq:GoalOp}$ not shown). 
    The compliment CEF $\bar{\kappa}_{\bz}$ determines the zones of validity for the $\eta$-predictions (solid purple arcs) and options (purple dashed lines). Blue and Green zones are the valid states corresponding to $w_i$ and $y_j$. Options are valid if they are to states outside the red zone for \textit{both} maps (Red is when $\bar{\kappa}_{z_k}(z_{k},t_f)=0$).}
    \label{fig:stff-decomp}
\end{figure}
This theorem, proved in Appx.\ref{appx:STOK-thm} and visualized in Fig. \ref{fig:stff-decomp}, says: in a region of consistent dynamics $\mc R_{\ell}$, the time that a first-event occurs has probability $\xi_{\bos}^{\ell}(t_f|\bos)$, which conditions the predictions for all other systems.  This is a chain-rule factorization of $\widetilde{\eta}_{\pi}$ along with conditional independence: $\bz_f \perp\!\!\!\!\perp (x_f,x)|(t_f, \bz)$ and $(x_f,t_f) \perp\!\!\!\!\perp  \bz | x$. If no HL events occur ($\bar{\kappa}_{\mathbf{z},\ell}(\mathbf{z},t_f)=1$), then we can simply multiply the BL STOK $\eta_{\pi,\ell}$ with HL SPKs to form the full product-space STOK $\widetilde{\eta}_{\pi}$. The decomposition enables high-dimensional forward planning; this includes dynamic physiological planning that avoids the curse of dimensionality of the value function and kernel in the Hamilton Jacobi Bellman equation of Homeostatic RL \cite{laurencon2024continuous, laurenccon2021continuous}.

\subsection{Options from CTMDP Ensembles}
Having defined a CTMDP $\bar{\mathscr{M}}=\langle\mathscr{S}, \mc A_{x}, P_{\bos}, f_{\dg},f_{c}\rangle$, we can solve for an \textit{affordance set} of \textit{point-options} \cite{jinnai2019finding} $\mc O_{\mc G}$ each terminating at a single BL state that outputs a non-default variable $\ba_{\dg}$ through $F$. This is done by taking $F_x^{\bz}$ and converting it into a set $\mc F_{\mc G}$ of goal functions for goal-state indices $\mc G=\{\dg_1,...,\dg_N\}$, where each $\dg_i$ indexes a unique non-default $(\ba,x,a)_i$ tuple in the support of $F_x^{\bz}$, meaning each state-action which induces the same HL action will have its own goal function (e.g. each tree and lake in Fig. \ref{fig:stff-decomp} is a separate goal):
\begin{align*}
    &\mc F_{\mc G} = \{f_{\dg_1},...,f_{\dg_N}\},\quad \text{where: }f_{\dg_i}(x,a) = F_x^{\bz}((\ba,x,a)_i).
\end{align*}
    Now we solve TMDPs $\mathscr{M}_{\dg_i} = \langle\mc X, \mc A, P_x, f_{\dg_i},f_{c}\rangle$ for each goal function $f_{\dg_i}\in \mc F_{\mc G}$, sharing $P_x$ and the constraint function $f_{c}^{g_i}$ which encodes goal-variables $\dg_j \neq \dg_i$ as constraints:
\begin{align*}
    &\mc K \Pi \mc H_{\mc G} \gets \texttt{ensemble\_FI}(\bar{\mathscr{M}}),~ (\kappa_{\dg_i},\pi_{\dg_i},\eta_{\dg_i}) \in \mc K \Pi \mc H_{\mc G}, ~\forall \dg_i \in \mc G,\\
    &\quad \text{where: }(\kappa_{\dg_i},\pi_{\dg_i},\eta_{\dg_i})\leftarrow\texttt{FI}(\mathscr{M}_{\dg_i} = \langle\mc X, \mc A, P_x, f_{\dg_i},f_{c}^{g_i}\rangle).
\end{align*}

The function $\texttt{ensemble\_FI}(\cdot)$ computes feasibility iteration, $\texttt{FI}(\cdot)$, on all TMDPs and returns $\mc K \Pi \mc H_{\mc G}$, a set of solution tuples that can be split into individual sets $\mc K_{\mc G}, \Pi_{\mc G}$, and $\mc H_{\mc G}$. 
A set of options $\mc O_{\mc G}$ can then be created using definition \eqref{eq:option} on each tuple of objects $o_{\dg}=\texttt{option}(\kappa_{\dg_i},\pi_{\dg_i},f_{\dg_i},f_{c,j}^{g_i})$ indexed by $\dg_i$:
\begin{align*}
    &\mc O_{\mc G} = \{o_{ \dg_1},...,o_{\dg_N}\} \gets \texttt{option\_set}(\mc K_{\mc G}, \Pi_{\mc G}, \mc F_{\mc G},f_{c,j}^{g_i}).
\end{align*}
We now have an option decomposition for the original CTMDP $\bar{\mathscr{M}}$.

\subsection{Goal Kernel}\label{sec:goal-op}
Using the sets $\mc H_{\mc G}$, $\mc O_{\mc G}$, and kernels $(P_x, P_{\bz}, \pfun_{\bz}^{\ell})$, we define a goal kernel, originally formalized by Ringstrom et al. \cite{ringstrom2020jump}, as an ensemble of STOK factorizations (Eq. \eqref{eq:stff-factor}) that maps from an initial state to a final state-time under an option:
\begin{align}
&\quad G(\bz',x',t_f+t+1|(\bz,x)^{\ell},t,o_{\ell,\dg_i})=\label{eq:GoalOp}\\
&\nonumber\resizebox{1\hsize}{!}{$~~~\mathlarger\sum_{\mathclap{\stackrel{\bz_f,\ba_{f}}{a_f,x_f,t_f}}}~\underbrace{P_{\bz}(\bz'|\bz_f,\ba_{f})P_x(x'|x_f,a_f)}_{\text{One-step boundary-action update}}\underbrace{\pfun_{\ell}(\bz_f|\bz,t_f)\eta_{o_{\ell},\dg_i}(\ba_{f},a_f,x_f,t_f|x)}_{\text{STOK Factorization}}$}
\end{align}
where $\pi_{\ell,\dg_i} \in o_{\ell,\dg_i}$ 
and $G$ is defined for all $o_{\ell,\dg_i}\in \mc O_{\mc G},~\eta_{\pi_{\ell,\dg_i}} \in \mc H_{\mc G}$. The above equation assumes determinism for HL kernels and uses Eq. \eqref{eq:stff-factor}, for stochastic kernels we use the TEF-version. For goal kernels with terminal-action STOKs and options (Eq. \eqref{eq:term-actions}), we have to evolve the dynamics forward one time-step from the boundary-action (hence the $+1$ above) to initiate the next option. Options without terminal actions do not require this update.

\subsection{Forward Optimization with Option Tree Search}
\begin{figure}[h]
    \centering
    \includegraphics[width=\linewidth]{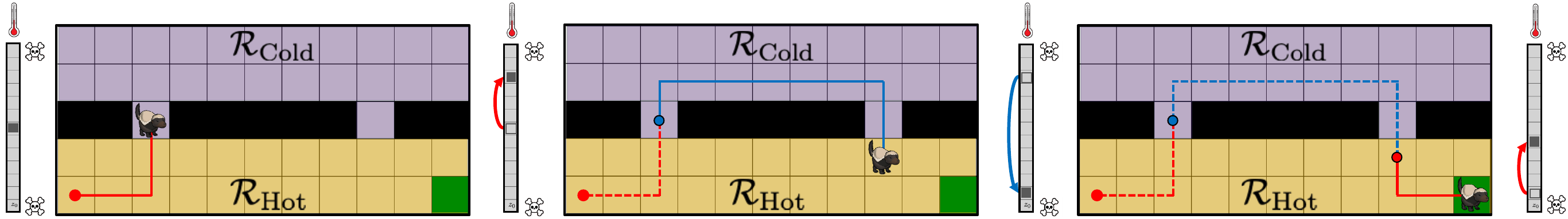}
    \caption{Regions $\mc R_{\text{Hot}}$ and $\mc R_{\text{Cold}}$ induce default variables $\alpha_{\text{Warm}}$ and $\alpha_{\text{Cool}}$ with one-step dynamics up or down in temperature space. The agent cannot go straight to the goal without overheating and must travel to $\mc R_{\text{Cold}}$ to cool down and re-enter $\mc R_{\text{Hot}}$ closer to the goal before freezing. The temperature state-space shows default dynamics predictions (red \& blue arcs) after each of the three options has terminated. }
    \label{fig:temp}
\end{figure}

Solving OKBEs using $G$ is intractable with DP, but we can optimize option sequences with tree search (TS) by forward-propagating state-vectors through $G$ to predict the resulting vectors (alg.\ref{alg:BFS}). The OKBE objective approximated by an open-loop policy $\mpol^*_{\bos}$ is:
\begin{align}   
    \!\!\!\!\!\!\!\!\kappa^*(\mathbf{s})&=\max_{o \in \mc O_{\mc G}}\left[f_1(\mathbf{s})+f_2(\mathbf{s})\expec_{\bos_{t_f}'\sim G_{\bos,o}}\kappa^*(\mathbf{s}_{t_f}')\right],\label{eq:forward-obj-1}\\
    \mpol^*_{\bos}&=\argmax_{\mpol\in \mc O^*} \kappa_{\mpol}(\bos),\label{eq:forward-obj-2}
\end{align}
where $\mpol^*_{\bos}$ is the best open-loop policy that approximates $\kappa^*(\mathbf{s})$ (exactly, under determinism), and $o^m\!\!\leftarrow\mpol^*_{\bos}(m)$ indexes the $m^{th}$ option in a sequence $\mpol$ from the set $\mc O^*$ of finite option sequences. Time-minimization has been omitted but can be incorporated. For stochastic kernels, tree search requires maintaining the joint distribution after each option, and can be achieved using sparse tensors if the number of non-zero values remains manageable, otherwise Monte-Carlo sampling is straightforward (see \ref{appx:forward_sampling}). Infeasible options are pruned, similar to temporally abstract partial models \cite{khetarpal2021temporally}. For this objective, Fig. \ref{fig:big-fig} depicts an agent solving a logic task with a precedence rule while avoiding fire and dehydration; Fig. \ref{fig:temp} shows an agent achieving a goal by traveling between hot and cold regions to regulate an internal temperature state to avoid dying.

Tree search for Eq. \eqref{eq:forward-obj-2} will find an equivalent solution to the original OKBEs \eqref{eq:high-dim-kappa} and \eqref{eq:forward-obj-1} under determinism if $\mc O$ and $\mc H$ have elements with a terminal state for each state in $\mc X$. As a theorem:

\begin{theorem}[State-Action Option Set]
Assume $\mc O_{\mc X \mc A}$ and $\mc H_{\mc X \mc A}$, are size-$|\mc X|$ sets of options and STOKs with a goal for each state in $\mc X$ paired with any final action in $\mc A$. If $\bar{\mathscr{M}}=\langle\mathscr{S} , \mc A_x, P_{\bos},\bar{f}_{\dg},\bar{f}_c\rangle$ is a deterministic CTMDP, then $\mc H_{\mc X\mc A}$ and $\mc O_{\mc X\mc A}$ is a sufficient set to find an optimal open-loop policy solution to $\bar{\mathscr{M}}$ with tree search.
\end{theorem}

The proof is trivial: if there is a feasible solution to the full problem it implies there is an optimal sequence of actions, and tree search over the state-action option set $\mc O_{\mc X \mc A}$ must contain the solution because the options can call every action from all states.

When the kernel $P_{\bs}$ is static and goals encoded in $\bar{f}_{\dg,\bs}$ are only in the higher spaces, we can create smaller sets $\mc O_{\mc X_{g}^a}$ and $H_{\mc X_{g}^a}$ with terminal goals in $\mc X_{g}^a$, which is the set of BL state-actions $(x,a)$ that induce all non-default goals $\ba_z$ through $F(\cdot|x,a)$:

\begin{theorem}[Affordance Option Set] \label{thm:static}
If $\bar{\mathscr{M}}=\langle\mathscr{S}, \mc A, P_{\bos},\bar{f}_{\dg,\bs},\bar{f}_c\rangle$, where $P_{\bos}=\lambda(P_{\bs},F,\zeta,P_x)$, and $P_{\bs}$ is static, then $\mc H_{\mc X_{g}^a}$ and $\mc O_{\mc X_{g}^a}$ are sufficient to find a solution with tree search.
\end{theorem}
A simple proof sketch: if a solution exists, then it doesn't require options to terminal states in $\mc X$ outside of $\mc X_{g}^a$ because the HL dynamics are invariant to time and only driven by states in $\mc X_{g}^a$. We show examples of solutions in Fig.\ref{fig:BFS-and-subimation} (TOP) for an illustration of tree search solving Eq. \eqref{eq:forward-obj-2} for a logic problem, and in sec. \ref{sec:verification} we show the solution for our motivating problem. Using the set $\mc O_{\mc X_{g}^a}$ over $\mc O_{\mc X\mc A}$ mitigates the complexity for the class of static problems. However, different option-sets and better pruning criteria could improve tree-search for problems with non-static transition kernels over $\mathscr{Z}$; for instance, the problem in Fig. \ref{fig:temp} can be solved with $\mc O_{\mc X \mc A}$, but intuitively there may be a more compact set of options that would suffice. We will leave this for future research.







\subsection{Modularity: Remapping Options and Tasks with Feature Functions}

The affordance function $F$ represents a directional influence between transition systems, but in order for a planning architecture to be modular we have to enrich the state-representation with features $\psi \in \Psi$ so HL actions can be indexed with richer semantics. For example, a lake should indicate the ability to drink and re-hydrate, rather than just the state's coordinate index.
\begin{figure}
    \centering
    \includegraphics[width=\linewidth]{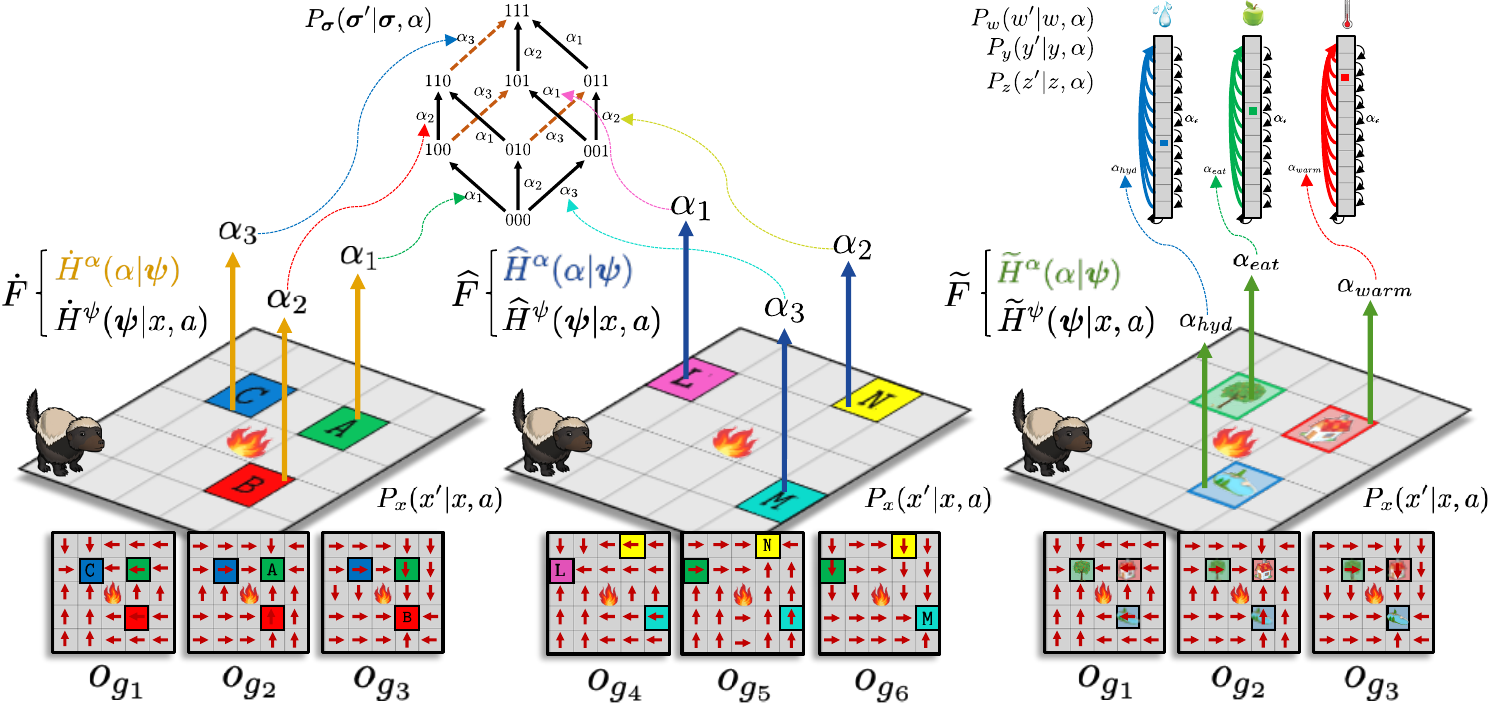}
    \caption{Option Remapping: (Left/Center) Different options with different feature functions $\dot H^\alpha$, $\dot H^\psi$, $\hat{H}^\alpha$ and $\hat{H}^\psi$ map to the same task. (Right) The same options can be remapped to different features and transition systems with $\tilde{H}^\psi$ and $\tilde{H}^\alpha$. Affordance functions $\dot F$, $\hat{F}$, and $\tilde{F}$ are defined as $F = H^\alpha \circ H^{\psi}$ between pairings.}
    \label{fig:enter-label}
\end{figure}
We can define an affordance function $F$ with feature functions $H^{\alpha}(\ba|\bpsi)$ and $H^\psi(\bpsi|x,a)$, which are functions of (bold) \textit{feature sets} $\bpsi\in 2^\Psi$. In Fig. \ref{fig:BFS-and-subimation}, we show state-actions mapping to feature sets, $\{\{\crule[blue]{0.18cm}{0.18cm},\texttt{A}\},\{\crule[blue]{0.18cm}{0.18cm},\texttt{B}\},\{\crule[blue]{0.18cm}{0.18cm},\texttt{C}\},\{\crule[green]{0.18cm}{0.18cm},\texttt{D}\},\{\crule[green]{0.18cm}{0.18cm},\texttt{E}\},\{\crule[green]{0.18cm}{0.18cm},\texttt{F}\}\}\subset 2^\Psi.$
Features map to an action in $\mc A_{\bz}$ to define an affordance function,
\begin{align*}
    F(\ba|x,a) = \sum_{\bpsi}H^{\ba}(\ba|\bpsi)H^{\psi}(\bpsi 
    |x,a).
\end{align*}
The feature functions facilitate representation reuse through kernel remapping.
If the components of the product-space kernel
$P_{\bos}=P_z\circ H^{\ba}\circ H^{\psi}\circ P_x$
change,
then options and STOKs $\eta_{\pi}$ can be remapped to new HL spaces and their SPKs $p_{\ell}$ (or vice-versa)
in the Goal Kernel (Eq. \eqref{eq:GoalOp}).
Modular goal kernels can adapt to changes in the modular primary product-space kernel $P_{\bos}$ because \textbf{only the affected sub-systems need to be updated with new local kernels} (i.e. STOKs, SPKs) to fix $G$. This modularity will also compliment a capacity called sublimated reasoning, which allows HL feasibility knowledge to be reused to solve new problems.

\subsection{Sublimation: Generalizable Knowledge from High-level TMDP Solutions}\label{sec:sublimation}

We have discussed solving high-dimensional CTMDPs, e.g. $\bar{\mathscr{M}}=\langle\Sigma \times \mc X, \mc A_x,\lambda(P_{\bs},F,\zeta,P_x),\bar{f}_{\dg},\bar{f}_c\rangle$, where the HL actions $\alpha\in \mc A_{\sigma}$ are not free variables. However, we can treat them as if they were free variables and solve problems only in the HL space, such as the TMDP: 
$\mathscr{M}_{sub,\sigma}=\langle\Sigma, \mc A_{\sigma}, P_{\bs},f_{\dg,\sigma},f_{c,\sigma}\rangle.$
Solving partial problems abstracted away from the product-space is called \textit{\underline{sub}limation}, analogous to physical sublimation where molecules break free from a solid lattice into gas. 

\begin{figure}[t]
\centering
\includegraphics[width=\linewidth]{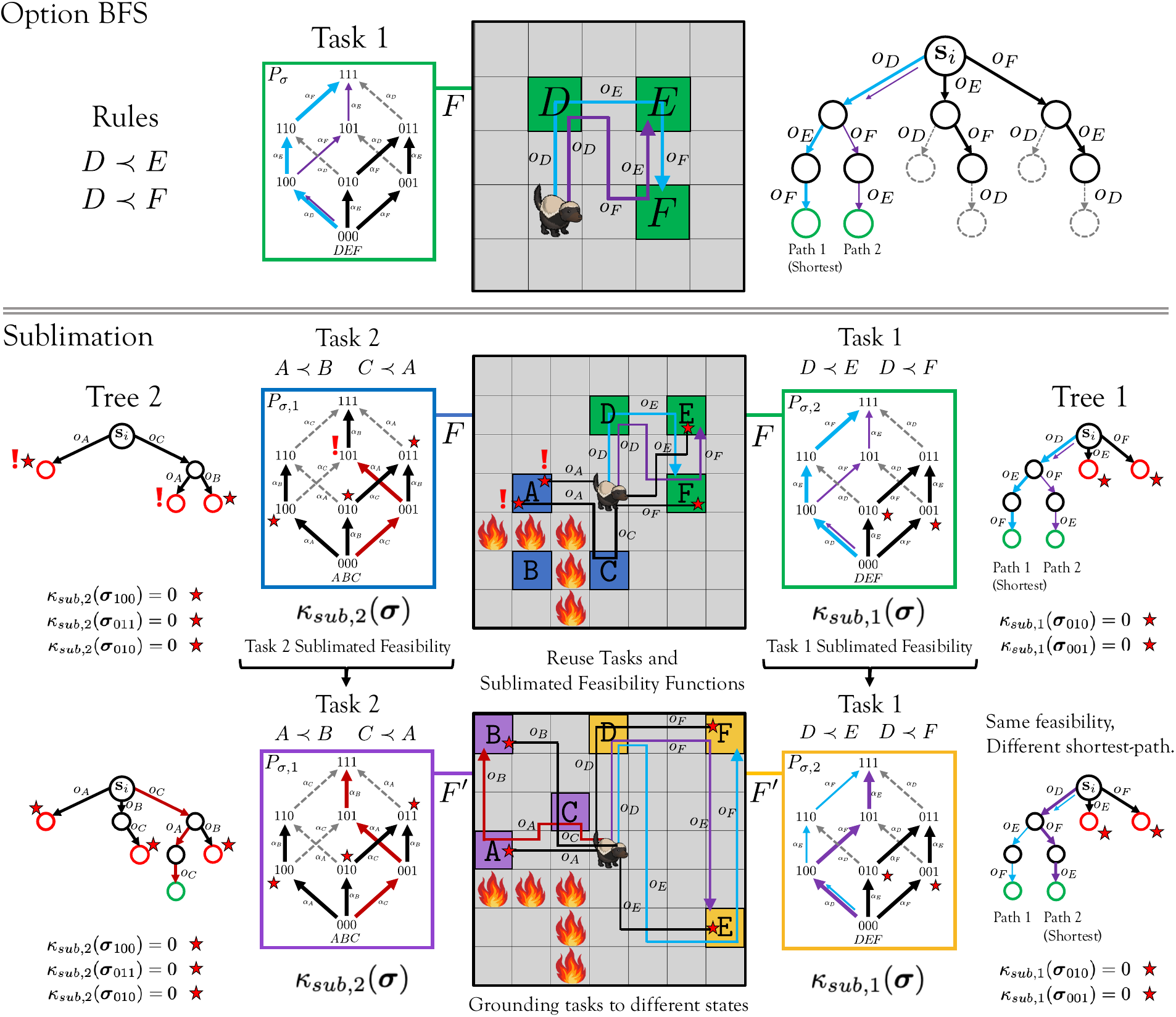}
\caption{Reusability in BL and HL spaces: (\textbf{TOP}) Same options with different tasks: The agent can use tree search to forward sample sequences of options as plans. The task rules (constraints indicated by gray dashed arrows) imposed on the logical space are $\texttt{D}\prec \texttt{E}$ and $\texttt{D}\prec \texttt{F}$, meaning only option sequences that achieve $D$ first will succeed, and other sequences run into dead-ends in logic-space. A feasible shortest-path leaf can be chosen as optimal.
(\textbf{BOTTOM}) The agent can use HL sublimated feasibility to prune options used in the BL space, where red stars indicate the absence of a feasible path ($\kappa_{sub,2}(\bs_{001})=0$), and exclamation points indicate no valid BL option calls.
Task 2 is abstractly possible ($\kappa_{sub,2}(\bs_{000})=1$) but it cannot be achieved under certain feature functions.
The lower maps shows how the sublimated feasibility functions can be reused when task kernels are remapped to different states-features $\Psi^\circ = \{\{\crule[red]{0.18cm}{0.18cm},\texttt{A}\},\{\crule[red]{0.18cm}{0.18cm},\texttt{B}\},\{\crule[red]{0.18cm}{0.18cm},\texttt{C}\},\{\crule[yellow]{0.18cm}{0.18cm},\texttt{D}\},\{\crule[yellow]{0.18cm}{0.18cm},\texttt{E}\},\{\crule[yellow]{0.18cm}{0.18cm},\texttt{F})\},$ and options. Remapping can make previously impossible tasks possible.
}
\label{fig:BFS-and-subimation}
\end{figure}

Interestingly, for certain forms of $f_{\dg,\sigma},f_{c,\sigma}$, solutions to these problems bound the feasibility of the full high-dimensional problem. For example, if the CFF $\kappa_{sub,\sigma}^*$ of a logic task informs the agent that the task is abstractly infeasible from a logic state, then it is practically infeasible from all state-vectors that include the logical state. But, even if the logical task is abstractly feasible, it does not imply that it is practically feasible from the agent's state-vector; the BL dynamics might make the problem impossible, or there could be unavoidable constraints to completing the task (see Fig.\ref{fig:BFS-and-subimation}).  

Formally, the Sublimation theorem is given as:

\begin{theorem}[Sublimation]
If $\bar{\mathscr{M}}=\langle\Sigma \times \mc X \times \mathscr{Z} , \mc A_x, P_{\bos},f_{\dg},f_c\rangle$ is a CTMDP, where $P_{\bos}=\lambda(\lambda(P_{\bs},P_{\bz}),F,\zeta,P_x)$, and $\mathscr{M}_{sub,\sigma}=\langle\Sigma, \mc A_{\sigma}, P_{\bs},f_{\dg,\sigma},f_{c,\sigma}\rangle$ is a sublimated TMDP using $P_{\bs}$ on space $\Sigma$ from $\bar{\mathscr{M}}$, where $f_{\dg,\sigma}(\boldsymbol{\sigma}) := \max_{\bz,x,a} f_{\dg}(\boldsymbol{\sigma},\bz,x,a)$ and $f_{c,\boldsymbol{\sigma}}(\boldsymbol{\sigma},\alpha_{\sigma})$ is a component from the separable constraint function $f_{c}$, then:
$$\widetilde{\kappa}^*(\boldsymbol{\sigma},\bz,x)\leq \kappa_{sub,\sigma}^*(\boldsymbol{\sigma}),$$
where $\widetilde{\kappa}^*$ is the full CFF and $\kappa_{sub,\sigma}^*$ is the CFF from $\mathscr{M}_{sub,\sigma}$.
\end{theorem}

The proof is provided in Appx.\ref{appx:sublimation}, and it generalizes to any sub-product-space in the full product-space. Sublimated feasibility information can be used to help prune tree branches that enter into an HL state which is infeasible, which is also practiced in the \textit{task and motion planning} literature in robotics \cite{garrett2021integrated, garrett2020pddlstream}, and is a high-dimensional version of affordance pruning \cite{xu2021deep, khetarpal2021temporally}. We see this in figure \ref{fig:BFS-and-subimation} (top) where the agent runs into dead-ends in logic space---the gray dashed circles indicate tree expansions that violate the task rules. Sublimated solutions allow us to terminate these branches early before reaching these gray transitions, as seen in figure \ref{fig:BFS-and-subimation} (bottom). Here, red stars indicate branch termination due to the sublimated CFF $\kappa_{sub,\sigma}^*$ returning $0$, saving $2$ node expansions for Task 1 compared to the top example. In Figure \ref{fig:BFS-and-subimation} (bottom), we also show how sublimated solutions can be reused as generalizable and transferable knowledge to solve different tasks. There are two three-goal tasks, which have an affordance function $F$ formed from feature functions, and the green/blue features. In the lower panel, the sublimated feasibility functions for the same two tasks are transferred and remapped to different BL states and features
with new feature functions $\tilde{H}^{\alpha}$ and $\tilde{H}^{\psi}$, creating a new affordance function $\tilde{F}$ and goal kernel $\tilde{G}$. The sublimated feasibility functions do not have to be re-computed, and can be used to help search over BL option sequences.  Thus, the sublimated solutions are a form of transferable and generalizable knowledge.


\begin{figure}[t]
    \centering
    \includegraphics[width=\linewidth]{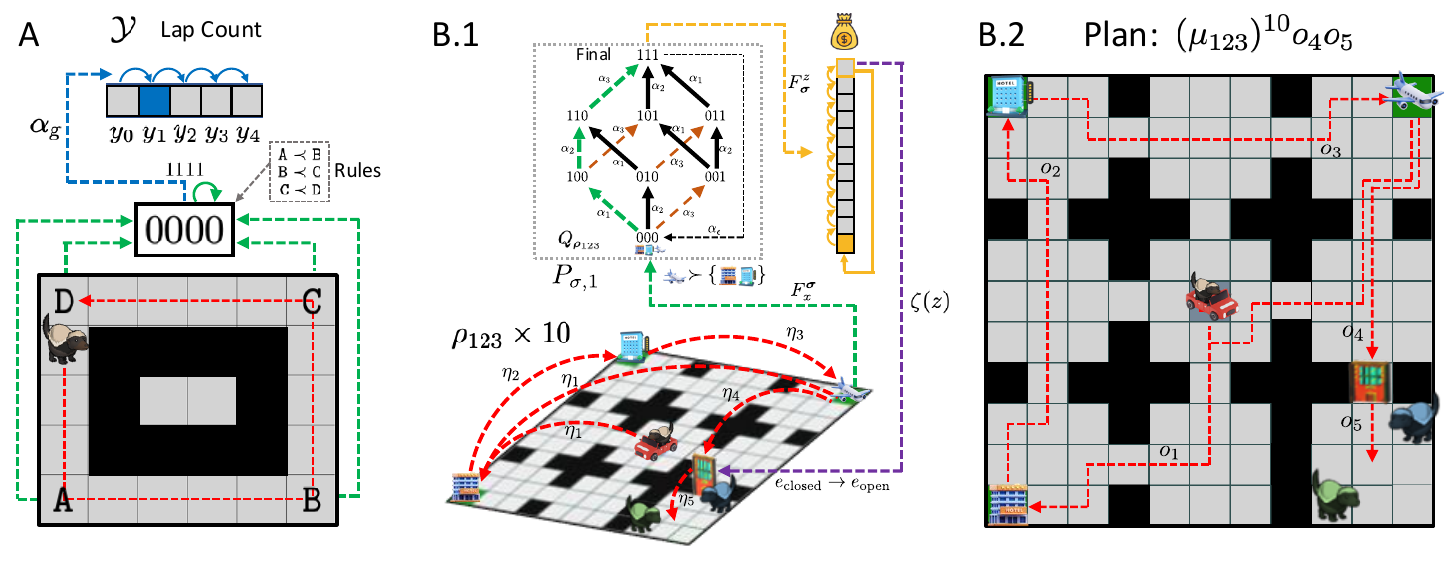}
    \caption{\textbf{A}: The agent has to complete $n$ cycles around the square hallways to complete a task, using four options for state $A, B, C,$ and $D$. 
    \textbf{B}: Transition Operators can be composed at multiple levels. The agent must pick up passengers at two hotels and take them to the airport for \$1. Sequences of options, $\mpol_{123}=(o_{\dg_1},o_{\dg_2},o_{\dg_3})$, can act as meta-actions in $m$-step plan kernel $G_m$ to solve the long horizon task of earning \$10 dollars to gain entry to the bar.}
    \label{fig:3level}
    \label{fig:3level}
\end{figure}

\subsection{Multi-level Planning with Abstract Actions}\label{sec:multi-level}
We can also compose multiple levels of hierarchy with $\lambda$. If we have a BL kernel $P_x$, and logic, wealth, and mode kernels $P_{\bs}$, $P_y$, $P_e$, then we can link them with $F_x^{\bs}$, $F_{\bs x}^y$ and $F_{yx}^e$:
\begin{align}
    &P_{\bos}(e',y',\bs',x'|e,y,\bs,x,a)=\sum_{\mathclap{\alpha_e,\alpha_y,\alpha_{\bs}}} P_e(e'|e,\alpha_e)F_{yx}^e(\alpha_e|y,x,a) \\[-0em]&\resizebox{1\hsize}{!}{$\times P_y(y'|y,\alpha_y)F^y_{\bs x}(\alpha_y|\bs,x,a) P_{\bs}(\bs'|\bs,\alpha_{\bs}) F_{x}^{\bs}(\alpha_{\bs}|x,a)P_x(x'|x,a,e),$}\label{eq:multi-lvl}
\end{align}\\[-1em]
where $F_{x}^{\bs}(\alpha_{\bs}|y,x,a)$ drives the task space, $F^y_{\bs x}(\alpha_y|\bs,x,a)$ increments the wealth space when a task is complete, and $F_{yx}^e(\alpha_e|y,x,a)$ changes the mode of the grid-world dynamics. In Fig. \ref{fig:3level}, an agent needs to repeat a three-goal task to earn \$10 (\$1 per task) to pay to get into the club to meet friends (the binary state $111$ resets to $000$ automatically, i.e. $P_{\bs}(\bs_{000}|\bs_{111},\cdot)=1$). The agent can purchase a ticket to the club with action $a_{\text{buy}}$ at $x_{\text{door}}$ if it has $y_{10} = \$10$, thereby transitioning the mode from $e_{\text{closed}}$ to $e_{\text{open}}$ through $F_{yx}^e$. Because the entire high-level space is static, Thm.\ref{thm:static} allows the agent to construct STOKs for each goal-state on $\mc X$ to build $G$ from $P_{\bos}$ and compute the plan with tree search. 

The agent can also use the solutions to the logic task, $\mpol_{123}$, as an abstract action to map from the initial state of the task to the final state of the task. Thus, the length of tree search branches can be cut down from 32 actions (repeating 3 sub-goals 10 times plus 2 sub-goals to the club) to 12 actions (10 abstract actions + 2). Thus, a tree for a size-$3$ option set would have $3^{30}$ leaves just to arrive at $\$10$ dollars, whereas using $\mpol_{123}$ with the other three options would have a tree of depth $10$ with $4^{10}$ leaves. This analysis leaves out the two final options for getting into the building, but BFS can ignore these options because their goal-feasibility is $0$ prior to arriving at $y_{10}$. We could also use only $\mpol_{123}$ in BFS (efficiently producing a ``tree" with one path), but we have no guarantees that only abstract actions will find the solution for all problems. 

In Fig. \ref{fig:3level}.A we recreate a counting task from neuroscience where a rat has to use four goal-conditioned options to cycle around hallways $4$ times with $\mpol_{\text{ABCD}}$ to get a food reward \cite{sun2020hippocampal}. Note that this is only one way of modeling the problem, it is possible to use one option with properly defined constraints to complete the task.


\begin{figure*}
    \centering
    \includegraphics[width=\linewidth]{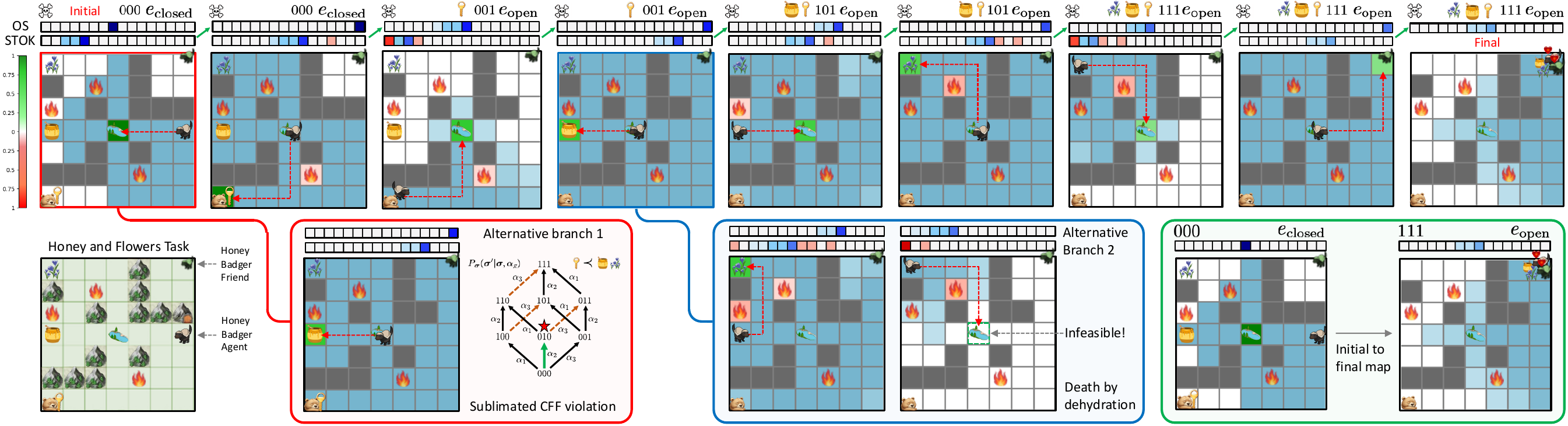}
    \caption{High-dimensional Verification: The agent must bring honey and flowers to a friend on the other side of a mountain range, and a key guarded by a sleeping bear is needed to open the mountain pass door. A task precedent constraint $\texttt{key} \prec \texttt{(honey, flowers)}$ is imposed on $P_{\bs}$ (dashed lines) as to not wake the bear with odor. The agent must obtain these items while staying alive by drinking water at the lake with a stochastic kernel $P_x$ ($95\%$ chance intended direction). The \textbf{top row} shows an optimal sequence of policies which induces probability distributions over goal satisfaction (green) and constraint violations (red) for each policy. The hydration space has two rows, one for the STOK update and one for the one-step (OS) update in Eq. \ref{eq:GoalOp}, indicated by the green arrow. Blue in the HL space represents the marginal probability of occupying an HL state after an OS or STOK update, and red squares in the HL space correspond to frozen marginal probability mass for HL \textit{or} BL constraint violations (e.g. panel two has frozen HL probability mass corresponding to the fire). This failure probability mass is subtracted from the subsequent plots for visual clarity, and we also omit bit vector marginal probabilities. Blue squares on non-task BL states indicate the probability an agent can reach the state without an HL constraint violation. The \textbf{bottom row} shows two possible different branches: one (red) where the agent chooses a goal that results in an infeasible state when evaluated by the sublimated CFF of the binary vector space and is thus pruned; the other alternative branch (blue) shows a sequence of options where all the probability mass hits the HL death state, rendering the task infeasible. The right-most panel (green) shows the initial-to-final state transition kernel given the sequence of options in the top row; the task is satisfied with p=0.3.}\label{fig:verify}
    
\end{figure*}

\subsection{Interpretability: High Dimensional Verification}\label{sec:verification}
We have shown that OKBEs preserve event-interpretability to create compositional STOK functions. STOK composition allows us to plan over options while aggregating goal success and constraint violation probability, and propagating timing information to HL state-spaces. We also showed that with sublimated CFF and CEF functions, HL feasibility information can be passed to the BL state-space to prune considered options during tree search. The STOKs, CEFs, and sublimated CFFs are three functions that participate in an bidirectional coordination of information propagation critical for scaling verification to high-dimensional world-models.

Now we can show how all of these components come together for our original motivating example in Fig. \ref{fig:goal-constraint}. Figure \ref{fig:verify} displays one full branch of the badger's option tree-search from the perspective of the STOK and SOK maps, shows zones of valid option calls (blue color over $\mc X$), and two alternative branches which violate constraints. The task is to obtain two items along with a key in order to open the door and bring the items to a friend in the top right, while staying hydrated. The plan completes the goal with probability $p_{\dg}=0.3$ and violates constraints with $p_c = 0.7$.


\section{Theoretical Connections and Considerations}\label{sec:connections}
The OKBEs have important theoretical connections, conflicts and synergies with other optimization frameworks that we now discuss.


\subsection{An Equivalence Between First-exit and Stationary OKBE Solutions Under Determinism}\label{sec:first-exit}
While the OKBE solutions do not have a direct correspondence to the standard BE solutions, an exception is first-exit (FE) objectives, where the value function $v_{\texttt{FE}}$ represents the accumulated cost until \textit{exiting} the non-boundary states and hitting a (deterministic) boundary state (i.e. goal), where the boundary-state value is set to $v_{\texttt{FE}}(x_g) = 0$. This is similar to exiting a TMDP problem by probabilistically completing the task or failing it. The FE shortest-path Bellman equation is given as \cite{bertsekas2012dynamic}:
\begin{align*}
    v_{\texttt{FE}}^*(x) &= \min_a \left[q(x)+\textstyle\expec_{P_x^a} v_{\texttt{FE}}^*(x')\right], &&\text{where:}~~v_{\texttt{FE}}^*(x_g) = 0\\
    \pi_{\texttt{FE}}^*(x) &= \argmin_a \left[q(x)+\textstyle\expec_{P_x^a} v_{\texttt{FE}}^*(x')\right].\\[-0.6cm]
\end{align*}
Constant state-costs $q(x)=c$ produce shortest-path (i.e. time-minimizing) policies so the value function $v_{\texttt{FE}}$ for a deterministic single-goal FE problem will contain all of the timing and goal-state hitting probability information. This is because the policy will only produce one shortest-path trajectory so the value function is information about that single trajectory. Thus, deterministic FE problems are a specific instance in which standard value functions are interpretable in terms of \textit{events} and \textit{time}, making them compatible with OKBE solutions. 
The time-to-goal is simply the path length, which is equal to $t_f=\frac{v_{\texttt{FE}}(x)}{c}$. 
However, determinism implies that if $x_\dg$ is not reachable from $x_i$, then $v_{\texttt{FE}}(x_i)=\infty$, and the path length to policy termination (from $x_i$ to $x_i$) is $0$. Therefore, we can extract the state-time termination information to compute:\\[-0.4cm]
\begin{align}
    &\eta^{+}_{\dg}(\xp,\tp|x) \!=\! \{
        1 ~\text{if:} (\tp = v_{\texttt{FE}}^*(x)/c)
        \land (\xp = x_g);~
        0 ~ \text{o.w.}\},\\
    &\eta^{-}_{\dg}(x_{\minus},t_{\minus}|x) \!=\! \{
        1 ~\text{if:} (v_{\texttt{FE}}^*(x) = \infty) \land (x_{\minus} = x)\land (t_{\minus} = 0);0~  \text{o.w.}\},\\[-0.9cm]
\end{align}
and $\kappa^*_{\dg}(x) = \textstyle\sum_{x_f,t_f}\eta^{+}_{\dg}(x_f,t_f|x),~\pi_{\dg}^{**}(x)=\pi_{\texttt{FE}}^*(x)$. Thus, the OKBE is like FE with constraint avoidance and with final state-time distributions over success and failure terminations.







\subsection{Reward-maximization}\label{sec:reward-max}

There is a fundamental tension between reward-maximization and high-dimensional planning. Optimally stitching policies (options) together is challenging because precise timing matters for controlling many systems. Standard RL algorithms are designed for stationarity and do not incorporate enough structure, often offloading the problem of overcoming the hidden non-Markovian and non-stationary aspects of a problem to trainable neural networks. We have provided the additional structure needed for event-based temporal reasoning. 

Making direct comparisons of our work to reward-maximization Bellman equations is tricky, as they cannot efficiently solve our problems. However, we can discuss the extra theory that would be needed to have the same capabilities of OKBE solutions in a limited stationary domain. Consider a stationary kernel $P_{\bos}$ of a logical task state-space $\Sigma$ connected to a base state-space $\mc X$ with $F_x^{\bs}$. In order to use the solutions of a reward-max objective function to construct and plan as we can with $\kappa$, $\pi$, and $\eta$, we could first solve a set of infinite-horizon discounted reward-maximization problems using \textit{sparse} reward functions $\mc R = \{r_{\dg_1},...,r_{\dg_N}\}$ (rewarding goals and penalize constraints and outputting $0$ otherwise) to generate a set of value function $\mc V = \{v_{\dg_1},...,v_{\dg_N}\}$ and policies $\Pi = \{\pi_{\dg_1},...,\pi_{\dg_N}\}$. To construct a makeshift STOK $\widetilde{\eta}_{\pi}$ with a policy, we would need to extract the goal and constraint hitting-times from the controlled Markov dynamics of the reward-max policy, $P_{\pi}(i,j)=P(x_j|x_i,\pi(x_i))$. We can define terminal states $\mc T$ as the union of rewarded and penalized states in the sparse reward-function $r$, along with the non-terminal states $\mc N$. Segregating the indices of $P_{\pi}$ into block matrices allows us to predict the events of the reward-maximization policy:\\[-0.4cm]
\begin{align*}    
\widetilde{\eta}_{\pi}(x_{j},t_f|x_i)=(P_{\pi,\mc N \mc N}^{t_f-1}P_{\pi,\mc N \mc T})(i,j) = \widetilde{P}_{\pi,t_f}(i,j),\\[-0.6cm]
\end{align*}
where $\widetilde{P}_{\pi,t_f}(i,j)$ is the probability that the agent, starting from $x_i$, hits a rewarded state $x_j$ for the first time at time $t_f$. However, $\widetilde{\eta}_{\pi}$ will not sum to one like a proper kernel if $P_{\pi}$ is not an absorbing chain where all probability mass will hit a rewarding state.

Notice the extra work required to create a makeshift STOK. We first had to solve for the value function and policy by backward induction and then, because value functions and rewards are not event-interpretable, we obtain the hitting state-time probabilities by a forward-process. However, for an OKBE, we get forward roll-out event probabilities directly only using backward induction. To construct a goal kernel from makeshift STOKs, we would have to perform the two-step process for every policy $\pi_{\dg}$ in $\Pi$. The value functions $v_{\dg}$ provide no useful information here and can be discarded, they are not useful for larger hierarchical problem unless they represent path-length or costs, which can be additively summed from sub-problems (explained in sec. \ref*{sec:connections}.\ref{sec:first-exit}).  


Unfortunately, for a sparse-reward problem in high-dimensions, there is no known reward or value function decomposition in which a set of local option value functions $v_o$ can be optimized for specific rewarded states, and summed as a sequence $\mpol$ to equal the true value function of the original problem, $v_\mpol=v^*$. Therefore, it is not value functions that are critical for cross-task generalization, but predictive representations. However, not all predictive representations have nice properties. For example, the successor representation (SR) $S_{\pi}$ with $\gamma \in [0,1)$,\\[-0.4cm]
\begin{align}
     S_\pi=\sum_{t=0}^\infty \gamma^{t}P_{\pi}^t=(I-\gamma P_{\pi})^{-1},\quad \text{where:  } \mathbf{v}_{\pi}=S_\pi\mathbf{r},\\[-0.6cm]
\end{align}
is a linear operator derived from the infinite horizon discounted Bellman equation; it produces a value function vector $\mathbf{v}_{\pi}$ given any reward function vector $\mathbf{r}$. Here, $S_\pi(i,j)$ represents the expected number of $\gamma$-weighted state-occupancies of the agent in $x_j$ when starting from $x_i$ under the controlled dynamics \cite{dayan1993improving}. 
However, unlike S(T)OKs, SRs do not have an event space, the discount factor is inseparable, they do not compose with each other to form composite SRs, and do not decompose in high-dimensions; representing occupancy statistics is not realistic in high-dimensions and plays no role in the formation of useful hierarchical abstractions. The advantage of a STOK is that occupancy statistics do not matter, 
only the final goal-success and failure event distributions do.

With a STOK, one \textit{can} compute a standard value function for an option 
up to the first reward event at a terminal state, similar to first-occupancy RL \cite{moskovitz2021first}. This can be represented as a linear equation with a discount factor $\gamma \in [0,1]$, $$v_{o}(x)=\sum_{x_f}\sum_{t_f}\eta_{o}(x_f,t_f|x)\gamma^{t_f}r(x_f)~ \equiv ~
\mathbf{v}_{o}=E_{\eta}D_{\gamma}\mathbf{r},$$ mapping a sparse reward vector $\mathbf{r}$ to a value vector $\mathbf{v}_{o}$, where $E_{\eta}$ is $\mathbbm{R}^{n_x \times n_xn_t}$ and $D_{\gamma}$ is an $\mathbbm{R}^{n_xn_t \times n_x}$ discount matrix. Thus, with many STOKs it is possible to weight each task by importance; if the world-structure changes, tasks may need to be re-weighted (in fact, \textit{empowerment} can be used for intrinsic weighting, see \ref{sec:empower}).

Reward-maximization could in principle incentivize the emergence of STOK-like factorizations, or the compositional structure previously observed in recurrent neural networks \cite{driscoll2022flexible}; this is the argument of the Reward is Enough Hypothesis (RIEH) \cite{silver2021reward}, which postulates that all capacities of general intelligence subserve the accumulation of received reward signals. However, the idea that reward-maximization is sufficient as a \textit{normative} theory is questionable, and OKBEs support a fully reward-free method for an agent to make intrinsic \textit{value judgments}, as we now explain.

\subsection{There is Value in the World-Model: Towards a Naturalistic Theory of Intrinsic Motivation and Value with Empowerment}\label{sec:empower}
~\\
We have focused on reward-free optimization for \textit{instrumental} planning---that is, finding ways to realize states of the world---but, a few words must be said about the \textit{normative} perspective: high-dimensional reward functions $r(x,...,z)$ lack a normative explanation, and value functions are semantically uninterpretable by extension. RL agents do not have the freedom to interpret why the reward signals they are responsive to are salient. Moving from rewards to tasks appears to trade one quandary for another. The question ``where do rewards come from and why are they salient?" now applies to goals and constraints. Isn't explaining the normativity and saliency of high-dimensional goals $f_{\dg}(x,...,z)$ and constraints $f_{c}(x,...,z)$ just as difficult? The formalisms, however, are not equivalent. With OKBEs, there is a reciprocity between instrumental planning and intrinsic value arising from the STOK factorization. Since OKBEs create a goal kernel that has an efficient and interpretable factorization \eqref{eq:stff-factor}, Ringstrom showed it can also be an argument to an intrinsic motivation function called empowerment \cite{klyubin2005empowerment, salge2014empowerment, tiomkin2024intrinsic} to justify the value of goals and constraints. Empowerment combines controllability with observability and quantifies the freedom of the agent to reliably effect a variety of observable state-transformations under a world-model, and it has recently 
gained attention in the cognitive sciences as a proposed mechanism for learning and exploration \cite{brandle2023empowerment}.

Empowerment is formally the Shannon channel capacity $C$ of a transition kernel (e.g.) $P(x'|x,a)$ over a horizon of $n$ actions $\mathbf{a}^n$, $$\mathfrak{E}_n(P|x_i)=C(P_n|x_i)=\max_{p(\mathbf{a}^n)} I(A_n;X_{n}|x_i),$$
which is the maximum mutual information $I$ between sequences of $n$ actions as channel inputs (R.V. $A_{n}\! \sim \!p(\mathbf{a}^n)$) to the resulting $n$-step future states as channel outputs (R.V. $X_{n}\!\sim \!P_n(x_j|x_i,A_{n}\!=\! \mathbf{a}^n) = (\prod_{k=1}^nP_{\mathbf{a}(k)})(i,j)$), starting from $x_i$, where $p(\mathbf{a})$ is the input distribution over length-$n$ action-sequences. This quantifies how much information an agent can send to its future self.

There is a problem, however, for agents that accumulate knowledge of many coupled systems: it is not practical to compute empowerment on the primitive transition kernel $P_{\bos}$ of Eq. \eqref{eq:joint_kernel} because the space of action sequences grows exponentially in $n$, limiting empowerment to short spatiotemporal scales. However, since options $o_{\dg}$ are actions in the factorization of $G(\mathbf{s}_f,t_f|\mathbf{s},t,o_{\dg})$, computing the semi-Markov option empowerment \cite{ringstrom2022reward, ringstrom2023reward}, 
\begin{align}
   \mathfrak{E}_n(G|\mathbf{s})=C(G_n|\mathbf{s})=\max_{p(\textbf{o}^n)} I(O_n;ST_{n}|\mathbf{s}),
\end{align}
quantifies the maximum mutual information between sequences of $n$ options (R.V. $O_n$) and the resulting state-time vectors (R.V. $ST_{n}$) deep into the future after $n$ goal kernel jumps. This measures the agent's ability to control over all the subsystems in its Cartesian product-space $\mathscr{S}$ that it has abstract planning representations for in $G$, from logical spaces to its own internal physiological ``need spaces." This is the contribution of the OKBEs to intrinsic value. 

We can combine instrumental reasoning with intrinsic value to see the reciprocity at work. The agent can use $G$ to (instrumentally) solve tasks into the spatiotemporally distant future, then it can make a value judgment of the plan with the \textit{gain} in high-dimensional empowerment (called \textit{valence}) on $G$, itself, at $t_{\mpol}$:
\begin{align}
    \mpol^*=\argmax_{\mpol \in \mc O^m}\expec_{\mathbf{s}_{\mpol},t_{\mpol}\sim G_m(\cdot,\cdot|\mathbf{s}_i,\mpol)}\left[\mathfrak{E}_n(G|\mathbf{s}_{t_{\mpol}})-\mathfrak{E}_n(G|\mathbf{s}_i)\right],
\end{align}
where the plan kernel $G_m$ maps an initial node to a final leaf after $m$ options in $\mpol$ sequentially condition $G$, (see Fig. \ref{fig:big-fig}).

\begin{figure}
    \centering
    \includegraphics[width=\linewidth]{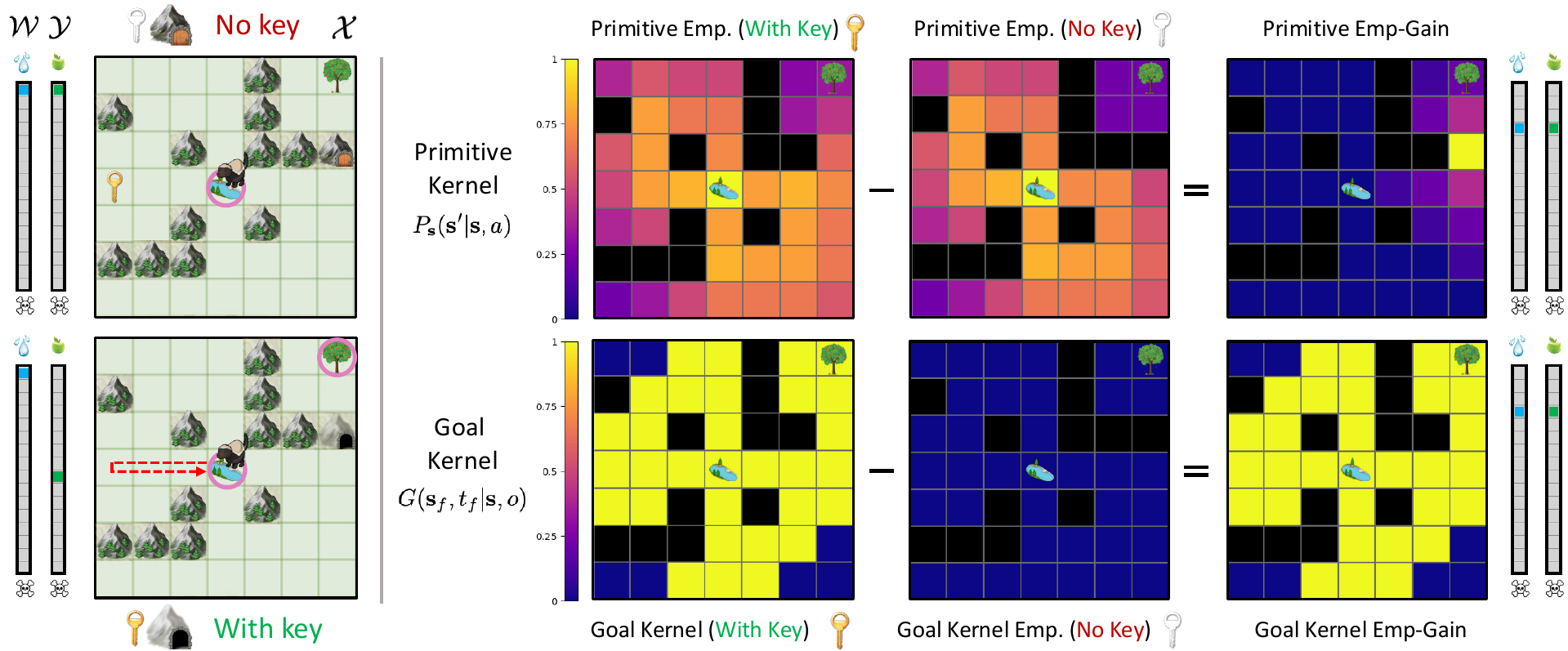}
    \caption{Empowerment Gain: (Left) Two gridworlds with kernels where the agent has a key vs. when it does not. (Right) Each row shows the absolute empowerment (AE) of the agent from each state of the gridworld and the rightmost plot shows the empowerment-gain (EG, $n=3$) which is an element-wise subtraction of the maps (all maps are normalized to $1$ (yellow)). All plots fix the internal states to $w=w_{12}$ and $y=y_{12}$. The top row shows AE and EG using the primitive kernel $P_{\bos}$, and the bottom row are the same plots using the Goal Kernel factorization of $G$, where the middle map is blue due to their being only one task that can be performed (AE of $0$). The EG of the primitive kernel is limited to a short range around the door, whereas the EG of the Goal Kernel meaningfully extends throughout the entire state-space, allowing the agent to make context sensitive value-judgments in high-dimensions.}
    \label{fig:emp}
\end{figure}


Agents can also use empowerment-gain \textit{without} instrumental planning to make value judgments about representations. If a key represented by a bit $\sigma$ can change the structure of an environment so that the agent is more expressive (or better able to sustain itself), then high-dimensional empowerment-gain can, through credit-assignment, quantify the value of a key to the agent,
\begin{align}    V(\sigma=1|\sigma=0;G,\mathbf{s},n)=\mathfrak{E}_n(G|\mathbf{s}_{\sigma_1})-\mathfrak{E}_n(G|\mathbf{s}_{\sigma_0}),\label{eq:EG}
\end{align}
where the empowerment difference is between the vectors with only a change to the bit for representing the possession of the key: $\mathbf{s}_{\sigma_1} \!= \!(\sigma_1,x,...,z)$ and $\mathbf{s}_{\sigma_0} \!= \!(\sigma_0,x,...,z)$.
Such an evaluation allows for agent-relative judgments about value, quantified as the change in the rate at which information about an agent's observable influence can propagate through a world-model. If the world-model $P_{\bos}$ changes through composition (Eq. \eqref{eq:joint_kernel-compo}), the agent can consider the value of yet-to-be realized possibilities evaluated against its constructed abstractions in $G$ (Eq. \eqref{eq:GoalOp}). Neither the OKBEs nor empowerment-gain involve received quantities, the quantities are \textit{derived} from the world-model and empowerment-gain is implicitly accumulated in the \textit{structure and state} of the agent, not a value function. Unlike utility functions which do not explain value, high-dimensional empowerment-gain is a theory of value that can adapt to an agent's idiosyncratic, history-dependent accumulation of structure, knowledge, and abstractions over a lifetime.  This agrees with perspectives that preferences and values are contextually computed and constructed, and are not normative primitives \cite{lichtenstein2006construction,srivastava2015learning,warren2011values,zhi2024beyond}. To the discussion of what constitutes an \textit{agent} \cite{sutton2022quest}, we submit: an agent is a system that can interpret the functional significance of something relative to its own structure and act on it as a reason. As Ringstrom explains, RL systems do not have this teleological capacity by design, as they have no ability to \textit{interpret} value through and for their own ontology \cite{ringstrom2023reward}.




\section{Conclusion}

In our work, we contributed new decision processes (TMDP, CTMDP) and showed how they define Option Kernel Bellman Equations (OKBEs) that directly optimize an option's initiation-to-termination spatiotemporal transition kernel, a STOK. These kernels compose by convolutional Chapman-Kolmogorov equations to create new STOKs for abstract actions. This is possible because OKBEs preserve information about goal satisfaction and constraint violation events, making STOKs predictive maps of a policy's \textit{forward} roll-out event-distribution, computed by \textit{backward} induction. In contrast, reward-maximization objectives sum rewards and costs into an expectation of a running total and fail to record spatiotemporal reward event information, thereby losing semantic interpretability in the value function (or Q-function). Furthermore, with a decomposition theorem, a high-dimensional STOK factorizes into base-level (BL) and high-level (HL) kernel components. An agent can thus optimize high-dimensional STOKs for many problems and aggregate them into a factorized goal kernel to use for verifiable planning. To mitigate some of the complexity, agents can pass BL goal feasibility information up to induce HL transition dynamics and pass HL (sublimated) feasibility and constraint-violation predictions down to the BL space to restrict which options can be used to construct goal and constraint-satisfying plans. Furthermore, we showed how both BL options and HL state-spaces can be remapped and reused without recomputation. All of this points to significant advantages of reachability over reward-maximization Bellman equations: reachability facilitates both high-dimensional instrumental planning and intrinsic motivation through empowerment, enabling context-sensitive value judgments that contemporary reinforcement learning theory does not currently capture. We believe that the factorizations we identified (Eq. \eqref{eq:joint_kernel}, Eq. \eqref{eq:stff-factor}) are critical for carving nature at its joints--it is not clear that these representations and algorithms will be discovered by current DRL network architectures with reward-maximization objectives. Therefore, we suggest that our factorizations should be key optimization targets for creating scalable world models within the latent-space of artificial neural networks. 


There are a few outstanding technical details which we did not develop in this paper. Firstly, OKBE solutions are risk-sensitive in the canonical form because the probability mass of constraint-violating event decreases the total probability of goal-satisfaction in the optimized CFF, thereby incentivizing overly cautious policies. Creating a risk-calibrated set of STOKs and options for multi-goal problems under uncertainty is an important theoretical problem we are pursuing. Secondly, we have made a seemingly intractable problem tractably \textit{representable} with the STOK factorization, and the problem can then be solved with tree-search, which can have exponential complexity. We introduced two basic theorems which state that a full option set can find deterministic solutions through tree search, and affordance option-set can find an optimal solution to product-spaces with static high-level transition kernels.  However, a full option set can have a prohibitive branching factor in tree search for large state-spaces, and even static state-spaces can lead to exponential blow-up in tree-search for large binary vector task-spaces. We believe improvements can be found which makes tree-search more efficient, whether it be by restricting the option/STOK set, or by using improved pruning criteria. Thirdly, we did not provide ways of computing closed-loop meta-policies over options to find optimal solutions to stochastic problems with tree-search, this remains an important problem for future work. 

We have shown that OKBEs promote compositionality, modularity, and interpretability, and we argued that STOKs are fundamental structures of generalization and flexibility. In doing so, we provided an alternative to computing or approximating value functions in high-dimensional space, and we made the case that reachability optimizations directly promote flexibility. We anticipate that taking advantage of the structure of compositional transition kernels will be pivotal for future progress in the fields of high-dimensional planning and intrinsic motivation.






\newpage
\section*{Appendix}
\bibliography{pnas-sample}

\begin{thebibliography}{100}

\bibitem{keramati2014homeostatic}
M Keramati, B Gutkin, Homeostatic reinforcement learning for integrating reward collection and physiological stability.
\newblock {\em\protect\JournalTitle{Elife}} \textbf{3}, e04811 (2014).

\bibitem{laurenccon2021continuous}
H Lauren{\c{c}}on, CR S{\'e}gerie, J Lussange, BS Gutkin, Continuous homeostatic reinforcement learning for self-regulated autonomous agents.
\newblock {\em\protect\JournalTitle{arXiv preprint arXiv:2109.06580}} (2021).

\bibitem{laurencon2024continuous}
H Laurencon, et~al., Continuous time continuous space homeostatic reinforcement learning (ctcs-hrrl): Towards biological self-autonomous agent.
\newblock {\em\protect\JournalTitle{arXiv preprint arXiv:2401.08999}} (2024).

\bibitem{gaon2020reinforcement}
M Gaon, R Brafman, Reinforcement learning with non-markovian rewards in {\em Proceedings of the AAAI conference on artificial intelligence}.
\newblock Vol.{}~34, pp. 3980--3987 (2020).

\bibitem{icarte2018using}
RT Icarte, T Klassen, R Valenzano, S McIlraith, Using reward machines for high-level task specification and decomposition in reinforcement learning in {\em International Conference on Machine Learning}.
\newblock (PMLR), pp. 2107--2116 (2018).

\bibitem{abel2016near}
D Abel, D Hershkowitz, M Littman, Near optimal behavior via approximate state abstraction in {\em International Conference on Machine Learning}.
\newblock (PMLR), pp. 2915--2923 (2016).

\bibitem{konidaris2019necessity}
G Konidaris, On the necessity of abstraction.
\newblock {\em\protect\JournalTitle{Current opinion in behavioral sciences}} \textbf{29}, 1--7 (2019).

\bibitem{ho2019value}
MK Ho, D Abel, TL Griffiths, ML Littman, The value of abstraction.
\newblock {\em\protect\JournalTitle{Current opinion in behavioral sciences}} \textbf{29}, 111--116 (2019).

\bibitem{pickett2002policyblocks}
M Pickett, AG Barto, Policyblocks: An algorithm for creating useful macro-actions in reinforcement learning in {\em ICML}.
\newblock Vol.{}~19, pp. 506--513 (2002).

\bibitem{topin2015portable}
N Topin, et~al., Portable option discovery for automated learning transfer in object-oriented markov decision processes. in {\em IJCAI}.
\newblock pp. 3856--3864 (2015).

\bibitem{lake2017building}
BM Lake, TD Ullman, JB Tenenbaum, SJ Gershman, Building machines that learn and think like people.
\newblock {\em\protect\JournalTitle{Behavioral and brain sciences}} \textbf{40}, e253 (2017).

\bibitem{du2021unsupervised}
Y Du, S Li, Y Sharma, J Tenenbaum, I Mordatch, Unsupervised learning of compositional energy concepts.
\newblock {\em\protect\JournalTitle{Advances in Neural Information Processing Systems}} \textbf{34}, 15608--15620 (2021).

\bibitem{du2024compositional}
Y Du, L Kaelbling, Compositional generative modeling: A single model is not all you need.
\newblock {\em\protect\JournalTitle{arXiv preprint arXiv:2402.01103}} (2024).

\bibitem{lecun2022path}
Y LeCun, A path towards autonomous machine intelligence version 0.9. 2, 2022-06-27.
\newblock {\em\protect\JournalTitle{Open Review}} \textbf{62} (2022).

\bibitem{tsividis2021human}
PA Tsividis, et~al., Human-level reinforcement learning through theory-based modeling, exploration, and planning.
\newblock {\em\protect\JournalTitle{arXiv preprint arXiv:2107.12544}} (2021).

\bibitem{team2021open}
OEL Team, et~al., Open-ended learning leads to generally capable agents.
\newblock {\em\protect\JournalTitle{arXiv preprint arXiv:2107.12808}} (2021).

\bibitem{dalrymple2024towards}
D Dalrymple, et~al., Towards guaranteed safe ai: A framework for ensuring robust and reliable ai systems.
\newblock {\em\protect\JournalTitle{arXiv preprint arXiv:2405.06624}} (2024).

\bibitem{leike2017ai}
J Leike, et~al., Ai safety gridworlds.
\newblock {\em\protect\JournalTitle{arXiv preprint arXiv:1711.09883}} (2017).

\bibitem{seshia2022toward}
SA Seshia, D Sadigh, SS Sastry, Toward verified artificial intelligence.
\newblock {\em\protect\JournalTitle{Communications of the ACM}} \textbf{65}, 46--55 (2022).

\bibitem{molinaro2023goal}
G Molinaro, AG Collins, A goal-centric outlook on learning.
\newblock {\em\protect\JournalTitle{Trends in Cognitive Sciences}} (2023).

\bibitem{juechems2019does}
K Juechems, C Summerfield, Where does value come from?
\newblock {\em\protect\JournalTitle{Trends in cognitive sciences}} \textbf{23}, 836--850 (2019).

\bibitem{dayan1993improving}
P Dayan, Improving generalization for temporal difference learning: The successor representation.
\newblock {\em\protect\JournalTitle{Neural Computation}} \textbf{5}, 613--624 (1993).

\bibitem{momennejad2017successor}
I Momennejad, et~al., The successor representation in human reinforcement learning.
\newblock {\em\protect\JournalTitle{Nature human behaviour}} \textbf{1}, 680--692 (2017).

\bibitem{stachenfeld2017hippocampus}
KL Stachenfeld, MM Botvinick, SJ Gershman, The hippocampus as a predictive map.
\newblock {\em\protect\JournalTitle{Nature neuroscience}} \textbf{20}, 1643--1653 (2017).

\bibitem{gershman2018successor}
SJ Gershman, The successor representation: its computational logic and neural substrates.
\newblock {\em\protect\JournalTitle{Journal of Neuroscience}} \textbf{38}, 7193--7200 (2018).

\bibitem{brunec2022predictive}
IK Brunec, I Momennejad, Predictive representations in hippocampal and prefrontal hierarchies.
\newblock {\em\protect\JournalTitle{Journal of Neuroscience}} \textbf{42}, 299--312 (2022).

\bibitem{carvalho2024predictive}
W Carvalho, MS Tomov, W de~Cothi, C Barry, SJ Gershman, Predictive representations: Building blocks of intelligence.
\newblock {\em\protect\JournalTitle{Neural Computation}} pp. 1--74 (2024).

\bibitem{piray2021linear}
P Piray, ND Daw, Linear reinforcement learning in planning, grid fields, and cognitive control.
\newblock {\em\protect\JournalTitle{Nature communications}} \textbf{12}, 4942 (2021).

\bibitem{piray2024reconciling}
P Piray, ND Daw, Reconciling flexibility and efficiency: Medial entorhinal cortex represents a compositional cognitive map.
\newblock {\em\protect\JournalTitle{bioRxiv}} pp. 2024--05 (2024).

\bibitem{sutton1999between}
RS Sutton, D Precup, S Singh, Between mdps and semi-mdps: A framework for temporal abstraction in reinforcement learning.
\newblock {\em\protect\JournalTitle{Artificial intelligence}} \textbf{112}, 181--211 (1999).

\bibitem{xia2021temporal}
L Xia, AG Collins, Temporal and state abstractions for efficient learning, transfer, and composition in humans.
\newblock {\em\protect\JournalTitle{Psychological review}} \textbf{128}, 643 (2021).

\bibitem{reverberi2012compositionality}
C Reverberi, K G{\"o}rgen, JD Haynes, Compositionality of rule representations in human prefrontal cortex.
\newblock {\em\protect\JournalTitle{Cerebral cortex}} \textbf{22}, 1237--1246 (2012).

\bibitem{lake2015human}
BM Lake, R Salakhutdinov, JB Tenenbaum, Human-level concept learning through probabilistic program induction.
\newblock {\em\protect\JournalTitle{Science}} \textbf{350}, 1332--1338 (2015).

\bibitem{frankland2020concepts}
SM Frankland, JD Greene, Concepts and compositionality: in search of the brain's language of thought.
\newblock {\em\protect\JournalTitle{Annual review of psychology}} \textbf{71}, 273--303 (2020).

\bibitem{driscoll2022flexible}
L Driscoll, K Shenoy, D Sussillo, Flexible multitask computation in recurrent networks utilizes shared dynamical motifs.
\newblock {\em\protect\JournalTitle{Biorxiv}} pp. 2022--08 (2022).

\bibitem{davidson2022creativity}
G Davidson, T Gureckis, B Lake, Creativity, compositionality, and common sense in human goal generation. psyarxiv (2022).

\bibitem{zhou2024compositional}
Y Zhou, BM Lake, A Williams, Compositional learning of functions in humans and machines.
\newblock {\em\protect\JournalTitle{arXiv preprint arXiv:2403.12201}} (2024).

\bibitem{mark2023flexible}
S Mark, et~al., Flexible and abstract neural representations of abstract structural knowledge.
\newblock {\em\protect\JournalTitle{bioRxiv}} pp. 2023--08 (2023).

\bibitem{el2023cellular}
M El-Gaby, et~al., A cellular basis for mapping behavioural structure.
\newblock {\em\protect\JournalTitle{bioRxiv}} pp. 2023--11 (2023).

\bibitem{samborska2022complementary}
V Samborska, JL Butler, ME Walton, TE Behrens, T Akam, Complementary task representations in hippocampus and prefrontal cortex for generalizing the structure of problems.
\newblock {\em\protect\JournalTitle{Nature Neuroscience}} \textbf{25}, 1314--1326 (2022).

\bibitem{kurth2023replay}
Z Kurth-Nelson, et~al., Replay and compositional computation.
\newblock {\em\protect\JournalTitle{Neuron}} \textbf{111}, 454--469 (2023).

\bibitem{eltetHo2023habits}
N {\'E}ltet{\H{o}}, P Dayan, Habits of mind: Reusing action sequences for efficient planning.
\newblock {\em\protect\JournalTitle{arXiv preprint arXiv:2306.05298}} (2023).

\bibitem{sharma2021map}
S Sharma, A Curtis, M Kryven, J Tenenbaum, I Fiete, Map induction: Compositional spatial submap learning for efficient exploration in novel environments.
\newblock {\em\protect\JournalTitle{arXiv preprint arXiv:2110.12301}} (2021).

\bibitem{bakermans2025constructing}
JJ Bakermans, J Warren, JC Whittington, TE Behrens, Constructing future behavior in the hippocampal formation through composition and replay.
\newblock {\em\protect\JournalTitle{Nature Neuroscience}} pp. 1--12 (2025).

\bibitem{sun2020hippocampal}
C Sun, W Yang, J Martin, S Tonegawa, Hippocampal neurons represent events as transferable units of experience.
\newblock {\em\protect\JournalTitle{Nature neuroscience}} \textbf{23}, 651--663 (2020).

\bibitem{bellman1957markovian}
R Bellman, A markovian decision process.
\newblock {\em\protect\JournalTitle{Journal of mathematics and mechanics}} pp. 679--684 (1957).

\bibitem{puterman2014markov}
ML Puterman, {\em Markov decision processes: discrete stochastic dynamic programming}.
\newblock (John Wiley \& Sons), (2014).

\bibitem{manna1971toward}
Z Manna, RJ Waldinger, Toward automatic program synthesis.
\newblock {\em\protect\JournalTitle{Communications of the ACM}} \textbf{14}, 151--165 (1971).

\bibitem{bastani2018verifiable}
O Bastani, Y Pu, A Solar-Lezama, Verifiable reinforcement learning via policy extraction.
\newblock {\em\protect\JournalTitle{Advances in neural information processing systems}} \textbf{31} (2018).

\bibitem{inala2020synthesizing}
JP Inala, O Bastani, Z Tavares, A Solar-Lezama, Synthesizing programmatic policies that inductively generalize in {\em 8th International Conference on Learning Representations}.
\newblock (2020).

\bibitem{trivedi2021learning}
D Trivedi, J Zhang, SH Sun, JJ Lim, Learning to synthesize programs as interpretable and generalizable policies.
\newblock {\em\protect\JournalTitle{Advances in neural information processing systems}} \textbf{34}, 25146--25163 (2021).

\bibitem{qiu2022programmatic}
W Qiu, H Zhu, Programmatic reinforcement learning without oracles in {\em The Tenth International Conference on Learning Representations}.
\newblock (2022).

\bibitem{silver2020few}
T Silver, KR Allen, AK Lew, LP Kaelbling, J Tenenbaum, Few-shot bayesian imitation learning with logical program policies in {\em Proceedings of the AAAI Conference on Artificial Intelligence}.
\newblock Vol.{}~34, pp. 10251--10258 (2020).

\bibitem{cui2024reward}
G Cui, Y Wang, W Qiu, H Zhu, Reward-guided synthesis of intelligent agents with control structures.
\newblock {\em\protect\JournalTitle{Proceedings of the ACM on Programming Languages}} \textbf{8}, 1730--1754 (2024).

\bibitem{lygeros2004reachability}
J Lygeros, 9.
\newblock {\em\protect\JournalTitle{Automatica}} \textbf{40}, 917--927 (2004).

\bibitem{amin2006reachability}
S Amin, A Abate, M Prandini, J Lygeros, S Sastry, Reachability analysis for controlled discrete time stochastic hybrid systems in {\em Hybrid Systems: Computation and Control: 9th International Workshop, HSCC 2006, Santa Barbara, CA, USA, March 29-31, 2006. Proceedings 9}.
\newblock (Springer), pp. 49--63 (2006).

\bibitem{abate2008probabilistic}
A Abate, M Prandini, J Lygeros, S Sastry, Probabilistic reachability and safety for controlled discrete time stochastic hybrid systems.
\newblock {\em\protect\JournalTitle{Automatica}} \textbf{44}, 2724--2734 (2008).

\bibitem{tkachev2013quantitative}
I Tkachev, A Mereacre, JP Katoen, A Abate, Quantitative automata-based controller synthesis for non-autonomous stochastic hybrid systems in {\em Proceedings of the 16th international conference on Hybrid systems: computation and control}.
\newblock pp. 293--302 (2013).

\bibitem{haesaert2018temporal}
S Haesaert, S Soudjani, A Abate, Temporal logic control of general markov decision processes by approximate policy refinement.
\newblock {\em\protect\JournalTitle{IFAC-PapersOnLine}} \textbf{51}, 73--78 (2018).

\bibitem{sutton1998introduction}
RS Sutton, AG Barto, {\em Reinforcement Learning: {A}n Introduction}.
\newblock (The MIT Press, Cambridge, MA), (1998).

\bibitem{silver2021reward}
D Silver, S Singh, D Precup, RS Sutton, Reward is enough.
\newblock {\em\protect\JournalTitle{Artificial Intelligence}} \textbf{299}, 103535 (2021).

\bibitem{vamplew2022scalar}
P Vamplew, et~al., Scalar reward is not enough: A response to silver, singh, precup and sutton (2021).
\newblock {\em\protect\JournalTitle{Autonomous Agents and Multi-Agent Systems}} \textbf{36}, 41 (2022).

\bibitem{skalse2023limitations}
J Skalse, A Abate, On the limitations of markovian rewards to express multi-objective, risk-sensitive, and modal tasks in {\em Uncertainty in Artificial Intelligence}.
\newblock (PMLR), pp. 1974--1984 (2023).

\bibitem{bowling2023settling}
M Bowling, JD Martin, D Abel, W Dabney, Settling the reward hypothesis in {\em International Conference on Machine Learning}.
\newblock (PMLR), pp. 3003--3020 (2023).

\bibitem{ringstrom2022reward}
TJ Ringstrom, Reward is not necessary: How to create a modular \& compositional self-preserving agent for life-long learning.
\newblock {\em\protect\JournalTitle{arXiv preprint arXiv:2211.10851}} (2022).

\bibitem{abel2021expressivity}
D Abel, et~al., On the expressivity of markov reward.
\newblock {\em\protect\JournalTitle{Advances in Neural Information Processing Systems}} \textbf{34}, 7799--7812 (2021).

\bibitem{abel2024three}
D Abel, MK Ho, A Harutyunyan, Three dogmas of reinforcement learning.
\newblock {\em\protect\JournalTitle{arXiv preprint arXiv:2407.10583}} (2024).

\bibitem{russell2003q}
SJ Russell, A Zimdars, Q-decomposition for reinforcement learning agents in {\em Proceedings of the 20th international conference on machine learning (ICML-03)}.
\newblock pp. 656--663 (2003).

\bibitem{dietterich2000hierarchical}
TG Dietterich, Hierarchical reinforcement learning with the maxq value function decomposition.
\newblock {\em\protect\JournalTitle{Journal of artificial intelligence research}} \textbf{13}, 227--303 (2000).

\bibitem{todorov2009efficient}
E Todorov, Efficient computation of optimal actions.
\newblock {\em\protect\JournalTitle{Proceedings of the national academy of sciences}} \textbf{106}, 11478--11483 (2009).

\bibitem{horowitz2014compositional}
MB Horowitz, EM Wolff, RM Murray, A compositional approach to stochastic optimal control with co-safe temporal logic specifications. in {\em IROS}.
\newblock pp. 1466--1473 (2014).

\bibitem{jonsson2016hierarchical}
A Jonsson, V G{\'o}mez, Hierarchical linearly-solvable markov decision problems. in {\em ICAPS}.
\newblock pp. 193--201 (2016).

\bibitem{saxe2017hierarchy}
AM Saxe, AC Earle, B Rosman, Hierarchy through composition with multitask lmdps in {\em International Conference on Machine Learning}.
\newblock (PMLR), pp. 3017--3026 (2017).

\bibitem{ringstrom2020jump}
TJ Ringstrom, M Hasanbeig, A Abate, Jump operator planning: Goal-conditioned policy ensembles and zero-shot transfer.
\newblock {\em\protect\JournalTitle{arXiv preprint arXiv:2007.02527}} (2020).

\bibitem{infante2022globally}
G Infante, A Jonsson, V G{\'o}mez, Globally optimal hierarchical reinforcement learning for linearly-solvable markov decision processes in {\em Proceedings of the AAAI Conference on Artificial Intelligence}.
\newblock Vol.{}~36, pp. 6970--6977 (2022).

\bibitem{hasanbeig2020deep}
M Hasanbeig, D Kroening, A Abate, Deep reinforcement learning with temporal logics in {\em International Conference on Formal Modeling and Analysis of Timed Systems}.
\newblock (Springer), pp. 1--22 (2020).

\bibitem{icarte2022reward}
RT Icarte, TQ Klassen, R Valenzano, SA McIlraith, Reward machines: Exploiting reward function structure in reinforcement learning.
\newblock {\em\protect\JournalTitle{Journal of Artificial Intelligence Research}} \textbf{73}, 173--208 (2022).

\bibitem{camacho2019ltl}
A Camacho, RT Icarte, TQ Klassen, RA Valenzano, SA McIlraith, Ltl and beyond: Formal languages for reward function specification in reinforcement learning. in {\em IJCAI}.
\newblock Vol.{}~19, pp. 6065--6073 (2019).

\bibitem{hasanbeig2019reinforcement}
M Hasanbeig, et~al., Reinforcement learning for temporal logic control synthesis with probabilistic satisfaction guarantees in {\em 2019 IEEE 58th conference on decision and control (CDC)}.
\newblock (IEEE), pp. 5338--5343 (2019).

\bibitem{hasanbeig2023certified}
H Hasanbeig, D Kroening, A Abate, Certified reinforcement learning with logic guidance.
\newblock {\em\protect\JournalTitle{Artificial Intelligence}} \textbf{322}, 103949 (2023).

\bibitem{li2017reinforcement}
X Li, CI Vasile, C Belta, Reinforcement learning with temporal logic rewards in {\em 2017 IEEE/RSJ International Conference on Intelligent Robots and Systems (IROS)}.
\newblock (IEEE), pp. 3834--3839 (2017).

\bibitem{littman2017environment}
ML Littman, et~al., Environment-independent task specifications via gltl.
\newblock {\em\protect\JournalTitle{arXiv preprint arXiv:1704.04341}} (2017).

\bibitem{tasseskill}
GN Tasse, D Jarvis, S James, B Rosman, Skill machines: Temporal logic skill composition in reinforcement learning in {\em The Twelfth International Conference on Learning Representations}.
\newblock (2024).

\bibitem{araki2021logical}
B Araki, et~al., The logical options framework in {\em International Conference on Machine Learning}.
\newblock (PMLR), pp. 307--317 (2021).

\bibitem{boutilier2000stochastic}
C Boutilier, R Dearden, M Goldszmidt, Stochastic dynamic programming with factored representations.
\newblock {\em\protect\JournalTitle{Artificial intelligence}} \textbf{121}, 49--107 (2000).

\bibitem{jothimurugan2021compositional}
K Jothimurugan, S Bansal, O Bastani, R Alur, Compositional reinforcement learning from logical specifications.
\newblock {\em\protect\JournalTitle{Advances in Neural Information Processing Systems}} \textbf{34}, 10026--10039 (2021).

\bibitem{neary2022verifiable}
C Neary, C Verginis, M Cubuktepe, U Topcu, Verifiable and compositional reinforcement learning systems in {\em Proceedings of the International Conference on Automated Planning and Scheduling}.
\newblock Vol.{}~32, pp. 615--623 (2022).

\bibitem{sutton1995td}
RS Sutton, Td models: Modeling the world at a mixture of time scales in {\em Machine Learning Proceedings 1995}.
\newblock (Elsevier), pp. 531--539 (1995).

\bibitem{precup1998theoretical}
D Precup, RS Sutton, S Singh, Theoretical results on reinforcement learning with temporally abstract options in {\em Machine Learning: ECML-98: 10th European Conference on Machine Learning Chemnitz, Germany, April 21--23, 1998 Proceedings 10}.
\newblock (Springer), pp. 382--393 (1998).

\bibitem{silver2012compositional}
D Silver, K Ciosek, Compositional planning using optimal option models.
\newblock {\em\protect\JournalTitle{arXiv preprint arXiv:1206.6473}} (2012).

\bibitem{ciosek2015value}
K Ciosek, D Silver, Value iteration with options and state aggregation.
\newblock {\em\protect\JournalTitle{arXiv preprint arXiv:1501.03959}} (2015).

\bibitem{carvalho2024combining}
WC Carvalho, et~al., Combining behaviors with the successor features keyboard.
\newblock {\em\protect\JournalTitle{Advances in Neural Information Processing Systems}} \textbf{36} (2024).

\bibitem{carvalho2023composing}
W Carvalho, A Filos, RL Lewis, S Singh, , et~al., Composing task knowledge with modular successor feature approximators.
\newblock {\em\protect\JournalTitle{arXiv preprint arXiv:2301.12305}} (2023).

\bibitem{machado2023temporal}
MC Machado, A Barreto, D Precup, M Bowling, Temporal abstraction in reinforcement learning with the successor representation.
\newblock {\em\protect\JournalTitle{Journal of Machine Learning Research}} \textbf{24}, 1--69 (2023).

\bibitem{barreto2017successor}
A Barreto, et~al., Successor features for transfer in reinforcement learning.
\newblock {\em\protect\JournalTitle{Advances in neural information processing systems}} \textbf{30} (2017).

\bibitem{barreto2018transfer}
A Barreto, et~al., Transfer in deep reinforcement learning using successor features and generalised policy improvement in {\em International Conference on Machine Learning}.
\newblock (PMLR), pp. 501--510 (2018).

\bibitem{barreto2019option}
A Barreto, et~al., The option keyboard: Combining skills in reinforcement learning.
\newblock {\em\protect\JournalTitle{Advances in Neural Information Processing Systems}} \textbf{32} (2019).

\bibitem{khetarpal2020can}
K Khetarpal, Z Ahmed, G Comanici, D Abel, D Precup, What can i do here? a theory of affordances in reinforcement learning in {\em International Conference on Machine Learning}.
\newblock (PMLR), pp. 5243--5253 (2020).

\bibitem{khetarpal2021temporally}
K Khetarpal, Z Ahmed, G Comanici, D Precup, Temporally abstract partial models.
\newblock {\em\protect\JournalTitle{Advances in Neural Information Processing Systems}} \textbf{34}, 1979--1991 (2021).

\bibitem{xu2021deep}
D Xu, et~al., Deep affordance foresight: Planning through what can be done in the future in {\em 2021 IEEE international conference on robotics and automation (ICRA)}.
\newblock (IEEE), pp. 6206--6213 (2021).

\bibitem{salge2014empowerment}
C Salge, C Glackin, D Polani, Empowerment--an introduction in {\em Guided Self-Organization: Inception}.
\newblock (Springer), pp. 67--114 (2014).

\bibitem{ringstrom2023reward}
TJ Ringstrom, ``Reward Is Not Necessary: Foundations for Compositional Non-Stationary Non-Markovian Hierarchical Planning and Intrinsically Motivated Autonomous Agents,'' PhD thesis,  University of Minnesota (2023).

\bibitem{white2017unifying}
M White, Unifying task specification in reinforcement learning in {\em International Conference on Machine Learning}.
\newblock (PMLR), pp. 3742--3750 (2017).

\bibitem{kalman1960new}
RE Kalman, A new approach to linear filtering and prediction problems.
\newblock {\em\protect\JournalTitle{~}} (1960).

\bibitem{attias2003planning}
H Attias, Planning by probabilistic inference in {\em International workshop on artificial intelligence and statistics}.
\newblock (PMLR), pp. 9--16 (2003).

\bibitem{toussaint2006probabilistic}
M Toussaint, A Storkey, Probabilistic inference for solving discrete and continuous state markov decision processes in {\em Proceedings of the 23rd international conference on Machine learning}.
\newblock pp. 945--952 (2006).

\bibitem{rawlik2013stochastic}
K Rawlik, M Toussaint, S Vijayakumar, On stochastic optimal control and reinforcement learning by approximate inference.
\newblock {\em\protect\JournalTitle{~}} (2013).

\bibitem{levine2018reinforcement}
S Levine, Reinforcement learning and control as probabilistic inference: Tutorial and review.
\newblock {\em\protect\JournalTitle{arXiv preprint arXiv:1805.00909}} (2018).

\bibitem{bertsekas2012dynamic}
D Bertsekas, {\em Dynamic programming and optimal control: Volume I}.
\newblock (Athena scientific) Vol.{}~1, (2012).

\bibitem{jinnai2019finding}
Y Jinnai, D Abel, D Hershkowitz, M Littman, G Konidaris, Finding options that minimize planning time in {\em International Conference on Machine Learning}.
\newblock pp. 3120--3129 (2019).

\bibitem{garrett2021integrated}
CR Garrett, et~al., Integrated task and motion planning.
\newblock {\em\protect\JournalTitle{Annual review of control, robotics, and autonomous systems}} \textbf{4}, 265--293 (2021).

\bibitem{garrett2020pddlstream}
CR Garrett, T Lozano-P{\'e}rez, LP Kaelbling, Pddlstream: Integrating symbolic planners and blackbox samplers via optimistic adaptive planning in {\em Proceedings of the international conference on automated planning and scheduling}.
\newblock Vol.{}~30, pp. 440--448 (2020).

\bibitem{moskovitz2021first}
T Moskovitz, SR Wilson, M Sahani, A first-occupancy representation for reinforcement learning.
\newblock {\em\protect\JournalTitle{arXiv preprint arXiv:2109.13863}} (2021).

\bibitem{klyubin2005empowerment}
AS Klyubin, D Polani, CL Nehaniv, Empowerment: A universal agent-centric measure of control in {\em 2005 ieee congress on evolutionary computation}.
\newblock (IEEE), Vol.{}~1, pp. 128--135 (2005).

\bibitem{tiomkin2024intrinsic}
S Tiomkin, I Nemenman, D Polani, N Tishby, Intrinsic motivation in dynamical control systems.
\newblock {\em\protect\JournalTitle{PRX Life}} \textbf{2}, 033009 (2024).

\bibitem{brandle2023empowerment}
F Br{\"a}ndle, LJ Stocks, JB Tenenbaum, SJ Gershman, E Schulz, Empowerment contributes to exploration behaviour in a creative video game.
\newblock {\em\protect\JournalTitle{Nature Human Behaviour}} \textbf{7}, 1481--1489 (2023).

\bibitem{lichtenstein2006construction}
S Lichtenstein, P Slovic, The construction of preference: An overview.
\newblock {\em\protect\JournalTitle{The construction of preference}} \textbf{1}, 1--40 (2006).

\bibitem{srivastava2015learning}
N Srivastava, P Schrater, Learning what to want: context-sensitive preference learning.
\newblock {\em\protect\JournalTitle{PloS one}} \textbf{10}, e0141129 (2015).

\bibitem{warren2011values}
C Warren, AP McGraw, L Van~Boven, Values and preferences: defining preference construction.
\newblock {\em\protect\JournalTitle{Wiley Interdisciplinary Reviews: Cognitive Science}} \textbf{2}, 193--205 (2011).

\bibitem{zhi2024beyond}
T Zhi-Xuan, M Carroll, M Franklin, H Ashton, Beyond preferences in ai alignment.
\newblock {\em\protect\JournalTitle{arXiv preprint arXiv:2408.16984}} (2024).

\bibitem{sutton2022quest}
RS Sutton, The quest for a common model of the intelligent decision maker.
\newblock {\em\protect\JournalTitle{arXiv preprint arXiv:2202.13252}} (2022).

\end{thebibliography}
\section*{ }



\newpage
\onecolumn
\section*{Author Note:}
The proofs in this appendix are under peer-review, but all results, such as the STOK factorization, have been empirically tested and verified through computation.

\section*{Glossary}

\begin{longtable}{ll}
  \textbf{Term} & \textbf{Definition and Description} \\
  \hline
  \endfirsthead
  \textbf{Term} & \textbf{Definition and Description} \\
  \hline
  \endhead
  \hline
  \endfoot
  \textbf{Acronyms} & \\
  \hline
  BL & Base-Level: The foundational state-space in a hierarchical system. \\
  CEF & Cumulative Event Function: A function that sums the probability of achieving a goal over time. \\
  CTMDP & Compositional Task Markov Decision Process: A decision process for high-dimensional, modular planning. \\
  DP & Dynamic Programming: A method for solving complex problems by breaking them into simpler subproblems. \\
  DRL & Deep Reinforcement Learning: A combination of deep learning and reinforcement learning. \\
  FE & First-Exit: A problem where the goal is to reach a boundary state. \\
  HL & High-Level: Abstract state-spaces or actions in a hierarchical system. \\
  MDP & Markov Decision Process: A framework for modeling sequential decision-making in stochastic environments. \\
  OKBE & Option Kernel Bellman Equations: Bellman equations for optimizing and constructing an option kernel. \\
  RL & Reinforcement Learning: A machine learning paradigm where agents learn by interacting with an environment. \\
  SR & Successor Representation: A predictive representation of state occupancy under a policy. \\
  STEF & State-Time Event Function: A function that predicts the probability of events at a given state-time (e.g. STIF, STFF, STOK). \\
  STFF & State-Time Feasibility Function: A function that predicts the feasibility of achieving a goal over time. \\
  STIF & State-Time Infeasibility Function: A function that predicts the probability of failing to achieve a goal. \\
  STOK & State-Time Option Kernel: A transition kernel for options that predicts state-time outcomes. \\
  SOK & State Option Kernel: A simplified version of STOK that marginalizes over time. \\
  TEF & Temporal Event Function: Predicts the probability of an event occurring at a specific time. \\
  TMDP & Task Markov Decision Process: A MDP for tasks defined by goals and constraints. \\
  \hline
  \textbf{Variables} & \\
  \hline
  $\alpha$ & High-level action variable: Represents actions in high-level state-spaces. \\
  $\beta$ & Termination function: Determines when an option terminates. \\
  $\bs$ & Binary state vector: A vector of binary states (e.g., task completion flags). \\
  $\delta$ & Kronecker delta: A function that returns 1 if inputs are equal, otherwise 0. \\
  $\gamma$ & Discount factor: A factor that reduces the weight of future rewards or events. \\
  $\psi$ & Feature variable: Represents a feature in a feature set. \\
  $\bpsi$ & Feature set: A set of features used for mapping states to actions. \\
  $\tau$ & Time index: A specific time step in a trajectory. \\
  $\bos$ & Full state vector: The product of base-level and high-level state vectors. \\
  $t$ & Time variable: Represents discrete time steps. \\
  $x$ & Base-level state variable: Represents the agent's current state in the base-level space. \\
  $z$ & High-level state variable: Represents abstract states (e.g., hydration level). \\
  $\bz$ & High-level state vector: A vector of high-level state variables. \\
  \hline
  \textbf{Functions} & \\
  \hline
  $f_1$ & Achievement function: Combines goal and constraint functions ($f_{\dg} \cdot f_c$). \\
  $f_2$ & Continuation function: Combines the negation of the goal function with the constraint function ($(1 - f_{\dg}) \cdot f_c$). \\
  $f_c$ & Constraint function: Defines the probability of violating a constraint at a given state-action. \\
  $f_{\dg}$ & Goal function: Defines the probability of satisfying a goal at a given state-action. \\
  $\mathfrak{E}$ & Empowerment: A measure of an agent's ability to control its environment. \\
  $F$ & Affordance function: Links base-level actions to high-level state transformations. \\
  $G$ & Goal kernel: Maps initial high-dimensional state-vector (and time) to final state-vector and time under an option. \\
  $H^{\alpha}$ & Feature-to-action map: Maps features to high-level actions. \\
  $H^{\psi}$ & State-to-feature map: Maps states to features. \\
  $\lambda$ & Composition function: Combines transition kernels into a product-space kernel. \\
  $P$ & Transition kernel: Defines the probability of transitioning between states. \\
  $\chi$ & State Option Kernel (SOK): A simplified version of STOK that marginalizes over time. \\
  $\pfun$ & State Prediction Kernel (SPK): A function that predicts the final state after $t_f$ time steps.\\
  $\eta^+$ & State-Time Feasibility Function (STFF): Predicts the probability of achieving a task over time. \\
  $\eta^-$ & State-Time Infeasibility Function (STIF): Predicts the probability of failing a task over time. \\
  $\eta_z$ & State-Time Event Function (STEF): Predicts the probability of inducing an event in a space (e.g. $\mc Z$) at a given state-time. \\
  $\eta^{**}_{\pi}$ & State-Time Option Kernel (STOK): Probability distribution over success and failure events under a policy $\pi$ (Is a STEF). \\
  $\kappa^{*}_{\pi}$ & Cumulative Feasibility Function (CFF): Sums the probability of achieving a task across all times. \\
  $\kappa_z$ & Cumulative Event Function (CEF): Sums the probability of an event happening up to a given final time. \\
  $\bar{\kappa}$ & Compliment Cumulative Event Function: Sums the probability that an event does not happen up to a given final time. \\
  $\mpol$ & Meta-policy: An open or closed-loop policy that outputs options. Can be used as an abstract action. \\
  $\pi$ & Policy: A function that maps states to actions. \\
  $\xi$ & Temporal Event Function (TEF): Predicts the probability of an event occurring at a specific time. \\
  $\zeta$ & Mode function: A function that sets the dynamics mode of a transition kernel conditioned on another state-space. \\
  $v$ & Value Function: Represents the expected cumulative reward from a state under a policy. \\
  \hline
  \textbf{Sets} & \\
  \hline
  $\mc A$ & Action space: A set of actions. \\
  $\mc A_{z}$ & High-level action space: A set of high-level actions. \\
  $\mc A_{\bz}$ & High-level action product space: The Cartesian product set of all possible high-level actions. \\
  $\mc F$ & Affordance goal functions: A set of goal functions $\{f_{\dg_1}, f_{\dg_2}, \dots\}$, each specifying a goal for a state of a non-null action of $F$. \\
  $\mc G$ & Goal set: A set of goals $\{\dg_1, \dg_2, \dots\}$, where each goal is associated with a goal function $f_{\dg_i}$. \\
  $\mc H$ & STOK set: A set of STOKs for $n$ individual problems. \\
  $\mc O$ & Option set: A set of options $\{o_1, o_2, \dots\}$, where each option is a policy paired with a termination function. \\
  $\mc R$ & Region: Defines a consistent set of default dynamics. \\
  $\mathscr{R}$ & Set of all regions: A set of regions $\{\mc R_1, \mc R_2, \dots\}$. \\
  $\mathscr{S}$ & Full product space: A Cartesian product of all state-spaces, both base-level and high-level, $\mathscr{S} = \mc X \times \mathscr{Z}$. \\
  $\mc V$ & Set of value functions \\  
  $\mc X$ & Base-level state space: The set of all possible base-level states. \\
  $\mc Z$ & High-level state space: The set of all possible high-level states (e.g., hydration levels). \\
  $\mathscr{Z}$ & A Cartesian product of all high-level state-spaces. \\
  $\Pi$ & Policy set: A set of policies for $n$ individual problems.\\

\end{longtable}

\newpage

\section{State-Time Option Kernels Sum to One}\label{appx:eta-kappa}
Given a TMDP and its OKBE solutions $\kappa^+_{\pi},\pi,\eta_{\pi}^{+},\eta_{\pi}^{-}$, we will prove the following three equations:
\begin{align}
    &\kappa^+_{\pi}(x_i)=\sum_{\xp}\sum_{\tp}\eta^+_{\pi}(\xp,\tp|x_i),\label{eq:no1}\\
    &\kappa^-_{\pi}(x_i)=\sum_{\xm}\sum_{\tm}\eta^-_{\pi}(\xm,\tm|x_i),\label{eq:no2}\\
    &\sum_{x_f}\sum_{t_f}\eta^{**}_{\pi_o}(x_f,t_f|x_i)=1,\qquad\text{where:}~~\eta^{**}_{\pi_o}(x_f,t_f|x_i):=\eta^-_{\pi}(x_f,t_f|x_i) + \eta^+_{\pi}(x_f,t_f|x_i).\label{eq:no3}
\end{align}

We start by defining the block matrix for the policy dynamics as an absorbing Markov chain. To do this we will \textit{clone} the constraint states $\mc X_{c}\subset \mc X$ and goal states $\mc X_{\dg}\subset \mc X$, where the cloned states (indicated by a tilde) $\tilde{\mc X}_{\dg} = \mc X_{\dg}$ and $\tilde{\mc X}_{c}=\mc X_{c}$ have transitions into them with probability $\mathbbm{1}_{\kappa}(x_i)f_\dg(x_i,\pi(x_i))f_c(x_{\dg},\pi(x_i))$ and $\mathbbm{1}_{\kappa}(x_i)(1-f_c(x_i,\pi(x_i)))+\bar{\mathbbm{1}}_{\kappa}(x_i)$, and where the compliment of those probabilities are multiplied for the normal probability dynamics. The block matrices (which are a function of $\pi$ and $\kappa$),
\begin{align}
    &P_{\mc N +}(\tilde{x}_i|x_i):=\mathbbm{1}_{\kappa}(x_i)f_\dg(x_i,\pi(x_i))f_c(x_{\dg},\pi(x_i)),\\
    &P_{\mc N -}(\tilde{x}_i|x_i):=\mathbbm{1}_{\kappa}(x_i)(1-f_c(x_i,\pi(x_i)))+\bar{\mathbbm{1}}_{\kappa}(x_i),
\end{align}
have transitions from non-terminal to success terminations given by the goal function $f_{\dg}$.
The non-terminal to non-terminal block matrix,
\begin{align}
    P_{\mc N \mc N}(x_j|x_i):=f_2(x_i,\pi(x_i))P(x_j|x_i,\pi(x_i)),\label{eq:NtoNdef}
\end{align}
is the probability of not entering into a cloned success or failure state multiplied by the controlled dynamics. The cloned states are absorbing states for goal achievements and constraint violations that accumulate success or failure probability mass, and allows probability mass to continue flowing through the chain if the goal or constraint violation event does not occur in stochastic scenarios. 

We can segregate these cloned states in a block matrix $B_{\pi}$ for our original policy dynamics:
\begin{align}
    B_{\pi}= \begin{bmatrix}
    P_{\mc N \mc N} & P_{\mc N -} & P_{\mc N +} \\
    0 & I & 0 \\
    0 & 0 & I
\end{bmatrix},
\end{align}
The block matrix $P_{\mc N \mc N}$ are the non-terminal to non-terminal state transitions, $P_{\mc N -}$ are the non-terminal-to-constraint-violation transition dynamics (to the cloned constraint states), $P_{\mc N +}$ are the non-terminal to goal-satisfaction transition dynamics (to the cloned goal states), and absorbing state dynamics are described by identity matrices $I$. If $P_{\pi}(x_j|x_i)=P(x_j|x_i,\pi(x))$ is the original Markovian policy dynamics, then it is straightforward to see that, 
$$P_{\pi}(x_j|x_i) = P_{\mc N\mc N}(x_j|x_i)+P_{\mc N -}(\tilde{x}_j|x_i)+P_{\mc N +}(\tilde{x}_j|x_i).$$ 
By taking the block matrix $B_{\pi}$ to the $t^{th}$ power, we get the cumulative probability of ending up in the cloned terminal states:
\begin{align}
B_{\pi}^t=\begin{bmatrix}
    P_{\mc N \mc N}^t & \sum_{\tau=0}^{t-1}P_{\mc N \mc N}^{\tau}P_{\mc N -} & \sum_{\tau=0}^{t-1}P_{\mc N \mc N}^{\tau}P_{\mc N +} \\
    0 & I & 0 \\
    0 & 0 & I
\end{bmatrix}.\label{eq:B-mat-pow}
\end{align}
 In the top-center and top-right blocks we have a sum of the matrix representations for the STIF and STFF at each time $\tau$. Each matrix multiplication in the sum represents the probability that the agent transitions into the cloned termination state $x_f$ at time $t_f$ when starting from $x_i$ and remaining in the non-terminal state $\mc N$ for $t_f-1$ time-steps:
\begin{align}
    \textcolor{black}{\eta^-_{\pi}(x_f,t_f|x_i)}=\textcolor{teal}{f_2(x_i,a_{\pi})}\textcolor{teal}{\expec_{x'\sim P_{\pi}}}\textcolor{red}{\eta_{\pi}^{-}(x_f,t_f-1|x')}=\textcolor{black}{\Big(}\textcolor{teal}{P_{\mc N \mc N}}\textcolor{red}{P_{\mc N \mc N}^{t_f-1}P_{\mc N -}}\textcolor{black}{\Big)(i,f)}=\textcolor{black}{\left(P_{\mc N \mc N}^{t_f}P_{\mc N -}\right)(i,f)},\label{eq:absorb_to_eta_fail}\\
    \textcolor{black}{\eta^+_{\pi}(x_f,t_f|x_i)}=\textcolor{teal}{f_2(x_i,a_{\pi})}\textcolor{teal}{\expec_{x'\sim P_{\pi}}}\textcolor{blue}{\eta_{\pi}^{+}(x_f,t_f-1|x')}=\textcolor{black}{\Big(}\textcolor{teal}{P_{\mc N \mc N}}\textcolor{blue}{P_{\mc N \mc N}^{t_f-1}P_{\mc N +}}\textcolor{black}{\Big)(i,f)}=\textcolor{black}{\left(P_{\mc N \mc N}^{t_f}P_{\mc N +}\right)(i,f)},\label{eq:absorb_to_eta_suc}
\end{align}



Summing over the column indices in \eqref{eq:B-mat-pow} gives us the time-conditioned cumulative (in)feasibility feasibility functions $\kappa_{\pi}$ which are the probabilities that the agent enters into a cloned success or failure termination state $\tilde{x}_f$ at time $t_f$ when starting from $x_i$:
\begin{align}
    \sum_{f_-}\left(\sum_{\tau=0}^{t_f-1}P_{\mc N \mc N}^{\tau}P_{\mc N -}\right)(i,f_-)\stackrel{\eqref{eq:absorb_to_eta_fail}}{=}\sum_{x_f}\sum_{\tau_f=0}^{t_f}\eta^-_{\pi}(x_f,\tau_f|x_i)=\kappa^-_{\pi}(x_i,t_f),\label{eq:absorb_cum_neg}\\
    \sum_{f_+}\left(\sum_{\tau=0}^{t_f-1}P_{\mc N \mc N}^{\tau}P_{\mc N +}\right)(i,f_+)\stackrel{\eqref{eq:absorb_to_eta_suc}}{=}\sum_{x_f}\sum_{\tau_f=0}^{t_f}\eta^+_{\pi}(x_f,\tau_f|x_i)=\kappa^+_{\pi}(x_i,t_f),\label{eq:absorb_cum_pos}
\end{align}
where $\kappa^+_{\pi}$ is simply a different name for the cumulative feasibility function $\kappa^{**}_{\pi}$ from the OKBEs, but using a $+$ instead of $**$ for notational convenience. Thus we can see the relationship between the time-conditioned $\kappa^+_{\pi}$ and $\eta^+_{\pi}$ from the perspective of an absorbing Markov chain. To get the time-independent $\kappa$ in the OKBEs, we can take $t_f$ to the infinite limit to obtain the following block matrix:
\begin{align}
 B_{\pi}^{\infty} = \lim\limits_{t\rightarrow \infty} B_{\pi}^t \!=\!\lim\limits_{t\rightarrow \infty}\begin{bmatrix}
    P_{\mc N \mc N}^t & \sum_{\tau=0}^{t-1}P_{\mc N \mc N}^{\tau}P_{\mc N -} & \sum_{\tau=0}^{t-1}P_{\mc N \mc N}^{\tau}P_{\mc N +} \\
    0 & I & 0 \\
    0 & 0 & I
\end{bmatrix}
\!=\begin{bmatrix}
    \lim\limits_{t\rightarrow \infty}P_{\mc N \mc N}^t & (I-P_{\mc N \mc N})^{-1}P_{\mc N -} & (I- P_{\mc N \mc N})^{-1}P_{\mc N +} \\
    0 & I & 0 \\
    0 & 0 & I
\end{bmatrix},
\end{align}
where the fundamental matrix (L.H.S.) $(I-P_{\mc N \mc N})^{-1}=\lim\limits_{t \rightarrow \infty}\sum_{\tau=0}^{t}P_{\mc N \mc N}^{\tau}$ is the solution to the Neumann series (R.H.S.), which exists if $I-P_{\mc N \mc N}$ is invertible. The fundamental matrix of an absorbing Markov chain represents the expected state occupancies in $\mc N$ until absorbing into the terminal set $\mc T$ under the policy, over an infinite horizon; its invertibility is important because it means that all possible trajectories under a policy are finite, exit $\mc N$, and contribute to a discrete event. We can obtain the total cumulative (in)feasibility function $\kappa(x)=\lim_{t\rightarrow \infty}\kappa(x,t)$ (which, with a slight abuse of notation, drops the time variable implicitly assuming it to be infinite):
\begin{align}
    \lim_{t_f\rightarrow \infty}\kappa_{\pi}^-(x,t_f)=\lim\limits_{t_f\rightarrow \infty}\sum_{f_-}\left(\sum_{\tau=0}^{t_f-1}P_{\mc N \mc N}^{\tau}P_{\mc N -}\right)(i,f_-)\stackrel{\eqref{eq:absorb_to_eta_fail}}{=}\sum_{x_f}\sum_{t_f=0}^{\infty}\eta^-_{\pi}(x_f,t_f|x_i)=\kappa^-_{\pi}(x_i),\label{eq:absorb_cum_neg_total}\\
    \lim_{t_f\rightarrow \infty}\kappa_{\pi}^+(x,t_f)=\lim\limits_{t_f\rightarrow \infty}\sum_{f_+}\left(\sum_{\tau=0}^{t_f-1}P_{\mc N \mc N}^{\tau}P_{\mc N +}\right)(i,f_+)\stackrel{\eqref{eq:absorb_to_eta_suc}}{=}\sum_{x_f}\sum_{t_f=0}^{\infty}\eta^+_{\pi}(x_f,t_f|x_i)=\kappa^+_{\pi}(x_i),\label{eq:absorb_cum_pos_total}
\end{align}
To prove Eqs. \eqref{eq:no1}, \eqref{eq:no2}, \eqref{eq:no3}, the above limits must exist, that is, the columns of $B_{\pi}^{\infty}$ corresponding to the non-terminal to non-terminal block $\lim\limits_{t\rightarrow\infty}P_{\mc N\mc N}^{t}$ contain no probability mass in the infinite limit, which also makes the Neumann series convergent. In the theory of Markov chains, this occurs if all of the states are \textit{transient}, i.e. a state is transient if there is a non-zero probability of never being re-visited under the dynamics. All states being transient implies that all of the probability mass will drain out of $\mc N$ in the infinite limit. In our case, we know that all states of $\mc N$ must be transient because all infeasible states in the state-space with $\kappa(x)=0$ are part of the failure set $\mc \tm$.  Therefore, the Markov chain is guaranteed to be absorbing. All states being transient implies that $P_{\mc N \mc N}$ has a spectral radius of $\mpol(P_{\mc N \mc N})<1$, which in turn implies that $\lim\limits_{t\rightarrow \infty}P_{\mc N \mc N}^t = 0$. This also implies $I-P_{\mc N \mc N}$ is invertible (and the associated sequence is convergent) because if the spectrum of $P_{\mc N \mc N}$ is $\boldsymbol{\lambda}$ then the spectrum of $I-P_{\mc N \mc N}$ is $\boldsymbol{\nu}=1-\boldsymbol{\lambda}$, which has no zero eigenvalues (given that $0< |\lambda| < 1 $ for all $\lambda\in\boldsymbol{\lambda}$). This proves Eqs. \eqref{eq:no1} and \eqref{eq:no2}.

The rows of $B_{\pi}^{\infty}$ tell us that the cumulative probability of the agent remaining in the non-terminal states $\mc N$ is zero, and all probability mass that doesn't flow into the goal-success terminating states flows into the failure states. For each row $i$, summing over the columns (where $f_{\mc N}$, $f_-$ and $f_+$ are indices for the non-terminal, failure, and success blocks) gives us:
\begin{align}
    &\sum_f B^{\infty}_{\pi}(i,f)=\lim\limits_{t_f\rightarrow \infty}\left[\sum_{f_{\mc N}}P_{\mc N \mc N}^{t_f}(i,f_{\mc N})+\sum_{f_-}\left(\sum_{\tau=0}^{t_f-1}P_{\mc N \mc N}^{\tau}P_{\mc N -}\right)(i,f_-)+\sum_{f_+}\left(\sum_{\tau=0}^{t_f-1}P_{\mc N \mc N}^{\tau}P_{\mc N +}\right)(i,f_+)\right]=1,\\
    &\sum_f B^{\infty}_{\pi}(i,f)=\cancel{\lim\limits_{t_f\rightarrow \infty}\sum_{f_{\mc N}}P_{\mc N \mc N}^{t_f}(i,f_{\mc N})}+\lim\limits_{t_f\rightarrow \infty}\sum_{f_-}\left(\sum_{\tau=0}^{t_f-1}P_{\mc N \mc N}^{\tau}P_{\mc N -}\right)(i,f_-)+\lim\limits_{t_f\rightarrow \infty}\sum_{f_+}\left(\sum_{\tau=0}^{t_f-1}P_{\mc N \mc N}^{\tau}P_{\mc N +}\right)(i,f_+)=1,\\
    \stackrel{\eqref{eq:absorb_cum_neg_total} \eqref{eq:absorb_cum_pos_total}}{\implies}&\sum_{x_f}\sum_{t_f}\eta^-_{\pi}(x_f,t_f|x_i) + \sum_{x_f}\sum_{t_f}\eta^+_{\pi}(x_f,t_f|x_i)=\kappa_{\pi}^-(x_i) + \kappa_{\pi}^+(x_i)=1,\\ 
    &\quad \text{where:}~~\sum_{x_f}\sum_{t_f}\eta^-_{\pi}(x_f,t_f|x_i) = 1-\kappa_{\pi}^+(x_i) = \kappa^-_{\pi}(x_i),\label{eq:compkappa-to-kappa}\\
    \implies &\sum_{x_f}\sum_{t_f}\eta^{**}_{\pi_o}(x_f,t_f|x_i) = 1,\\
    &\quad \text{where:}~~\eta^{**}_{\pi_o}(x_f,t_f|x_i):=\eta^-_{\pi}(x_f,t_f|x_i) + \eta^+_{\pi}(x_f,t_f|x_i).
\end{align}
This concludes the proof of Eq. \eqref{eq:no3} that a policy's STIF and STFF create a state-time option kernel $\eta^{**}_{\pi_o}$ through addition. $\hfill \square$
\\
\subsubsection*{Extra note}
Note that the above derivation reveals that the $\kappa$-OKBE is optimizing the cumulative task-success probability of the absorbing Markov chain, which has the linear algebraic form, $$\kappa(x_i) = \mathbf{e}^T_i(I- P_{\mc N \mc N})^{-1}P_{\mc N +}\mathbf{1},$$ where $\mathbf{e}^T_i$ is a one-hot basis vector and $\mathbf{1}$ is a vector of ones summing over the columns.

\newpage
\section{STOK Factorization Theorem}
We will prove the following theorem and corollary. First we define the class of transition kernels for the proof.

\subsection{Definition of the transition kernel} 
The transition kernel has the factorized form:
\begin{align}
    P(\bos'|\bos,a)= P(z_{k_n}',...,z_{k_1}',x'|z_{k_n},...,z_{k_1},x,a)=\lambda(P_{\bz},\fff,P_x) = \prod_k\sum_{\alpha_k}P_{z_k}(z_k'|z_k,\alpha_{k})F_{k}(\alpha_k|x,a)P_{x}(x'|x,a),
\end{align}
where we will use $\bos = (x,z_{k_1},...,z_{k_n})$.  This is slightly more general that the factorization discussed in Eq. \eqref{eq:joint_kernel} of the main text, because we allow for the fact that affordance functions can condition one HL state-space from another HL space, not just from BL to HL.

\subsection{Theorem and Corollary}
The theorem:
\begin{theorem}[STOK Factorization]\label{appx:STOK-thm}
If $\bar{\mathscr{M}}=\langle\mathscr{Z},\mc X, \mc A_x, P_{\bos},f_{\dg},f_{c,\ell}^{g}\rangle$ where ($f_\dg$,$f_{c,\ell}^{\dg}$) are separable, $P_{\bos} =$ $\lambda(P_{\bz},\fff,P_x)$, $\mc P = \{\pfun^{z_1}_{\alpha_1},...,\pfun^{z_n}_{\alpha_m}\}$, $\eta^{**}_{\pi_{\dg},\ell}$ is the STOK of TMDP $\mathscr{M}_{\dg,\ell}=\langle\mc X,\mc A,P_x,f_{\dg_i},f_{c,\ell}^{g}\rangle$, $(F_x^{\bz},\zeta,f_{c,\ell}^{g})$ is homogeneous, $f_{\dg}$ is not a function of HL states, 
then the product-space STOK $\widetilde{\eta}^{**}_{\pi}$ is:
\begin{align}
     &\widetilde{\eta}^{**}_{\pi_i}\!(\bz_f,x_f,t_f|(\bz,x)^{\ell})=\xi_{\bos,\ell}(t_f|\bz,x)\pfun_{\pi_{\dg}}(x_f|x,t_f)\prod_{\mathclap{k}}\pfun_{k}(z_{f}^k|z^k,t_f)
\end{align}
where $\xi_{\bos,\ell}$ is a fully factorizable TEF defined on $\mc X \times \mathscr{Z}$ using $\eta_{\pi,\ell}$ along with $\eta_{z,\ell}$ STEFs, and where $\xi_{\bos,\ell}(t_f|\bz,x)=\left(\left(1-\prod_{k}\bar\kappa_{s_k}^d(s_{k},t_f)\right) - \left(1-\prod_{k}\bar\kappa_{s_k}^d(s_{k},t_f-1)\right)\right)$, for a set $\{\kappa_{s_k}\}_{n}$ of $n$ compliment CEFs for each space $\mc S_k$ in $\mathbf{S}$.
\end{theorem}
And the corollary:
\begin{corollary}\label{appx:STOK-cor}
    Assuming the same antecedent conditions of Thm. \ref{appx:STOK-thm}, if $\bar{\kappa}_{\mathbf{z},\ell}(\mathbf{z},t_f)= 1$, then the STOK factorization reduces to:
    \begin{align}
        \widetilde{\eta}^{**}_{\pi_i}\!(\bz_f,x_f,t_f|(\bz,x)^{\ell})=\eta^{**}_{\pi_i,\ell}(x_f,t_f|x)\prod_{\mathclap{k}}\pfun_{k}(z_{f}^k|z^k,t_f)
    \end{align}
\end{corollary}
\subsection{Proof Outline}\label{appx:proof_outline}
For the STOK factorization proof we will take the following approach. The proof will show that the equality between the true product-space STOK and the factorized STOK holds for each step $d$ of feasibility iteration, assuming that all functions ($\kappa_{\pi}, \eta_{\pi},\eta_{z},\bar{\kappa}_{z},\bar{\kappa}_{\bz},\xi$) used in the proof are initialized to zero at $d_0$, and each function is progressively constructed over each iteration of dynamic programming. We focus on zero-initializations because, as have discussed in Appx.\ref{sec:fixedpnt}, the Bellman operator for feasibility iteration has a fixed point, but solutions are not necessarily unique. Only solutions with the zero-initializations guarantee that the CFF $\kappa$ accurately reports the true probability of goal satisfaction. Since $\kappa$ is used to define $\eta$ at the $t_0$ boundary condition and we proved in Appx.\ref{appx:eta-kappa} that $\kappa(x) = \sum_{x_f,t_f}\eta(x_f,t_f|x)$, so zero-initializations enforce that $\eta$ will sum to equal $\kappa$ during each step of feasibility iteration.  At a high-level the OKBEs on a full high-dimensional problem would have the following updates during feasibility iteration (where $d$ is the iteration count):
\begin{align}    (\kappa^{d_0}_{\bos},\pi^{d_0}_{\bos},\eta^{d_0}_{\bos})\rightarrow(\kappa^{d_1}_{\bos},\pi^{d_1}_{\bos},\eta^{d_1}_{\bos})\rightarrow(\kappa^{d_2}_{\bos},\pi^{d_2}_{\bos},\eta^{d_2}_{\bos})\rightarrow ... \rightarrow (\kappa_{\bos}^{d_{\infty}},\pi^{d_{\infty}}_{\bos},\eta_{\bos}^{d_{\infty}})
\end{align}
What we will prove is that, instead of this intractable computation, we can compute the following feasibility iteration updates:
\begin{align}    \underbrace{(\kappa_x^{d_0},\pi^{d_0}_{x},\{\bar{\kappa}^{d_0}\}_{n})}_{\text{Sec.}\ref{appx:d0}}\rightarrow\underbrace{(\kappa_x^{d_1},\pi^{d_1}_{x},\{\bar{\kappa}^{d_1}\}_{n})}_{\text{Sec.}\ref{appx:d1}}\rightarrow \underbrace{(\kappa^{d_2}_x,\pi^{d_2}_x,\{\bar{\kappa}^{d_2}\}_{n})}_{\text{Sec.}\ref{appx:d2}}\rightarrow ... \rightarrow \underbrace{(\kappa_x^{d_{\infty}},\pi^{d_{\infty}}_x,\{\bar{\kappa}^{d_{\infty}}\}_{n})}_{\text{Sec.}\ref{appx:dn}}
\end{align}
where $\{\bar{\kappa}^{d_{\infty}}\}_{n}$ is a set of $n$ compliment CEF functions for each state-space and $\pi_x$ is a policy computed only on $\mc X$ using $f_{1,x}^{\ell}$ and $f_{2,x}^{\ell}$. Thus, every function we compute during feasibility iteration will be defined on one of the $n$ state-spaces comprising $\mathscr{S}$. Using the components $(\kappa^{d_{\infty}},\pi^{d_{\infty}}_x,\{\bar{\kappa}^{d_{\infty}}\}_{n})$ along with a set of SPKs $\{\rho^{d_{\infty}}\}_{n}$ (that can be computed independently of feasibility iteration) \textbf{we prove that we can perfectly construct $\eta_{\bos}^{d}$ for all steps $d$ of feasibility iteration with the factorization}:
\begin{align}
    \eta_{\bos,\pi}^d(\bz_f,x_f,t_f|\bz,x)&=\pfun_{x,\pi}^{\ell}(x_f|x,t_f)\pfun_{\bz}^{\ell}(\bz_f|\bz,t_f)\xi_{\bos,\pi}^d(t_f|\bz,x)\\
    &=\pfun_{x,\pi}^{\ell}(x_f|x,t_f)\Big(\prod_{k}\pfun_{z_k}^{\ell}(z_{k,f}|z_{k},t_f)\Big)\Big(\Big(1-\prod_{k}\bar\kappa_{s_k}^d(s_{k},t_f)\Big) - \Big(1-\prod_{k}\bar\kappa_{s_k}^d(s_{k},t_f-1)\Big)\Big)
\end{align}
Furthermore, STOKs have STIF and STFF components, $\eta^+$ and $\eta^-$, each with the same definition for $t_f>t_0$ but slightly different definitions at $t_0$. We will not prove the result for each of these functions because it would be nearly identical proofs. Since we already showed in Appx.\ref{appx:eta-kappa} that $\eta_{\pi}^{**}(x_f,t_f|x)=\eta_{\pi}^{+}(x_f,t_f|x)+\eta_{\pi}^{-}(x_f,t_f|x)$, we will instead combine both of the success and failure event probabilities into one definition of the STOK.


Before the main part of the STOK factorization proof, we will first define regions and sub-regions, and then we will define random variables which indicate the occupancy of an agent being in a region and random variables for a history of an absence of events (which combines events for violating region occupancies, goal-satisfaction, constraint-violation). These random variables will play a key role in the proof by enabling a region-conditioned conditional independence between the state-dynamics of many systems given time.

\textbf{Part 1:} (sec. \ref{part1}) We begin by defining random variables which will help us decompose our problem.
\begin{enumerate}[itemsep=0pt, parsep=0.5pt]
    \item Region Random Variables (sec. 
    \ref{aapx:regions} and sec. \ref{sec:RVs}): Define region RVs that indicate whether an agent exists in a region or not.
    \item Goal and Constraint Random variables (sec. \ref{aapx:goals_constraints}): Define RVs that indicate if goal-events or constraint-violation events have occurred
    \item History Random Variables (sec. \ref{aapx:histroy}): Define an RV that indicates whether a goal, constraint-violation, or region-violation event has occurred over a time-span.
    \item First-event Random Variable (sec. \ref{aapx:firstevent}): Define an RV for the first time an event occurs at a given time-step.
\end{enumerate}

 \textbf{Part 2:} (sec. \ref{part2}) Given these random variables, we can break down the Bellman operator on a high-dimensional STOK and show how it decomposes into components $\rho$, $\eta$ and $\bar{\kappa}$ over every step $d$ of feasibility iteration. Specifically there will be subsections where:

 \begin{enumerate}[itemsep=0pt, parsep=0.5pt]
    \item We show how the STOK factorization holds at step $d_1$ in sec. \ref{appx:d1}.
    \item We show how the STOK factorization holds at step $d_2$ in sec \ref{appx:d2}.
    \item By induction, we then show how the STOK factorization holds at step $d+1$ in sec \ref{appx:dn}.
\end{enumerate}

Note that for Part 2, every STOK factorization will also include a policy that is only computed on the BL state-space, and thus each of these steps will show that we can substitute this BL policy in for the policy on the full product-space.  By proving that the STOK factorization holds for every step of feasibility iteration, it holds when the CFF converges (which we proved in sec. \ref{sec:fixedpnt}), and this will conclude the proof.

\subsubsection{Notation}
We will use $\mc S$ as generic state-spaces in the full product space $\mathscr{S} = \prod_{i}\mc S_i$. $\bos = (s_1,....,s_n)$ will be a generic state-vector, where $s_j$ is a component corresponding to $\mc S_j$. The set of all state-spaces is $\mathbf{S} = \{\mc S_1,...,\mc S_{n}\}= \{\mc X \times \mc Z_1,...,\mc Z_{n-1}\}$. The vector $\bos = (\bz,x)$ and both will often be used in equations interchangeably. $\mc Z$ will be used specifically for high-level state-spaces and $\mc X$ for low-level state-spaces (with states $z \in \mc Z$, $x\in \mc X$), where $\bos = (x,z_1,....,z_n)$ is another way of writing the state-vectors. $\mathscr{Z} = \prod_k \mc Z_k$ is the Cartesian product of high-level state-spaces and $\mathscr{S} = \mathscr{Z} \times \mc X$. For random variables, we will often mark their realizations using a superscript on the upper-left part of the variable.  For example, $^{0\!}R_j^{\ell}$ and $^{1\!}R_j^{\ell}$ means the Bernoulli R.V. $R_j^{\ell}$ evaluates to $0$ and $1$ respectively.  We will also use the notation $\bos\boa = (s^0,...,s^n,a,\alpha^1,...,\alpha^{n-1})$ for state-action vectors, where $s^0 = x$ is the base-level state, $s^k = z^k$ for $k>0$ is a high-level state, and $\boa = (a,\alpha^1,...,\alpha^{n-1})$ will be used as a generic action vector over both base-level and high-level actions. It is also worth noting that an action vector $\boa$ is fully determined by a BL state-action $(x,a)$ and the affordance function $F(\ba|x,a)$ where $a$ is the only free variable. Thus, in various parts of the proof we will simply write $\boa$ without explicitly writing out the affordance function because it will take up too much space in the equations.  The variable $sa=(s,a)$ is used for compact notation, where $\bos\boa^j\equiv\bos\boa(j)\equiv sa^j$ are all equivalent.
\subsubsection{Assumptions}\label{assumptions}
In this proof, without loss of generality, we will not concern ourselves with environment modes $e$ because they are equivalent to HL actions $\alpha$ and a mode-functions $\zeta(e|x)$ is equivalent to a deterministic affordance function. Including modes and mode functions does not change the result.  Also, many equations require an implicit deterministic time transition kernel $T(t+1|t)=1$ which is almost always left out for compactness. Affordance functions $F$ will also be factorized with factors $\{F_1,...,F_n\}$, $F(\ba|\bos,\boa) = \prod_k F_k(\ba^j|\bos^k,\boa^k)$, and we will assume that these functions are deterministic. If $F$ is stochastic, the proof has nearly identical steps, but instead of regions of default dynamics being conditioned on one of many region-specific variables $\ba_{\ell}$, it would be conditioned on one of many region-specific \textit{distributions} $\bod_{\bos a}^{\ell}:\mc A_{\ba}\rightarrow [0,1]$ and $F$ would be redefined $F:\mc S \times \mc A \rightarrow \Delta(\mc A_{\ba})$, where $\Delta(\mc A_{\ba})$ is a probability simplex over $\mc A_{\ba}$ (the set of all action vectors) where $\bod \in \Delta(\mc A_{\ba})$, i.e. $P_{\bz,\bod_{\bos a}^{\ell}}(\bz'|\bz) = \sum_{\ba_i}P_{\bz}(\bz'|\bz,\ba_i)\bod_{\bos a}^{\ell}(\ba_i)$. Thus, we avoid stochasticity in $F$ for simpler notation and fewer summations.

\subsection{Part 1: Definitions of Random Variables and Functions}\label{part1}
In this subsection we define random variables (R.V.) for events. In this proof, we concern ourselves with computing the probability of a first-event across the full product space of dynamics and there will be multiple event types. We will use the convention that a realization of $0$ in a Bernoulli R.V. will correspond to a notable event in the control setting, because the conjunction of multiple events will be an event, and thus, naturally multiplying $0$ realizations produces another $0$. The absence of a salient event will be a realization of $1$.
\subsubsection{Regions}\label{aapx:regions}
A region $\mc R^{\ell}$ is defined as the set of state-action vectors which output the same distribution over high-level actions $\alpha$,
\begin{align}
    \mc R^{\ell} = \{(\bos,a) : \fff(\ba^{\ell}|\bos,a)=1\},
\end{align}
where the superscript $\ell$ is an index of a unique output $\ba_{\ell}$ from $\fff$. Given that $F$ is a conditional distribution, every $(\bos,a)$ in $\mc S \times \mc A$ exists in a unique region such that the set of all regions $\mathscr{R}$ partition the state-vector-action space: $\mc S \times \mc A = \bigcup_{\ell}\mc R_{\ell}, \forall\mc R_{\ell}\in\mathscr{R}$.

A \textit{sub-region} $\mc R_j^{\ell}$ is simply the projection of all region vectors onto the state-action space $\mc S_j \times \mc A_{j}$:
\begin{align}
    \mc R_j^{\ell} = \text{proj}_{\mc S_j \times \mc A_j}(\mc R^{\ell}).
\end{align}

\subsubsection{Region Random Variables}\label{sec:RVs}

Define $\mathbf{d}: \mc S\rightarrow [0,1], ~s.t. \sum_{i}\mathbf{d}(i)=1$, as a PMF over the Cartesian product space $\mathscr{S}=\mc X\times \mc Z_1 \times ... \times \mc Z_n$. Also define the $\mathbf{d}_j:\mc S_j\rightarrow [0,1],~s.t. \sum_i\mathbf{d}_j(i)=1,$ over the individual state-space $\mc S_j \in \mathbf{S}$ as the marginal of $\mathbf{d}$, marginalizing over all state-spaces except $\mc S_j$.

A \textit{sub-region} R.V. $R_{j}^{\ell}$ on a single state-action space $\mc S \times \mc A_{j}$ is defined:
\begin{align}
    R_j^{\ell}(sa)=\begin{cases} 
1, & \text{if } (s,a) \in \mc R_{j}^{\ell}, \text{ (in region)}\\
0, & \text{if } (s,a) \not\in \mc R_{j}^{\ell}, \text{ (not in region)}.
    \end{cases}
\end{align}
with PMF $p_{j|\mathbf{d}_j}$ defined by a probability distribution $\mathbf{d}_j:\mc S_k\rightarrow [0,1],~s.t. \sum_i\mathbf{d}_j(i)=1,$ over the individual state-space $\mc S_j \in \mathbf{S}$, we have the probability of existing in that sub-region $\mc R_j^{\ell}$ (this is also a function of the policy $\pi$ which will be suppressed in the notation):
\begin{align}
    p_{j|\mathbf{d}}(R_{j,sa}^{\ell} = 1) = \mathbf{d}_j(s)\pi(a|s),\quad p_{j|\mathbf{d}}(R_{j,sa}^{\ell} = 0) = 1-\mathbf{d}_j(s)\pi(a|s).
\end{align}
The region Bernoulli R.V. on the full product-space $\mathscr{S}$ is $\mathfrak{R}$, defined as:
\begin{align}
    \mathfrak{R}^{\ell}(\bos\boa)=\prod_{j}R_j^{\ell}(\bos\boa^{j}).
\end{align}
where $\bos\boa^{j}$ is the $j^{th}$ component of $\bos\boa = (sa^0,...,sa^j,...,sa^n)$. This means, $\mathfrak{R}(\bos\boa)= 1$ if there is no sub-region violation (all components of the state vector are in their corresponding sub-region $\mc R_j^{\ell}$) and $\mathfrak{R}^{\ell}(\bos\boa)= 0$ if there is at least one sub-region. 

The probability that $\bos\boa$ is in the region $\mc R^{\ell}$ is:
\begin{align}
    p_{\mathfrak{R}}(\mathfrak{R}^{\ell}_{\bos\boa} = 1) = \prod_{j=0}^n p_j(R_{j,sa}^{\ell} = 1),\quad
    p_{\mathfrak{R}}(\mathfrak{R}^{\ell}_{\bos\boa} = 0) = 1-\prod_{j=0}^n p_j(R_{j,sa}^{\ell}= 1).
\end{align}
Thus, we have $p_{\mathfrak{R}|\bod}(\mathfrak{R}^{\ell}_{\bos\boa} = 0)+p_{\mathfrak{R}|\bod} (\mathfrak{R}^{\ell}_{\bos\boa} = 1) = 1$.

\subsubsection{Goal, Constraint, and Infeasibility Random Variables} \label{aapx:goals_constraints}
In addition to region random variables, we introduce goal success and constraint violation Bernoulli R.V.s for a sub-space $\mc S_j\times \mc A_j$: $G_j(\bos\boa) = \{0 ~~\text{if: } (sa^j \in \mc G_j, ~~1~~ \textbf{o.w.}\},$ and $C_j(\bos\boa) = \{0 ~~\text{if: } sa^j \in \mc C_{j}, ~~1~~ \textbf{o.w.}\}$, where $\mc G_j$ and $\mc C_j$ are goal and constraint sets specific to space $\mc S_j \times \mc A_j$. Notice that these R.V.s encode goal-achievement and constraint-violation "events" as zeros so that the and together when multiplied.

Over the entire space $\mathscr{S} \times \mathscr{A} = (\mc S_1 \times ... \times \mc S_n) \times (\mc A_x \times ... \times \mc A_n)$, the product-space goal and constraint RVs are defined as:
$$\mathfrak G(\bos\boa) = \{1 ~~\text{if: } \prod_{j}G_j(\bos\boa(j))=1, ~~0~~ \textbf{o.w.}\}, \quad
    \mathfrak C(\bos\boa) = \{1 ~~\text{if: } \prod_{j}C_j(\bos\boa(j)), ~~0~~ \textbf{o.w.}\}.$$
This R.V. has Bernoulli probabilities which form the achievement and constraint functions for individual spaces (which comprise the separable goal and constraint functions over the entire product-space):

\subsubsection{History Random Variable}\label{aapx:histroy}
The three goal-success, constraint violation, and region-violation events are captured in the STOK. We can now construct a history R.V. $\mathfrak{H}_t(\bos_{0:t})$ which indicates whether or not there has been one of these events in the trajectory $\bos_{0:t} = (\bos_{0},\bos_{1},...,\bos_{t})$,
\begin{align}
    \mathfrak{H}_t(\bos_{0:t}) = \begin{cases} 
1, & \text{if } \prod_{\tau=0}^t \mathfrak{R}_\tau(\bos\boa_{0:t})\mathfrak G_\tau(\bos\boa_{0:t})\mathfrak C_\tau(\bos\boa_{0:t})=1, \text{ (all 1s)}, \\
0, & \text{if } \prod_{\tau=0}^t \mathfrak{R}_\tau(\bos\boa_{0:t})\mathfrak G_\tau(\bos\boa_{0:t})\mathfrak C_\tau(\bos\boa_{0:t})=0, \text{ (at least one 0)},
\end{cases}
\end{align}
where, $p_{\mathfrak{H}}(\mathfrak{H}_{t_f}^{1:n} = 0) = \prod_{t=t_0}^{t_f}p_{\mathfrak{G}}(\mathfrak{G}_t = 0)p_{\mathfrak{C}}(\mathfrak{C}_t = 0)p_{\mathfrak{R}^{\ell}}(\mathfrak{R}_t^{\ell} = 0),
    p_{\mathfrak{H}}(\mathfrak{H}_{t_f}^{1:n} = 1) = 1-\prod_{t=t_0}^{t_f}p_{\mathfrak{G}}(\mathfrak{G}_t = 0)p_{\mathfrak{C}}(\mathfrak{C}_t = 0)p_{\mathfrak{R}^{\ell}}(\mathfrak{R}_t^{\ell} = 0).$

And the probability of the R.V. is,
\begin{align}
    p_{\mathfrak{H}^{\ell}}(\mathfrak{H}_{t_0}=1) = p_{\mathfrak{R}^{\ell}|\bod}(\mathfrak{R}_{t_0}^{\ell}(\bos\boa))p_{\mathfrak{R}|\bod}(\mathfrak{R}_{t_0}^{\ell}(\bos\boa)).
\end{align}

\subsubsection{First Event Random Variable}\label{aapx:firstevent}
Lastly, we will introduce a R.V. which incorporates all events into one variable (goal success, constraint violation, region violation). Let the first-event R.V. $\mathfrak{E}^{\ell}$ be defined as:
\begin{align}
    \mathfrak{E}^{\ell}_{t_f}(\bos\boa_{t_0:t_f}) = \mathfrak{H}_{t_f-1}(\bos\boa_{t_0:t_f-1})\big(1-\mathfrak{R}_{t_f}^{\ell}(\bos\boa_{t_f})\mathfrak G_{t_f}(\bos\boa_{t_f})\mathfrak C_{t_f}(\bos\boa_{t_f})\big) \label{appx:event-and-history}
\end{align}
where $\mathfrak{E}^{\ell}_t(\bos\boa) = 1$ if there is a region violation for the first time at $t$ and $\mathfrak{E}^{\ell}_t(\bos\boa) = 0$ if there is no region violation for the first time at $t$. 

Equivalently, the first event R.V. is defined:
\begin{align}
    \mathfrak{E}^{\ell}_{t_f}(\bos\boa_{t_0:t_f}) = \mathfrak{E}_{t_f:t_1}(\bos\boa_{t_1:t_f})\big(1-\mathfrak{R}_{t_0}^{\ell}(\bos\boa_{t_0})\mathfrak G_{t_0}(\bos\boa_{t_0})\mathfrak C_{t_0}(\bos\boa_{t_0})\big) \label{appx:event-and-history-2}
\end{align}
This is the more important definition, which gets rid of the history R.V. and just muliplies on the probability of and event to the first time $t_0$.
The probability of $\mathfrak{E}^{\ell}_t$ taking on the Boolean value of $1$ is given as:
\begin{align}
     p_{\mathfrak{E}}(\mathfrak{E}^{\ell}_{t_f} = 1) &= p_{\mathfrak{H}}(\mathfrak{H}_{t_f-1} = 0) \big(1-p_{\mathfrak{G}}(\mathfrak{G}_{t_f} = 0)p_{\mathfrak{C}}(\mathfrak{C}_{t_f} = 0)p_{\mathfrak{R}}(\mathfrak{R}_{t_f} = 0)\big), 
\end{align}
or equivalently, and more usefully, without the history R.V.:
\begin{align}
     p_{\mathfrak{E}}(\mathfrak{E}^{\ell}_{t_f} = 1) &= p_{\mathfrak{E}}(\mathfrak{E}_{t_f:t_1} = 1) \big(1-p_{\mathfrak{G}}(\mathfrak{G}_{t_0} = 0)p_{\mathfrak{C}}(\mathfrak{C}_{t_0} = 0)p_{\mathfrak{R}}(\mathfrak{R}_{t_0} = 0)\big), \\
    p_{\mathfrak{E}}(\mathfrak{E}^{\ell}_{t_f} = 0) &= 1-p_{\mathfrak{E}}(\mathfrak{E}^{\ell}_{t_f} = 1)
\end{align}

As we discussed in the proof outline (\ref{appx:proof_outline}), the achievement function will include any event, not just the goal event, and the completion function will define the probability that there is no event, because our proof will be for the entire STOK which combines the STFF and STIF definition at $t_0$.

\subsubsection{Achievement and Continuation Functions}

For the TMDP, the defined Bernoulli distributions corresponding to the goal-completion, constraint-violation, and region-occupation RVs are used to define the separable achievement and continuation functions $f_1$ and $f_2$. As we mentioned in the proof outline section (sec. \ref{appx:proof_outline}), we group together goal satisfaction, constraint-violation, and region-exiting events together into one function $f_1$, so the achievement function will be $f_1(\bos,\boa) = 1-f_2(\bos,\boa)$. The region-exiting function for region $\mc R_\ell$ on state-action space $\mc S_k \times \mc A_k$ will be defined:
\begin{align}
    f_{\ell,k}(s^k,a^k) = 1-F(\alpha_{\ell}|s^k,a^k)=1-p_R(R_{k,sa}^{\ell}),
\end{align}
which is the probability that the agent is not inducing the region-action $\alpha_{\ell}$ from $(s,a)$. Also, recall that the absence of events are encoded as a realization of $1$ where events are encoded as $0$, thus, both functions are defined as:
\begin{align}
    f_2(\bos,\boa) &= p_{\mathfrak{G}}(\mathfrak{G} = 1)p_{\mathfrak{C}}(\mathfrak{C} = 1)p_{\mathfrak{R}}(\mathfrak{R}^{\ell} = 1),~~ \text{Prob. of no events, $\neg$ Goal-success $\land$ $\neg$ Constraint-violation $\land$ $\neg$ Region-violation},\\
    f_1(\bos,\boa)&=1-p_{\mathfrak{G}}(\mathfrak{G} = 1)p_{\mathfrak{C}}(\mathfrak{C} = 1)p_{\mathfrak{R}}(\mathfrak{R}^{\ell} = 1), ~~ \text{Prob. of any event. Goal-success $\lor$ Constraint-violation $\lor$ Region-violation},\\
    f_2(\bos,\boa) &= \bar{f}_{\dg}(x,a) \prod_k f_{c\ell k}(s_k,\boa^k),\\
    f_1(\bos,\boa) &= 1-f_2(\bos,\boa) = 1-\bar{f}_{\dg}(x,a) \prod_k f_{c\ell k}(s_k,\boa^k),
\end{align}
with $\boa^k = \boa(k)$ and where we use the defined function:
\begin{align}
    f_{c\ell k}(s_k,a_k) &:= f_{c,k}(s_k,a_k)f_{\ell,k}(s_k,a_k)\\
    \bar{f}_{\dg}(x,a) &:= 1-f_{\dg}(x,a)
\end{align}
and,
\begin{align}
    f_{1,k}(s_k,a_k) &= 1-p_{G}(G_k = 1)p_{C}(C_k = 1)p_{R}(R_k = 1),\\
    &= 1- f_{\dg}(s_k,a_k))f_{c\ell k}(s_k,a_k)\\
    f_{2,k}(s_k,a_k) &= p_{G}(G_k = 1)p_{C}(C_k = 1)p_{R}(R_k = 1),\\
    &= f_{\dg}(s_k,a_k)f_{c\ell k}(s_k,a_k)
\end{align}
is the product of continuation functions which indicate the absence of an event on each space $\mc S_k \times \mc A_k$.

\subsubsection{An important identity for the compliment CEF}
Before continuing with the STOK decomposition, we provide an important identity that will be used in the proof. We write a time-augmented Option Kernel Bellman Equation  as,
\begin{align}
    {\kappa}_{\pi}(s,t_f)&=
f_1(s,a)+f_2(s,a)\expec_{s'\sim P} {\kappa}_{\pi}(s',t_f-1),\\
&=(1-f_2(s,a))+f_2(s,a)\expec_{s'\sim P} {\kappa}_{\pi}(s',t_f-1),
\end{align}
which reduces to the standard OKBE if $t_f$ is summed over on both sides. Now we derive the recursive form for the compliment CFF. Using Eqs. \eqref{eq:absorb_cum_pos} and \eqref{eq:compkappa-to-kappa}, we have:
\begin{align}
    1-\bar{\kappa}_{\pi}(s,t_f)&=
(1-f_2(s,a))+f_2(s,a)\expec_{s'\sim P}(1-\bar{\kappa}_{\pi}(s',t_f-1)),\\
    1-\bar{\kappa}_{\pi}(s,t_f)&=
(1-f_2(s,a))+f_2(s,a)-f_2(s,a)\expec_{s'\sim P}\bar{\kappa}_{\pi}(s',t_f-1),\\
1-\bar{\kappa}_{\pi}(s,t_f)&=
1-f_2(s,a)\expec_{s'\sim P}\bar{\kappa}_{\pi}(s',t_f-1),\\
\bar{\kappa}_{\pi}(s,t_f)&=
f_2(s,a)\expec_{s'\sim P}\bar{\kappa}_{\pi}(s',t_f-1).\label{eq:compkappa_identity}
\end{align}
where $\kappa_{\pi}(s,-1)=0$ and $\bar{\kappa}_{\pi}(s,-1)=1$.  

\subsection{Part 2: The STOK Factorization holds over each step of Feasibility Iteration}\label{part2}
Here we being the main part of the proof where we define the Bellman Operators which will update the functions during feasibility iteration, and then we show how the STOK factorization will hold for each step of feasibility iteration until convergence if we initialize our functions to zero.

\subsubsection*{Definition for the $\kappa$ Bellman Operator}
We now define a Bellman operator $\mc B_{\kappa}$ acting on $\kappa$ for updates $\kappa^{d+1}\leftarrow \mc B\kappa^d$ in two equivalent ways. The first using probability notation, the second using achievement and continuation functions:
\begin{align}
    \kappa^{d+1}=~&\mc B\kappa^d = \max_a\left[(1-p(\mathfrak{R}_{\bos\boa}=1)p(\mathfrak{G}_{\bos\boa}=1)p(\mathfrak{C}_{\bos\boa}=1))+p(\mathfrak{R}_{\bos\boa}=1)p(\mathfrak{G}_{\bos\boa}=1)p(\mathfrak{C}_{\bos\boa}=1)\sum_{\bos'}P(\bos'|\bos,a)\kappa^d(\bos')\right],\\
    \kappa^{d+1}=~&\mc B\kappa^d = \max_a \left[f_1(\bos,a)+f_2(\bos,a)\Big(\sum_{x'}P(x'|x,a)\prod_k \sum_{\alpha_k}\Big(P(z_k'|z_k,\alpha_{k})F_k(\alpha_k|x,a)\Big)\Big)\kappa^d(\bz',x')\right],\label{appx:bellman-op}
\end{align}

\subsubsection*{Definition for the $\pi$ Bellman Operator}
We now define a Bellman operator $\mc B_{\pi}$ acting on $\pi$ for updates $\pi^{d+1}\leftarrow \mc B\pi^d$ in two equivalent ways. The first using probability notation, the second using achievement and continuation functions:
\begin{align}
    \pi^{d+1}=~&\mc B\pi^d = \argmin_{a\in\mc A_x^{*}} \left[f_2(\bos,a)\sum_{x'}P(x'|x,a)\prod_k \sum_{\alpha_k}P(z_k'|z_k,\alpha_{k})F_k(\alpha_k|x,a)\sum_{z_{k,f},t_f}\big(t_f+1\big)\eta^d(\bz_f,x_f,t_f|z_k',x')\right],\label{appx:pi-bellman-op}
\end{align}

\subsubsection*{Definition for the $\eta$ Bellman Operator}
We define a Bellman operator $\mc B_{\eta}$ acting on $\eta$ for updates $\eta^{d+1}_{\pi}\leftarrow \mc B\eta^d_{\pi}$ in two equivalent ways. The first using probability notation, and the second using achievement and continuation functions:
\begin{align}
    \eta^{d+1}_{\pi}&=\mc B\eta^d_{\pi} = \mathbbm{1}_{t_0}(t_f)(1-p(\mathfrak{R}_{\bos\boa}=1)p(\mathfrak{G}_{\bos\boa}=1)p(\mathfrak{C}_{\bos\boa}=1))\delta_{if}+\bar{\mathbbm{1}}_{t_0}(t)p(\mathfrak{R}_{\bos\boa}=1)p(\mathfrak{G}_{\bos\boa}=1)p(\mathfrak{C}_{\bos\boa}=1)\expec_{\bos'\sim P_{\bos}^{\pi}}\eta^d_{\pi}(\bos_f,t_f-1,\mathfrak{E}_{t_f-1}|\bos'),\\
    \eta^{d+1}_{\pi}&=\mc B\eta^d_{\pi} = \mathbbm{1}_{t_0}(t_f)f_1(\bos,a)\delta_{if}+\bar{\mathbbm{1}}_{t_0}(t)f_2(\bos,\pi(\bos))\sum_{\bos'}P(\bos'|\bos,\pi(\bos))\eta^d_{\pi}(\bos_f,t_f-1,\mathfrak{E}_{t_f-1}|\bos'),\label{appx:eta-bellman-op}
\end{align}
where $\mathbbm{1}_{t_0}(t_f)$ is an indicator function for $t_0$ and $\bar{\mathbbm{1}}_{t_0}(t_f)=1-\mathbbm{1}_{t_0}(t_f)$. We will mostly focus on the updates for $t_f>0$ and thus leave out the $\mathbbm{1}_{t_0}$ function in many of the equations, unless we are directly addressing the $t_0$ time-step. 

\subsubsection{Feasibility Iteration Initialization at step $d_0$}\label{appx:d0}
All functions that we perform feasibility iteration on will be initialized to zero at step $d_0$: $\kappa_{\bos}^{d_0}(\bz,x) = 0, 
    \kappa_{\pi}^{d_0}(x) = 0, 
    \kappa_{z_k}^{d_0}(z_k,t_f) = 0,~ \forall k, 
    \pi_{x}^{d_0}(x) = a_0, 
    \pi_{\bos}^{d_0}(\bos) = \boa_0, 
    \eta_{\pi}^{d_0}(x_f,t_f|x) = 0, 
    \eta_{\bos}^{d_0} (\bz_f,x_f,t_f|\bz,x) = 0.$ where $a_0$ and $\boa_0$ are arbitrary initial actions. Trivially, the STOK factorization holds using the functions because everything is zero.


\subsubsection{Feasibility Iteration step $d_1$}\label{appx:d1}
\subsubsection*{Initial conditions for  $\kappa$ at $d_1$}
Let $\bar{f}_{\dg}(s,a) = 1-f_{\dg}(s,a)$ where $f_{\dg}(s,a)=1$ if a goal is satisfied and $0$ if not. Let $\bar{f}_{c}(s,a) = 1-f_{c}(s,a)$ where $f_{c}(s,a)=0$ if a constraint is not violated and $1$ if it is. Thus $\bar{f}_{\dg}(s,a)=0$ means a goal is satisfied and $1$ mean it is not.
\begin{align}
    f_{1}(\bos,a)  
    &=f_{\dg}(x,a)\prod_k f_{c\ell k}(z_k,\alpha^k),\\
    f_{2}(\bos,a) 
    &=\bar{f}_{\dg}(x,a)\prod_kf_{c\ell k}(z_k,\alpha^{k}),
\end{align}
For the $\kappa$-OKBE equation, $\kappa^{d_0}(\bos)=0$ cancels out of the equation, so $\kappa^{d_1}$ is:
\begin{align}
    \kappa_{\bos}^{d_1}(\bos) &= \max_{a}\left[f_1(\bos,a)+\cancel{f_2(\bos,a)\expec_{\bos' \sim P_{\bos}}\kappa_{\bos}^{d_0}(\bos') } \right]\\
    &=\max_{a}f_1(\bos,a)=\max_{a}f_{\dg,x}(x,a)f_{c\ell x}(x,a)f_{c\ell \bz}(\bz,\ba),\\
    &=f_{c\ell \bz}(\bz,\ba)\max_{a}\left[f_{\dg,x}(x,a)f_{c \ell x}(x,a)\right]\\
    &=f_{c\ell \bz}(\bz,\ba)\kappa_{x}^{d_1}(x_\bos)\label{appx:smaller}
\end{align}
and the optimal action set is therefore:
\begin{align}
    \mc A_{\bos}^{*}(\bos) = \argmax_{a}\left[\cancel{f_{c\ell \bz}(\bz,\ba)}f_{\dg,x}(x,a)f_{c \ell x}(x,a)\right].
\end{align}
where, similarly $\kappa_{x}^{d_1}$ in Eq. \eqref{appx:smaller} is given as,
\begin{align}
    \kappa_{x}^{d_1}(x_\bos) = \max_{a}\left[f_{\dg,x}(x,a)f_{c \ell x}(x,a)+\cancel{f_{2,x}(x,a)\expec_{x' \sim P_{x}}\kappa_{x}^{d_0}(x')} \right],
\end{align}
Notice that the action sets $\mc A^*$ for this OKBE is equivalent to the action set of a reduced $\kappa$-OKBE on $\mc X$:
\begin{align}
    \mc A_x^* &= \argmax_{a}\left[f_{\dg,x}(x,a)f_{c \ell x}(x,a)+\cancel{f_{2,x}(x,a)\expec_{x' \sim P_x}\kappa_{\bos}^{d_0}(x')}  \right]
\end{align}
Thus, $\mc A^*_\bos=\mc A^*_x$ for step $d_1$.  This will matter when we define the policy update for $d_1$ next, because it will allow us to replace the product-space policy with the reduced policy.

\subsubsection*{Updating $\pi$ for step $d_1$}
The policy for step $d_1$ is simple because $\eta^{d_0}$ evaluates to $0$ for all $t_f$:
\begin{align}
    \pi^{d_1}_{\bos}(\bos) &= \argmin_{a\in \mc A^*_{\bos}}\left[f_2(\bos,\boa)\expec_{\bos'\sim P_{\bos}}\sum_{\bos_f}\sum_{t_f}(t_f+1)\eta^{d_0}(\bos_f,t_f|\bos')\right]= \argmin_{a\in \mc A^*_{\bos}} 0
\end{align}
The same is true for a reduced policy optimization only on the BL:
\begin{align}
    \pi^{d_1}_x(x_{\bos}) = \argmin_{a\in \mc A^*_x}\left[f_2(x_{\bos},a)\expec_{x'\sim P_{x}}\sum_{x_f}\sum_{t_f}(t_f+1)\eta^{d_0}(x_f,t_f|x')\right]= \argmin_{a\in \mc A^*_x} 0
\end{align}
which implies that at $d_1$, so long as we have a systematic way of choosing an action from the equivalent sets $\mc A^*_x = \mc A^*_{\bos}$, we obtain equivalent policies. By equivalent we mean that, even through $\pi^d_{\bos}$ is defined on $\mathscr{S}$, and $\pi_x^d(x)$ is defined on $\mc X$, it is the case that the policies output the same actions when we take the projection of $\bos$ onto $\mc X$, i.e. $x_{\bos}$: $\pi^d_{\bos}(\bos) = \pi^d_{x}(x_\bos)$. A systemically way of choosing an action is straightforward: we can simply rank the priorities of actions in $\mc A^*_x$ and $\mc A^*_{\bos}$ with arbitrary indices so that in the case that there are equivalent actions that minimize time-to-go, we choose the same action from both sets.

\subsubsection*{Initial factorization of $\eta_{\pi,\bos}$ on step $d_1$ and $t_0$}\label{sec:t0-conditions}
We start by defining the STOK at time $t_0$. The STOK can be broken down via the chain rule and conditioning variables can be dropped because $f_1$ and $f_2$ are separable functions. 

\subsubsection*{Factorizing the STOK} 

We apply the chain rule to $\eta$ to obtain (for $t_f >0$):
\begin{align}
     \eta_{\pi}^{d_1}(\bz_f,x_f,t_f,\mathfrak{E}^{\ell}_{t_f}|(\bz,x)_{\ell})&=f_2(\bz,\ba_{\ell})f_2(x,a)\expec_{\bos'\sim P_{\bos}}\Bigg(\widetilde{\pfun}_{\bz}^{\ell, d_1}(\bz_f|\bz',t_f-1,x_f,x',\mathfrak{E}^{\ell}_{t_f})\widetilde{\pfun}_{\pi}^{\ell,d_1}(x_f|\bz',x',t_f-1,\mathfrak{E}^{\ell}_{t_f})\xi_{\bos,\pi}^{d_0}(t_f-1,\mathfrak{E}^{\ell}_{t_f}|\bz',x')\Bigg).
\end{align}
We now have three expectations $\expec_{\bos'\sim P_{\bos}}=\expec_{x'\sim P_x}\expec_{\ba\sim F}\expec_{\bz'\sim P_{\bz}}$ that can not yet be paired with a distinct factor because the conditioning variables are from multiple state-spaces for each function. Note that the tilde in $\widetilde{\pfun}$ is used to denote that the function uses all of the conditioning variables (this will soon be dropped).

For time $t_f=t_0$, and step $d_1$, the STOK $\eta^{d_0}$ evaluates to zero, therefore we have:
$f_2(\bos,\ba) = f_{2,x}(x,a)\prod_k f_{2,z_k,\ell}(z_k,\ba(k))$ and $f_1(\bos,\ba)=1-f_2(\bos,\ba)$; from Eq. \eqref{appx:eta-bellman-op}, at $t_0$ the products-space STOK is defined:

\begin{align}\label{eq:boundary}
    &\widetilde{\eta}_{\pi}^{d_1}(\bz_f,x_f,t_0,\mathfrak{E}^{\ell}_{t_0}|(\bz,x)_i)= f_1(\bos,\ba) = \big(1-f_{2,x}(x_i,\pi(x_i))\prod_k f_{2,z_k,\ell}(z_k,a_k)\big)\delta_{if}
\end{align}
Notice above that this function evaluates to $1$ if $(\bz_f,x_f) = (\bz_i,x_i)$ due to the Kronecker delta $\delta_{if}$.  We will also write the product-space stock in the following form using the chain rule:
\begin{align}
    \widetilde{\eta}_{\pi}^{d_1}(\bz_f,x_f,t_0,\mathfrak{E}^{\ell}_{t_0}|\bz,x_i)=\widetilde{\pfun}_{\pi}(x_f|\bz,x_i,t_0,\mathfrak{E}^{\ell}_{t_0})\widetilde{\pfun}_{\bz,\ell}(\bz_f|\bz,x,x_f,t_0,\mathfrak{E}^{\ell}_{t_0})\widetilde{\xi}_{\bos}^{d_1}(t_0,\mathfrak{E}^{\ell}_{t_0}|\bz,x).\label{appx:chainrule}
\end{align}
The function $\rho$ is the \textit{state prediction kernel} (SPK), and we will see shortly that we can reduce it down to a smaller domain. We can write an equivalency between \eqref{eq:boundary} and \eqref{appx:chainrule} using the fact that
since $\widetilde{\pfun}_{\pi}$ and $\widetilde{\pfun}_{\bz,\ell}$ must evaluate to $1$ when $t=t_0$ and the final state is the same as the initial (or else it evaluates to $0$), and $\widetilde{\xi}_{\bos}^{d_1}(t_0,\mathfrak{E}^{\ell}_{t_0}|\bz,x)=1$  we have:
\begin{align}
    =~&\widetilde{\pfun}_{\pi}(x_f|\bz,x_i,t_0,\mathfrak{E}^{\ell}_{t_0})\widetilde{\pfun}_{\bz,\ell}(\bz_f|\bz,x,x_f,t_0,\mathfrak{E}^{\ell}_{t_0})\widetilde{\xi}_{\bos}^{d_1}(t_0,\mathfrak{E}^{\ell}_{t_0}|\bz,x),\\
    &\text{where: }\quad \widetilde{\xi}_{\bos}^{d_1}(t_0,\mathfrak{E}^{\ell}_{t_0}|\bz_j,x_i) = \big(1-f_{2,x}(x_i,\pi(x_i))\prod_k f_{2,z_k,\ell}(z_k,a_k)\big)\\
    &\text{where: }\quad \widetilde{\pfun}_{\pi}(x_f|\bz,x_i,t_0,\mathfrak{E}^{\ell}_{t_0}) = \delta_{if}\\
    &\text{where: }\quad \widetilde{\pfun}_{\bz,\ell}(\bz_f|\bz_j,x_i,x_f,t_0,\mathfrak{E}^{\ell}_{t_0}) = \delta_{jf}.
\end{align}
We can then substitute in \textit{reduced} $\pfun$ functions (without the tilde, defined on a smaller domain) because since only $t_0$ time-steps have passed, the initial and final state distributions must be sharp and conditionally independent of other state-spaces:
\begin{align}
    \widetilde{\eta}_{\pi}^{d_1}(\bz_f,x_f,t_0,\mathfrak{E}^{\ell}_{t_0}|\bz,x_i)=\pfun_{\pi}^{\ell}(x_f|x,t_0,\mathfrak{E}^{\ell}_{t_0})\pfun_{\bz}^{\ell}(\bz_f|\bz_i,t_0,\mathfrak{E}^{\ell}_{t_0})\xi^{d_1}_{\bos}(t_0,\mathfrak{E}^{\ell}_{t_0}|\bz, x),
\end{align}

\subsubsection*{The State-Prediction Kernel $\rho$ can be computed outside feasibility iteration}
While the above definition for $\rho$ was for $t_0$, we can say something about this function for all $t_f>t_0$. Notice here that $\mathfrak{E_{t_f-1}}=1$ and indicates that the agent has remained in region $\ell$ from time $t_0$ up to $t_f-1$ where this is a potential region exiting event. This means we know that the variable $\ba_{\ell}$ holds over the past and $\rho$ is simply defined in terms of a Markov matrix power:
\begin{align}   
\pfun_{\bz}^{\ell}(\bz_f|\bz',t_f-1, \mathfrak{E_{t_f-1}})&= P_{\bz,\ba_{\ell},c}^{t_f-1}(i,f),\\
    &= \prod_k P_{z_k,\alpha_{\ell},c}^{t_f-1}(i,f) = \prod_k \pfun_{z_k}^{\ell}(z_f|z,t_f),\label{eq:HL-SPK-factorization}
\end{align}
where $P_{\bz,\ba_{\ell}}(i,j) = P_{\bz}(\bz'|\bz,\ba_{\ell})$ and $P_{z_k,\alpha_{\ell}}(i,j) = P_{z}(z'|z,\alpha_{k,\ell})$ are  Markov chain matrix set to the action $\alpha_{\ell}$ and the constraint states dictated by $f_{c,\ell}(z,\alpha_{\ell})$ are treated as absorbing. Because $\rho$ is defined independently of feasibility iteration, it does not need to be indexed by the iteration variable $d$ in the equations.

A simple remark can be made here:
\begin{remark}
    The probability $p(T=t|\mathfrak{E}_t=1)=1$ and $p(T=t|\mathfrak{E}_t=0)=0$ and this will remain true as $d\rightarrow \infty$ (the time R.V. $T$ is simply stating when the first event happens) so it is redundant.
    Therefore, we can drop $\mathfrak{E}_t$ from all subsequent equations (unless necessary to make a theoretical argument) and replace $\ba$ with $\ba_{\ell}$ since we know that the HL variables will be in region $\mc R_{\ell}$ up until the first-event.
\end{remark} 

We apply the conditioning variable reduction at $t_0$, as seen in Eq. \eqref{eq:boundary}:
\begin{align}    
    \widetilde{\eta}_{\pi}^{d_1}(\bz_f,x_f,t_0,\mathfrak{E}^{\ell}_{t_0}|\bz,x_i)=\pfun_{\bz}^{\ell}(\bz_f|\bz_i,t_0,\mathfrak{E}^{\ell}_{t_0})\pfun_{\pi}^{\ell}(x_f|x,t_0,\mathfrak{E}^{\ell}_{t_0})\xi^{d_1}_{\pi}(t_0,\mathfrak{E}^{\ell}_{t_0}|\bz, x)=f_1(\bos,\pi(\bos_i))\delta_{if}.\label{eq:eqfort_1}
    \end{align}


Regrouping like terms we obtain:
\begin{align}
\eta^{d_1}_{\pi}(\bos_f,t_1|\bos)=
\bar{\mathbbm{1}}_{t_0}(t_1)f_2(x,a)\expec_{x'\sim P_x}\left[\pfun_{\pi}^{\ell}(x_f|x',t_1-1)
f_2(\bz,\ba_{\ell})\expec_{\bz'\sim P_{\bz}}\left[\pfun_{\bz}^{\ell}(\bz_f|\bz',t_1-1)
\xi^{d_0}_{\bos,\pi}(t_1-1|\bz',x')\right]\right].\label{appx:fulldecomp}
\end{align}
Since $\eta^{d_1}_{\pi}(\bos_f,t_0|\bos)$ has its own conditions at $t_0$, we will leave the $\mathbbm{1}_{t_0}(t_1)f_1(\bos,a)$ term out of subsequent equations unless needed.

\subsubsection*{Factorizing the Temporal Event Function at Feasibility Iteration Step $d_1$}\label{sec:TEFatd0}

The TEF $\xi$ in equation \eqref{appx:fulldecomp} is a function of the full state-vector $\bos'=(\bz',x')$, but it can be factorized over all $n$ HL state-spaces and the $1$ BL state-space. The strategy will to be show than the inseparable Bellman update can be decomposed into separate Bellman updates over every state-space by starting at time $t_0$ and feasibility iteration step $d_1$, and show a decomposition holds for any $t_f$ and for all steps $d$, by induction.
\subsubsection*{Boundary condition} We start with the boundary time $t_0$. Recall from Eq. \eqref{appx:eta-bellman-op}, $\eta$ evaluated at $t_0$ is:
\begin{align}
    \eta_{\bos,\pi}^{d_1}(\bos_f,t_0|\bos)&= f_1(\bos,\boa) = 1-f_{2,s_{k_0}}(x,a_{\pi})f_{2,s_{k_1}}(s_{k_1},\alpha_{\ell})...f_{2,s_n}(s_{k_n},\alpha_{\ell}),\\
    &=1-\left((1-\eta_{s_1}^{d_1}(s_{k_1,f},t_0|s_{k_1}))(1-\eta_{s_2}^{d_1}(s_{k_2,f},t_0|s_{k_2}))...(1-\eta_{s_n}^{d_1}(s_{k_n,f},t_0|s_{k_n}))\right).
\end{align}
By summing over $\bos_f$ we obtain the following relationship to $\kappa_{\bos}$, which as we can see, has a tractable factorization,
\begin{align}    
    \xi^{d_1}_{\bos,\pi}(t_0|\bos)&=\sum_{\bos_f}\eta_{\bos,\pi}^{d_1}(\bos_f,t_0|\bos)\\
    &= \sum_{\bos_f}\left(1-\left((1-\eta_{s_1}^{d_1}(s_{k_1,f},t_0|s_{k_1}))(1-\eta_{s_2}^{d_1}(s_{k_2,f},t_0|s_{k_2}))...(1-\eta_{s_n}^{d_1}(s_{k_n,f},t_0|s_{k_n}))\right)\right), \\
    &= 1-\left((1-\sum_{s_{1,f}}\eta_{s_1}^{d_1}(s_{k_1,f},t_0|s_{k_1}))(1-\sum_{s_{2,f}}\eta_{s_2}^{d_1}(s_{k_2,f},t_0|s_{k_2}))...(1-\sum_{s_{n,f}}\eta_{s_n}^{d_1}(s_{k_n,f},t_0|s_{k_n}))\right), \\
    &=1-\prod_{k}\big(1-\kappa_{s_k}^{d_1}(s_{k},t_0)\big),\hspace{3em} \text{where: }~~\kappa_{s_k}(s_{k},t_0)=\sum_{s_{k,f}}\eta_{z}(s_{k,f},t_0|s_{k}),\label{eq:compliment-kappa}\\
    &=1-\prod_{k}\bar{\kappa}^{d_1}_{s_k}(s_k,t_0)\label{eq:compliment-kappa-2}\hspace{6.00em} \text{where: }~~\bar{\kappa}_{s_k}(s_k,t_0)= 1- \kappa_{s_k}(s_k,t_0),~~\\
    &=1-\bar{\kappa}_{\bos}^{d_1}(\bos,t_0),\hspace{7.3em} \text{where: }~~\bar{\kappa}_{\bos}(\bos,t_0)=\prod_{k}\bar{\kappa}_{s_k}(s_{k},t_0)=\bar{\kappa}_{\pi}(x,t_0)\prod_{k}\bar{\kappa}_{z_k}(z_{k},t_0),\label{eq:longerversion}\\
    &=\kappa_{\bos}^{d_1}(\bos,t_0).
\end{align}
The function $\bar{\kappa} = 1-\kappa$ is the compliment cumulative event function. In this proof, we will use the form implied by Eq. \eqref{eq:longerversion}:
\begin{align}
    \xi^{d_1}_{\bos}(t_0|\bos)=1-\bar{\kappa}_{\pi}^{d_1}(x,t_0)\prod_{k}\bar{\kappa}_{z_k}^{d_1}(z_{k},t_0).\label{eq:initialform}
\end{align}
For feasibility iteration step $d_1$, we can see that the STOK factorization must hold when $t_f=t_0$:
\begin{align}
    \eta^{d_1}_{\pi}(\bz_f,x_f,t_0|\bz,x) = \rho_{\pi}^{\ell}(x_{f}|x,t_0)\prod_{k}\rho_{z_k}^{\ell}(z_{f,k}|z_k,t_0)\big(1-\bar{\kappa}_{\pi}^{d_1}(x,t_0)\prod_{k}\bar{\kappa}_{z_k}^{d_1}(z_{k},t_0)\big).
\end{align}
This is true for $t_0$, but also for $t_f>t_0$ because of the zero-initiation.  Thus, for step $d_1$ we have:
\begin{align}
    \eta^{d_1}_{\pi}(\bz_f,x_f,t_f|\bz,x) &=\xi_{\bos}^{d_1}(t_f|\bos)\rho_{\pi}^{\ell}(x_{f}|x,t_f)\prod_k\rho_{z_k}^{\ell}(z_{f,k}|z_k,t_f), \quad \forall t_f. \label{appx:final_fact1}
\end{align}
To anticipate the general form of the equation, note that the $1$ in Eq. \eqref{eq:initialform} will correspond to $\bar{\kappa}_{z_k}(z_{k},t_f)=1$ when evaluated on $t_f=0$. Now we move on to step $d_2$, where we will see this more general form.

\subsubsection{Feasibility Iteration Step $d_2$}\label{appx:d2}

For step $d_2$, first we address the $\kappa$-OKBE update and then the $\pi$-OKBE update and show that we can again substitute $\pi_x$ in for $\pi_{\bos}$.

\subsubsection*{Form of the $\kappa$-OKBE on step $d_2$}

For $\kappa^{d_2}$, we can substitute in $\kappa_{\bos}^{d_1}(\bos,x) = f_{c\ell \bz}(\bz,\ba)\kappa_{x}^{d_1}(x_\bos)$ from Eq. \eqref{appx:smaller}:

\begin{align}
    \kappa_{\bos}^{d_2}(\bz,x) &= \max_{a\in \mc A}\left[f_{1,x}(x,a)\prod_k f_{c\ell k}(z_k,\alpha_k) +f_{1,x}(x,a)\prod_k f_{c\ell k}(z_k,\alpha_k)\sum_{x'}P(x'|x,a)\sum_{\ba}\sum_{\bz'}F(\ba|x,a)P(\bz'|\bz,\ba_{\ell})\kappa^{d_1}_{\bos}(\bz',x')\right],\label{kappa_step_d2}\\
    \kappa_{\bos}^{d_2}(\bz,x)&=\max_{a\in \mc A}\left[f_{1,x}(x,a)\prod_k f_{c\ell k}(z_k,\alpha_k) +f_{1,x}(x,a)\prod_k f_{c\ell k}(z_k,\alpha_k)\sum_{x'}P(x'|x,a)\sum_{\ba}\sum_{\bz'}F(\ba|x,a)P(\bz'|\bz,\ba_{\ell})f_{c\ell \bz}(\bz',\ba_{\ell})\kappa_{x}^{d_1}(x_\bos')\right].\label{kappa_d1}
\end{align}

\subsubsection*{Equivalent actions sets for $\kappa$ on step $d_2$}
The equivalent action sets for the product-space $\kappa$-OKBE and the reduced BL $\kappa$-OKBE are equal (that is, $\mc A^{*,d_2}_{\bos}= \mc A^{*,d_2}_{x_\bos} $) because we can drop the functions of $z_k$ as they are constants with respect to $x$:
\begin{align}
    \mc A^{*,d_2}_{\bos} &= \argmax_{a\in \mc A}\left[f_{1,x}(x,a)\prod_k f_{c\ell k}(z_k,\alpha_k) +f_{1,x}(x,a)\prod_k f_{c\ell k}(z_k,\alpha_k)\sum_{x'}P(x'|x,a)\sum_{\alpha_k}\sum_{\bz'}F(\ba_k|x,a)P(\bz'|\bz,\ba_{\ell})f_{c\ell \bz}(\bz',\ba_{\ell})\kappa_{x}^{d_1}(x_\bos')\right],\\
    &= \argmax_{a\in \mc A}\left[f_{\dg}(x,a)f_{c\ell x}(x,a_{\pi}) +\bar{f}_{\dg}(x,a)f_{c\ell x}(x,a_{\pi})\sum_{x'}P(x'|x,a)\kappa^{d_1}_x(x')\right]= \mc A^{*,d_2}_{x_\bos}.
\end{align}

\subsubsection*{Initial conditions for  $\pi$ on step $d_2$}
The policy for step $d_2$ is:
\begin{align}
    \pi^{d_2}_{\bos}(\bos) &= \argmin_{a\in \mc A^*_{\bos}}\left[f_2(\bos,\boa)\expec_{\bos'\sim P_{\bos}}\sum_{\bos_f}\sum_{t_f}(t_f+1)\eta^{+ d_1}(\bos_f,t_f|\bos')\right],\label{eq:pol-reduce-0}\\
    &=\argmin_{a\in \mc A^*_{\bos}}\left[f_2(x,a)\prod_k f_2(z_{k},\alpha_k)\expec_{\bos'\sim P_{\bos}}\sum_{z_{k,f}}\sum_{x_f}\sum_{t_f}(t_f+1)\big(1-\bar{\kappa}_{\pi}^{d_1}(x',t_f)\prod_{k}\bar{\kappa}_{z_k}^{d_1}(z_{k}',t_f)\big)\rho_{\pi}^{\ell}(x_{f}|x',t_f)\rho_{z_k}^{\ell}(z_{f,k}|z_k',t_f)\right],
\end{align}
where $\expec_{\bos'\sim P_{\bos}} = \expec_{x'\sim P_{x}}\expec_{\ba\sim F}\expec_{\bz'\sim P_{\bz}}$. Functions with only high-level $z$ variables will always be constant with respect to the $\argmin$ function (the HL dynamics are consistent within region $\mc R_{\ell}$) and therefore can be dropped. This results in the form of the policy optimization for the reduced policy in Eq. \eqref{eq:pol-reduce-0}:
\begin{align}
    &=\argmin_{a\in \mc A^*_{x}}\left[f_2(x,a)\expec_{x'\sim P_{x}}\sum_{x_f}\sum_{t_f}(t_f+1)\big(1-\bar{\kappa}_{\pi}^{d_1}(x',t_f)\big)\rho_{\pi}^{\ell}(x_{f}|x',t_f)\right],\\
    &=\argmin_{a\in \mc A^*_{x}}\left[f_2(x,a)\expec_{x'\sim P_{x}}\sum_{x_f}\sum_{t_f}(t_f+1)\xi_{\pi}^{d_1}(t_f|x)\rho_{\pi}^{\ell}(x_{f}|x',t_f)\right],\\
    &=\argmin_{a\in \mc A^*_{x}}\left[f_2(x,a)\expec_{x'\sim P_{x}}\sum_{x_f}\sum_{t_f}(t_f+1)\eta_{\pi}^{d_1}(x_f,t_f|x')\right],\\
    \pi^{d_2}_{\bos}(\bos)&=\pi_x^{d_2}(x_{\bos}).\label{eq:pol-reduce-1}
\end{align}
Thus, the policy $\pi_x^{d_2}(x_{\bos})$ for the reduced problem on $\mc X$ can be substituted for $\pi_{\bos}^{d_2}(\bos)$. We now address the factorization of $\eta_{\bos}^{d_2}$ at time $t_1$.

For step $d_2$, the STOK factorization holds when evaluated $t_0$ because it has the $t_0$ boundary conditions discussed in sec. \ref{sec:t0-conditions}. Thus we now only need to analyze time $t_1$ (as later times the function only evaluates to zero).

\subsubsection*{Computing $\xi_{\bos,\pi}$ evaluated at $t_1$}
We can now use this boundary result (at $t_0$) to obtain $\xi_{\bos,\pi}$ evaluated at $t_1$. We start with $\eta$ evaluated at $t_1$, substitute in the factorization for $\xi_{\bos,\pi}$ (Eq. \eqref{eq:initialform}) into Eq. \eqref{eq:first_eq}, and distribute the terms:
\begin{align}
\eta^{d_2}_{\bos}(\bos_f,t_1|\bos)&=f_2(x,a)\expec_{x'\sim P_{x}}\left(\pfun_{\pi}^{\ell}(x_f|x',t_0)f_2(\bz,\ba_{\ell})\expec_{\bz'\sim P_{\bz}}\left(\pfun_{\bz}^{\ell}(\bz_f|\bz',t_0)\xi_{\bos,\pi}^{d}(t_0|\bz',x')\right)\right),\label{eq:first_eq}\\
&=f_2(x,a)\expec_{x'\sim P_{x}}\left(\pfun_{\pi}^{\ell}(x_f|x',t_0)f_2(\bz,\ba_{\ell})\expec_{\bz'\sim P_{\bz}}\left(\pfun_{\bz}^{\ell}(\bz_f|\bz',t_0)
\bigg(1-\bar{\kappa}^{d}_{\pi}(x',t_0)\prod_{k}\bar{\kappa}^{d}_{z_k}(z_{k}',t_0)\bigg)\right)\right),\\
&=\left(f_2(x,a)\expec_{x'\sim P_{x}}\pfun_{\pi}^{\ell}(x_f|x',t_0)\right)\left(f_2(\bz,\ba_{\ell})\expec_{\bz'\sim P_{\bz}}\pfun_{\bz}^{\ell}(\bz_f|\bz',t_0)\right)\\
&\quad\quad\quad\quad -
f_2(x,a)\expec_{x'\sim P_{x}}\left(\pfun_{\pi}^{\ell}(x_f|x',t_0)f_2(\bz,\ba_{\ell})\expec_{\bz'\sim P_{\bz}}\left(\pfun_{\bz}^{\ell}(\bz_f|\bz',t_0)\bar{\kappa}^{d}_{\pi}(x',t_0)\prod_{k}\bar{\kappa}^{d}_{z_k}(z_{k}',t_0)\right)\right),\\
&=\left(f_2(x,a)\expec_{x'\sim P_{x}}\pfun_{\pi}^{\ell}(x_f|x',t_0)\right)\prod_{k}\left(f_{2,k}(z_k,\alpha_k)\expec_{z_k'\sim P_{z_k}}\pfun_{z_k}^{\ell}(z_{k,f}|z_k',t_0)\right)\\
&\quad\quad\quad\quad-\left(f_2(x,a)\expec_{x'\sim P_{x}}\pfun_{\pi}^{\ell}(x_f|x',t_0)\bar{\kappa}^{d}_{\pi}(x',t_0)\right)\prod_{k}\left(f_{2,k}(z_k,\alpha_k)\expec_{z_k'\sim P_{z_k}}\pfun_{z_k}^{\ell}(z_{k,f}|z_k',t_0)\bar{\kappa}^{d}_{z_k}(z_{k}',t_0)\right).
\end{align}
Note again that $f_2(x,a)\expec_{x'\sim P_{x}}\pfun_{\pi}^{\ell}(x_f|x',t_0)=\pfun_{\pi}^{\ell}(x_f|x,t_1)$ in the above equation is taking the one-step expectation of the Markov dynamics under $\ba_{\ell}$.
By summing over $\bos_f$ in $\eta_{\bos}$ we can obtain $\xi_{\bos,\pi}^{d_2}$ evaluated at $t_1$:
\begin{align}
    \xi_{\bos,\pi}^{d_2}(t_1|\bos)&=\sum_{\bos_f}\eta^{d_2}_{\bos}(\bos_f,t_1|\bos),\\
    &=\sum_{x_f}\left(f_2(x,a)\expec_{x'\sim P_{x}}\pfun_{\pi}^{\ell}(x_f|x',t_0)\right)\prod_k\Bigg(\sum_{z_{f,k}}f_{2,z_k}(z_k,\alpha_k)\expec_{z_k'\sim P_{z_k}}\pfun_{z_k}^{\ell}(z_{f,k}|z_k',t_0)\Bigg)\\
    &\qquad\qquad-\sum_{x_f}\left(f_2(x,a)\expec_{x'\sim P_{x}}\pfun_{\pi}^{\ell}(x_f|x',t_0)\bar{\kappa}^{d_1}_{\pi}(x',t_0)\right)\prod_{k}\Big(\sum_{z_{f,k}}f_{2,k}(z_k,\alpha_k)\expec_{z_k'\sim P_{z_k}}\pfun_{z_k}^{\ell}(z_{k,f}|z_k',t_0)\bar{\kappa}^{d_1}_{z_k}(z_{k}',t_0)\Big),\\
    &=\Bigg(f_2(x,a)\left(\cancel{\expec_{x'\sim P_{x}}\sum_{x_f}\pfun_{\pi}^{\ell}(x_f|x',t_0)}\right)\Bigg)\prod_k\Bigg(f_{2,z_k}(z_k,\alpha_k)\cancel{\expec_{z_k'\sim P_{z_k}}\sum_{z_{f,k}}\pfun_{z_k}^{\ell}(z_{f,k}|z_k',t_0)}\Bigg)\\
    &\qquad\qquad-f_2(x,a)\left(\expec_{x'\sim P_{x}}\bar{\kappa}^{d}_{\pi}(x',t_0)\cancel{\sum_{x_f}\pfun_{\pi}^{\ell}(x_f|x',t_0)}\right)\prod_{k}\Bigg(f_{2,k}(z_k,\alpha_k)\expec_{z_k'\sim P_{z_k}}\bar{\kappa}^{d}_{z_k}(z_{k}',t_0)\cancel{\sum_{z_{f,k}}\pfun_{z_k}^{\ell}(z_{k,f}|z_k',t_0)}\Bigg),\\
    &=f_2(x,a)\prod_k\Bigg(f_{2,z_k}(z_k,\alpha_k)\Bigg)-\left(f_2(x,a)\expec_{x'\sim P_{x}}\bar{\kappa}^{d}_{\pi}(x',t_0)\right)\prod_{k}\Big(f_{2,k}(z_k,\alpha_k)\expec_{z_k'\sim P_{z_k}}\bar{\kappa}^{d}_{z_k}(z_{k}',t_0)\Big).\label{eq:continuting}
\end{align}
Recall Eq. \eqref{eq:compkappa_identity}, reproduced here in the form of a Bellman update for $\bar{\kappa}$ for a Bellman Operator $\bar{\kappa}^{d+1}\leftarrow\mc B_{\bar{\kappa}}\bar{\kappa}^d$: 
\begin{align}
    \bar{\kappa}^{d+1}(s,t_f)&=
f_2(s,a_{\pi})\expec_{s'\sim P}\bar{\kappa}^{d}(s',t_f-1).\label{eq:individual-bellop}
\end{align}
Notice that an equation of the form $f_2(s,a)\expec_{s'\sim P}\bar{\kappa}(s',t_f)$ is equivalent to shifting the time by $1$ in the CEF, i.e. $\bar{\kappa}(s,t_f+1)$. Therefore, continuing from Eq. \eqref{eq:continuting} we have:
\begin{align}
    \implies \xi_{\bos,\pi}^{d_2}(t_1|\bos)&=\bar{\kappa}^{d_2}(x,t_0)\prod_k \bar{\kappa}_{z_k}^{d_2}(z_k,t_0)-\bar{\kappa}_{\pi}^{d_2}(x,t_1)\prod_k\bar{\kappa}_{z_k}^{d_2}(z_k,t_1),\\
    \xi_{\bos,\pi}^{d_2}(t_1|\bos)&=\bar{\kappa}_{\bos}^{d_2}(\bos,t_0)-\bar{\kappa}_{\bos}^{d_2}(\bos,t_1),\label{eq:TEF-as-compkappa}
\end{align}
where the compliment CEF $\bar{\kappa}_{\bos}^{d_2}$ on the full product space is product of the individual compliment CEFs $\bar{\kappa}_{s_k}$,
\begin{align}
    \bar{\kappa}_{\bos}^{d_2}(\bos,t_1)=\bar{\kappa}_{\pi}^{d_2}(x,t_1)\prod_k\bar{\kappa}_{z_k}^{d_2}(z_k,t_1).\label{eq:onlyt1}
\end{align}
Thus, by substituting Eq. \eqref{eq:onlyt1} into Eq. \eqref{eq:first_eq}, we can see that at feasibility iteration step $d_2$, the full STOK decomposes:
\begin{align}
    \eta_{\bos,\pi}^{d_2}(\bz_f,x_f,t_1|\bz,x)&=\rho_{\pi}^{\ell}(x_f|x,t_1)\prod_k\rho_{z_k}^{\ell}(z_{f,k}|z_k,t_1)\Big(\bar{\kappa}^{d_2}_{\pi}(x,t_0)\prod_k \bar{\kappa}_{z_k}^{d_2}(z_k,t_0)-\bar{\kappa}_{\pi}^{d_2}(x,t_1)\prod_k\bar{\kappa}_{z_k}^{d_2}(z_k,t_1)\Big),\\
    \eta_{\bos,\pi}^{d_2}(\bz_f,x_f,t_1|\bz,x)&=\xi_{\bos,\pi}^{d_2}(t_1|\bos)\rho_{\pi}^{\ell}(x_f|x,t_1)\prod_k\rho_{z_k}^{\ell}(z_{f,k}|z_k,t_1),\label{appx:final_fact2}
\end{align}
which is true when $t_f=0$ and $t_f=1$ as show in the previous sections, but also when $t_f>t_1$ due to the fact that all functions output zero due to the initialization. Having shown that the STOK factorization holds for $d_1$ and $d_2$, by induction we can now show that for any $d+1$ this STOK factorization holds for all $t_f$.

\subsubsection{Feasibility Iteration for any step $d+1$}\label{appx:dn}
We proved the STOK factorization for steps $d_1$ and $d_2$, but now we can easily extend this to the general result of $d+1$ for any $d$. For the $\kappa$-OKBE, its straightforward to show that the action sets, once again, are the same for the full problem and the reduced problem on $\mc X$ ($\mc A^*_{\bos} = \mc A^*_{x_{\bos}}$), and we will not reproduce these steps. For the policy, the exact same steps performed between the equations \eqref{kappa_step_d2} and \eqref{eq:pol-reduce-1} can be repeated for $d+1$.  Note that anytime we substitute a $\kappa$-OKBE like Eq. \eqref{kappa_d1} into the subsequent step, the $z$ terms will not affect the $\argmin$ function, this will hold for any $\kappa^{d+1}$, shown below (we omit the intermediate steps, which mirror \eqref{kappa_step_d2} through \eqref{eq:pol-reduce-1}), resulting in:
\begin{align}
    \pi^{d+1}_{\bos}(\bos) &= \argmin_{a\in \mc A^*_{\bos}}\left[f_2(\bos,\boa)\expec_{\bos'\sim P_{\bos}}\sum_{\bos_f}\sum_{t_f}(t_f+1)\eta^{d+1}_{\bos,\pi}(\bos_f,t_f|\bos')\right],\\
    &=\argmin_{a\in \mc A^*_{x_\bos}}\left[f_2(x,a)\expec_{x'\sim P_{x}}\sum_{x_f}\sum_{t_f}(t_f+1)\eta_{x,\pi}^{d+1}(x_f,t_f|x')\right],\\
    \pi^{d+1}_{\bos}(\bos)&=\pi_x^{d+1}(x_{\bos}).
\end{align}
Thus, we can again subistute in the reduced policy on $\mc X$ for the full policy on $\mc S$ to use for $\bar{\kappa}_{\pi_{x}}$ and $\rho_{\pi_{x}}$.

\subsubsection*{Computing $\eta_{\bos,\pi}^{d+1}$ for step $d+1$ for all $t_f$}
We can now see in Eq. \eqref{eq:TEF-as-compkappa} that the TEF is the difference of the compliment CEF between time-steps. By induction, we can derive a general form for $\xi_{\bos,\pi}$. It is straightforward to show that any $\xi_{\bos,\pi}(t_f-1|\bos)=\bar{\kappa}_{\bos}(\bos,t_f-2)-\bar{\kappa}_{\bos}(\bos,t_f-1)$ can be substituted into Eq. \eqref{eq:first_eq} to derive the TEF evaluated at the next time step as $\xi_{\bos,\pi}(t_f|\bos)=\bar{\kappa}_{\bos}(\bos,t_f-1)-\bar{\kappa}_{\bos}(\bos,t_f)$:

\begin{align}
\eta^{d+1}_{\bos,\pi}(\bos_f,t_f|\bos)&=f_2(x,a)\expec_{x'\sim P_{x}}\left(\pfun_{\pi}^{\ell}(x_f|x',t_f-1)f_2(\bz,\ba_{\ell})\expec_{\bz'\sim P_{\bz}}\left(\pfun_{\bz}^{\ell}(\bz_f|\bz',t_f-1)\xi_{\bos,\pi}^{d}(t_f-1|\bz',x')\right)\right),\\
&=\left(f_2(x,a)\expec_{x'\sim P_{x}}\pfun_{\pi}^{\ell}(x_f|x',t_f-1)\right)\left(f_2(\bz,\ba_{\ell})\expec_{\bz'\sim P_{\bz}}\pfun_{\bz}^{\ell}(\bz_f|\bz',t_f-1)\right)\\
&\qquad\qquad\times
\bigg(\bar{\kappa}_{\pi}^{d}(x',t_f-2)\prod_{k}\bar{\kappa}^{d}_{z_k}(z_{k}',t_f-2) - \bar{\kappa}_{\pi}^{d}(x',t_f-1)\prod_{k}\bar{\kappa}^{d}_{z_k}(z_{k}',t_f-1)\bigg),\\
&=\left(f_2(x,a)\expec_{x'\sim P_{x}}\pfun_{\pi}^{\ell}(x_f|x',t_f-1)\bar{\kappa}_{\pi}^{d}(x',t_f-2)\right)\prod_{k}\left(f_{2,s_k}(z_k,\alpha_k)\expec_{z_k'\sim P_{z_k}}\pfun_{z_k}^{\ell}(z_{f,k}|z_k',t_f-1)\bar{\kappa}_{z_k}^{d}(z_{k}',t_f-2)\right)\\
&\qquad\qquad-\left(f_2(x,a)\expec_{x'\sim P_{x}}\pfun_{\pi}^{\ell}(x_f|x',t_f-1)\bar{\kappa}_{\pi}^{d}(x',t_f-1)\right)\prod_{k}\Big(f_{2,k}(z_k,\alpha_k)\expec_{\bz'\sim P_{\bz}}\pfun_{z_k}^{\ell}(z_{k,f}|z_k',t_f-1)\bar{\kappa}_{z_k}^{d}(z_{k}',t_f-1)\Big).
\end{align}
We sum over $\bos_f$ to obtain the TEF $\xi_{\bos,\pi}$ at $t_f$:
\begin{align}
    \xi_{\bos,\pi}^{d+1}(t_f|\bos)&=\sum_{\bos_f}\eta^{d+1}_{\bos,\pi}(\bos_f,t_f|\bos)\\
    &=f_2(x,a)\left(\expec_{x'\sim P_{x}}\bar{\kappa}_{\pi}^d(x',t_f-2)\cancel{\sum_{x_f}\pfun_{\pi}^{\ell}(x_f|x',t_f-1)}\right)\prod_k\Bigg(f_{2,z_k}(z_k,\alpha_k)\expec_{z_k'\sim P_{z_k}}\bar{\kappa}_{z_k}^d(z_{k}',t_f-2)\cancel{\sum_{z_{f,k}}\pfun_{z_k}^{\ell}(z_{f,k}|z_k',t_f-1)}\Bigg)\\
    &\qquad-f_2(x,a)\expec_{x'\sim P_{x}}\left(\bar{\kappa}_{\pi}^d(x',t_f-1)\cancel{\sum_{x_f}\pfun_{\pi}^{\ell}(x_f|x',t_f-1)}\right)\prod_{k}\left(f_{2,k}(z_k,\alpha_k)\expec_{z_k'\sim P_{z_k}}\left(\bar{\kappa}_{z_k}^d(z_{k}',t_f-1)\cancel{\sum_{z_{f,k}}\pfun_{z_k}^{\ell}(z_{k,f}|z_k',t_f-1)}\right)\!\right)\!,\\
     &=\left(f_2(x,a)\expec_{x'\sim P_{x}}\bar{\kappa}_{\pi}^d(x',t_f-2)\right)\prod_k\left(f_{2,z_k}(z_k,\alpha_k)\expec_{z_k'\sim P_{z_k}}\bar{\kappa}_{z_k}^d(z_{k}',t_f-2)\right),\\
    &\qquad-\left(f_2(x,a)\expec_{x'\sim P_{x}}\bar{\kappa}_{\pi}^d(x',t_f-1)\right)\prod_{k}\Big(f_{2,k}(z_k,\alpha_k)\expec_{z_k'\sim P_{z_k}}\bar{\kappa}_{z_k}^d(z_{k}',t_f-1)\Big),\\
    &=\bar{\kappa}^{d+1}(x,t_f-1)\prod_k \bar{\kappa}_{z_k}^{d+1}(z_k,t_f-1)-\bar{\kappa}_{\pi}^{d+1}(x,t_f)\prod_k\bar{\kappa}_{z_k}^{d+1}(z_k,t_f),\label{eq:generalTEF-factored}\\
    &=\bar{\kappa}_{\bos}^{d+1}(\bos,t_f-1)-\bar{\kappa}_{\bos}^{d+1}(\bos,t_f),\label{eq:generalTEF}
\end{align}
which is the general form of the TEF factorization, where Eq. \eqref{eq:generalTEF} is equal to Eq. \eqref{eq:initialform} when $t_f=-1$.
and is an instance of Eq. \eqref{eq:onlyt1} when $t_f=1$. Note that the above Eqs. \eqref{eq:generalTEF-factored} and \eqref{eq:generalTEF},
\begin{align}
    \bar{\kappa}_{\bos}^{d+1}(\bos,t_f)=\bar{\kappa}_{\pi}^{d+1}(x,t_f)\prod_k\bar{\kappa}_{z_k}^{d+1}(z_k,t_f).
\end{align}
The general form for $\xi_{\bos}$ on the product-space is therefore:
\begin{align}
    \xi_{\bos,\pi}^{d+1}(t_f|\bos)&=\bar{\kappa}_{\bos}^{d+1}(\bos,t_f-1)-\bar{\kappa}_{\bos}^{d+1}(\bos,t_f),\\
    &=\prod_k \bar{\kappa}_{s_k}^{d+1}(s_k,t_f-1)-\prod_k\bar{\kappa}_{s_k}^{d+1}(s_k,t_f),
\end{align}
which can be equivalently written with the product-space (regular, non-compliment) CEFs by substituting $\bar{\kappa}_{s_k}^{d+1}(s_k,t_f) = 1-\kappa_{s_k}^{d+1}(s_k,t_f)$:
\begin{align}
    \xi_{\bos,\pi}^{d+1}(t_f|\bos)&=(1-\kappa_{\bos}^{d+1}(\bos,t_f-1))-(1-\kappa_{\bos}^{d+1}(\bos,t_f)),\\
    &=\kappa_{\bos}^{d+1}(\bos,t_f)-\kappa_{\bos}^{d+1}(\bos,t_f-1),\label{eq:final_xi_factorization}
\end{align}
where $\kappa(\bos,t_{0}-1)=0$.
\subsubsection*{A quick verification}
We have shown the $\xi_{\bos}$ decomposes into factors $\bar{\kappa}_{s_k}$ in an inductive step-wise fashion for all $d$ and $t_f$, and we did this by using Eq. \eqref{eq:compkappa_identity} on the RHS of our equations for a substitution. This is adequate for our purposes but we can verify our result by substituting these factors (Eq. \eqref{eq:final_xi_factorization}) back into our original Bellman equation (Eq. \eqref{appx:fulldecomp}) \textit{on both sides of the equation} (Eq.\eqref{eq:bothsides}) to verify that this entails the set of Bellman updates $\bar{\kappa}^{d+1}_{s_k}\leftarrow B_{\bar{\kappa}}\bar{\kappa}^{d}_{s_k}$ for each space $\mc S_k \in \mathbf{S}$:
\begin{align}
\sum_{\bz_f}\sum_{x_f}\eta^{d+1}_{\pi}(\bz_f,x_f,t_f|\bz,x)&=\sum_{\bz_f}\sum_{x_f}f_2(x,a)\expec_{x'\sim P_{x}}\left(\pfun_{\pi}^{\ell}(x_f|x',t_f-1)f_2(\bz,\ba_{\ell})\expec_{\bz'\sim P_{\bz}}\left(\pfun_{\bz}^{\ell}(\bz_f|\bz',t_f-1)\xi_{\bos,\pi}^{d}(t_f-1|\bz',x')\right)\right),\\
    \implies \xi^{d+1}_{\bos,\pi}(t_f|\bos)&=f_2(\bos,\ba_{\ell})\expec_{\bos'\sim P_{\bos}}
\xi_{\bos,\pi}^d(t_f-1|\bos'),\\
\bar{\kappa}_{\bos}^{d+1}(\bos,t_f-1)-\bar{\kappa}_{\bos}^{d+1}(\bos,t_f)&=f_{2}(\bos,\ba_{\ell})\expec_{\bos'\sim P_{\bos}}\Big[
\bar{\kappa}_{\bos}^{d}(\bos',t_f-2)-\bar{\kappa}_{\bos}^{d}(\bos',t_f-1)\Big],\label{eq:bothsides}\\
-\bar{\kappa}_{\bos}^{d+1}(\bos,t_f)&=
f_{2}(\bos,\ba_{\ell})\expec_{\bos'\sim P_{\bos}}\bar{\kappa}_{\bos}^{d}(\bos',t_f-2)-f_{2}(\bos,\ba_{\ell})\expec_{\bos'\sim P_{\bos}} \bar{\kappa}_{\bos}^{d}(\bos',t_f-1)-\bar{\kappa}_{\bos}^{d+1}(\bos,t_f-1),\\
(\text{Eq. }\eqref{eq:compkappa_identity})\qquad\qquad-\bar{\kappa}_{\bos}^{d+1}(\bos,t_f)&=
\cancel{\bar{\kappa}_{\bos}^{d+1}(\bos,t_f-1)}-f_{2}(\bos,\ba_{\ell})\expec_{\bos'\sim P_{\bos}} \bar{\kappa}_{\bos}^{d}(\bos',t_f-1)-\cancel{\bar{\kappa}_{\bos}^{d+1}(\bos,t_f-1)},\\
\prod_k\bar{\kappa}_{s_k}^{d+1}(s_k,t_f)&=
\prod_k f_{2}(s_k,\alpha_{\ell})\expec_{s_k'\sim P_{\bos}} \bar{\kappa}_{s_k}^{d}(s_k',t_f-1),
\\
\bar{\kappa}_{s_k}^{d+1}(s_k,t_f)\prod_{j\neq k}\bar{\kappa}_{s_j}^{d+1}(s_j,t_f)&=
\Bigg(f_{2,k}(s_k,\alpha_{\ell})\expec_{s_k'\sim P_{\bos}} \bar{\kappa}_{s_k}^{d}(s_k',t_f-1)\Bigg)\Bigg(\prod_{j\neq k} f_{2,j}(s_j,\alpha_{s_j})\expec_{s_j'\sim P_{\bos}} \bar{\kappa}_{s_j}^{d}(s_j',t_f-1)\Bigg),
\\
\bar{\kappa}_{s_k}^{d+1}(s_k,t_f)\cancel{\prod_{j\neq k}\bar{\kappa}_{s_j}^{d+1}(s_j,t_f)}&=
\Bigg(f_{2,k}(s_k,\alpha_{\ell})\expec_{s_k'\sim P_{\bos}} \bar{\kappa}_{s_k}^{d}(s_k',t_f-1)\Bigg)\cancel{\prod_{j\neq k}\bar{\kappa}_{s_j}^{d+1}(s_j,t_f)},\\
\implies \bar{\kappa}_{s_k}^{d+1}(s_k,t_f)&=
f_{2,k}(s_k,\alpha_{\ell})\expec_{s_k'\sim P_{\bos}} \bar{\kappa}_{s_k}^{d}(s_k',t_f-1), \quad \forall k.
\end{align}

Thus, the Bellman operator $\mc B_{\xi}$ breaks down into Bellman operators $\mc B_{\bar{\kappa}}$ for the compliment CEF function, which will compute $\bar{\kappa}^{d+1}\leftarrow B_{\bar{\kappa}}\bar{\kappa}^{d}$ for each space $\mc S_k \in \mathbf{S}$. 

\subsubsection*{The Form of the STOK Factorization for any step $d+1$}
We can now substitute our fully factorized decomposition of $\xi_{\bos,\pi}$ from Eq. \eqref{appx:fulldecomp} to get the STOK factorization at any iteration $d+1$:
\begin{align}
    \eta_{\pi}^{d+1}(\bos_f,t_f|\bos)&=f_2(x,a)f_2(\bz,\ba_{\ell})\expec_{\bz'\sim P_{\bz}}\expec_{x'\sim P_x}\pfun_{\pi}^{\ell}(x_f|x',t_f-1)\pfun_{\bz}^{\ell}(\bz_f|\bz',t_f-1)\xi_{\bos,\pi}^d(t_f-1|\bz',x'),\\
    &=f_2(x,a)f_2(\bz,\ba_{\ell})\expec_{\bz'\sim P_{\bz}}\expec_{x'\sim P_x}\Bigg[\pfun_{\pi}^{\ell}(x_f|x',t_f-1)\left(\prod_{k}\pfun_{z_k}^{\ell}(z_{k,f}|z_{k}',t_f-1)\right)\\    &\qquad\qquad\qquad\qquad\qquad\qquad\qquad\qquad\qquad\qquad\qquad\times \left(\Big(1-\prod_{k}\bar\kappa_{s_k}^{d}(s_{k}',t_f-1)\Big) - \Big(1-\prod_{k}\bar\kappa_{s_k}^{d}(s_{k}',t_f-2)\Big)\right)\Bigg],\\
    &=\pfun_{\pi}^{\ell}(x_f|x,t_f)\left(\prod_{k}\pfun_{z_k}^{\ell}(z_{k,f}|z_{k},t_f)\right)\left(\Big(1-\prod_{k}\bar\kappa_{s_k}^{d+1}(s_{k},t_f)\Big) - \Big(1-\prod_{k}\bar\kappa_{s_k}^{d+1}(s_{k},t_f-1)\Big)\right),\\
    \eta_{\pi}^{d+1}(\bz_f,x_f,t_f|\bz,x)&=\xi_{\bos,\pi}^{d+1}(t_f|\bz,x)\pfun_{\pi}^{\ell}(x_f|x,t_f)\pfun_{\bz}^{\ell}(\bz_f|\bz,t_f).\label{appx:final_fact3}
\end{align}
And by substituting $d = d_1$ or $d = d_2$, this factorization reproduces the factorizations we have previously derived in addition all subsequent $d>d_2$.
\subsection{Conclusion}
We have shown by induction with Eqs. \eqref{appx:final_fact1}, \eqref{appx:final_fact2}, and \eqref{appx:final_fact3}, that the STOK factorization holds for each step $d$ of feasibility iteration. Instead of computing feasibility iteration as,
\begin{align}    (\kappa^{d_0}_{\bos},\pi^{d_0}_{\bos},\eta^{d_0}_{\bos})\rightarrow(\kappa^{d_1}_{\bos},\pi^{d_1}_{\bos},\eta^{d_1}_{\bos})\rightarrow(\kappa^{d_2}_{\bos},\pi^{d_2}_{\bos},\eta^{d_2}_{\bos})\rightarrow ... \rightarrow (\kappa_{\bos}^{d_{\infty}},\pi^{d_{\infty}}_{\bos},\eta_{\bos}^{d_{\infty}}),
\end{align}
we can alternatively compute feasibility iteration as,
\begin{align}    (\kappa_x^{d_0},\pi^{d_0}_{x},\{\bar{\kappa}^{d_0}\}_{n})\rightarrow(\kappa_x^{d_1},\pi^{d_1}_{x},\{\bar{\kappa}^{d_1}\}_{n})\rightarrow(\kappa^{d_2}_x,\pi^{d_2}_x,\{\bar{\kappa}^{d_2}\}_{n})\rightarrow ... \rightarrow (\kappa_x^{d_{\infty}},\pi^{d_{\infty}}_x,\{\bar{\kappa}^{d_{\infty}}\}_{n}),
\end{align}
because temporal event function (TEF) $\xi_{\bos,\pi}$ has a definition using $\kappa_{\bos}$ which has components that are factorizable in terms of Eq. \eqref{eq:compliment-kappa} for all steps $d$ until convergence. This concludes the proof of thm. \ref{appx:STOK-thm}. \qed

\subsection{Recap}
We showed that if goal, constraint, and region-violations are encoded in a separable continuation function, then we can use state-prediction kernels $\rho$ that are consistent with region's dynamics, allowing us to apply the chain rule to a high-dimensional STOK and drop conditioning variables using conditional independence. This results in each of the $n$ spaces in the product-space having their own state-prediction kernels (SPK), along with a factorizable temporal event function (TEF) $\xi_{\bos,\pi}$. Thus, this factorization is defined up until the event in which a policy violates the region's conditional-independence properties or the policy completes the goal or fails the task if no region violation occurs.

\subsection{Corollary: No probability of inducing an HL event}

As a special case, we consider when no HL events occur up to time $t_f$. This means $\kappa_{z_k}(z_k,t_f)=0$ for all $k$ corresponding to an HL space $\mc Z_k$, (or equivalently, when $\bar{\kappa}_{\bz}(\bz,t_f)=1$). In the TEF definition, this leaves only $\kappa_{\pi}(x,t_f)$. The factorization then reduces to,
\begin{align}
    \xi_{\bos,\pi}(t_f|\bos)=\kappa_{\bos}(\bos,t_f)-\kappa_{\bos}(\bos,t_f-1) &= \left(1-\prod_{k}\bar\kappa_{s_k}(s_k,t_f)\right)-\left(1-\prod_{k}\bar\kappa_{s_k}(s_k,t_f-1)\right),\\
    &=\kappa_{\pi}(x,t_f)-\kappa_{\pi}(x,t_f-1),\\
    &=\xi_{x}^{\pi}(t_f|x).
\end{align}
This means we can multiply $\xi_{x}^{\pi}$ with $\pfun_{\pi}^{\ell}$ to produce $\eta_{\pi}$, 
\begin{align}
    \eta_{\pi}(x_f,t_f|x)=\xi_{x}^{\pi}(t_f|x)\pfun_{\pi}^{\ell}(x_f|x,t_f).
\end{align}
Thus, when $\bar{\kappa}_{\bz}(\bz,t_f)=1$ the STOK factorization reduces down to:
\begin{align}
    \hat{\eta}_{\pi}(\bz_f,x_f,t_f|\bz,x)&=\xi_{x}^{\pi}(t_f|x)\pfun_{\pi}^{\ell}(x_f|x,t_f)\prod_{k}\pfun_{z_k}^{\ell}(z_{k,f}|z_k,t_f),\\
    \hat{\eta}_{\pi}(\bz_f,x_f,t_f|\bz,x)&=\eta_{\pi}(x_f,t_f|x)\prod_{k}\pfun_{z_k}^{\ell}(z_{k,f}|z_k,t_f).
\end{align}
This concludes the proof of cor. \ref{appx:STOK-cor}. \qed

\newpage
\section{The OKBE Bellman Operator has a Fixed Point}\label{sec:fixedpnt}
Assuming elements of $\kappa$ are initialized in $[0,1]$, we can prove that the OKBE Bellman Operator is bounded and monotonic and therefore has a fixed point. Let $\mc B$ be the Bellman Operator defined:
\begin{align}
    \mc B(\kappa)(x) &= \max_{a}\left[f_{1}(x,a)+f_{2}(x,a)\sum_{x'}P(x'|x,a)\kappa(x')\right],
\end{align}
\subsection{Boundedness}
Assume $\kappa(x)$ is initialized in $[0,1]$. Recall $f_{\dg},f_{c},P,\kappa$ have the codomain $[0,1]$.

The following inequality holds:
\begin{align}
    \mc B(\kappa)(x)&=\max_{a}\left[f_{\dg}(x,a)f_c(x,a)+(1-f_{\dg}(x,a))f_c(x,a)\sum_{x'}P(x'|x,a)\kappa(x')\right],\\
    &= \max_{a}\left[f_c(x,a)\left(f_{\dg}(x,a)+(1-f_{\dg}(x,a))\sum_{x'}P(x'|x,a)\kappa(x')\right)\right],\\
    &\leq \max_{a}\left[f_{\dg}(x,a)+(1-f_{\dg}(x,a))\sum_{x'}P(x'|x,a)\kappa(x')\right],\\
    &\leq \max_{a}\left[f_{\dg}(x,a)+(1-f_{\dg}(x,a))\right],\\
    &= 1.
\end{align}
Thus, $\mc B(\kappa)(x) \leq 1$. Also, since each function ($f_{\dg},1-f_{\dg},f_c,P,\kappa$) has the codomain $[0,1]$ and the Bellman operator only involves addition and multiplication, we have $\mc B(\kappa)(x)\geq 0$. Therefore $\mc B(\kappa)(x)$ is bounded in the closed interval $[0,1]$ for all $x \in \mc X$.
\subsection{The OKBE Bellman Operator is Monotonic}

\newcommand{\kappaone}{\kappa_{1}}
\newcommand{\kappatwo}{\kappa_{2}}
\newcommand{\Bg}{f_{g}}
\newcommand{\Bc}{f_{c}}
\newcommand{\Bell}{\mathcal{B}}
Suppose $\kappaone(x) \le \kappatwo(x)$ for all $x$. We want to show
$\Bell(\kappaone)(x) \le \Bell(\kappatwo)(x)$ for each $x.$

Define
\[
F(\kappa,x,a)
:=
f_1(x,a)
+
f_2(x,a)\sum_{x'} P(x'|x,a)\kappa(x').
\]
Since $\kappaone(x') \le \kappatwo(x')$ for all $x'$, we have
\[
\sum_{x'} P(x'|x,a)\kappaone(x')
\le
\sum_{x'} P(x'|x,a)\kappatwo(x').
\]
Thus, for any action $a$,
\[
F(\kappaone,x,a) \le F(\kappatwo,x,a).
\]
Taking the maximum over $a$,
\[
\max_{a} F(\kappaone,x,a) 
\le 
\max_{a} F(\kappatwo,x,a).
\]
This implies,
\[
\Bell(\kappaone)(x) 
\le 
\Bell(\kappatwo)(x),
\quad \forall x \in \mc X.
\]
Therefore, $\Bell$ is monotonic.

\subsection{The OKBE Bellman Operator is Convergent}
Given that $\mc B:\mc K \rightarrow \mc K$ is monotonic and bounded within the closed interval $[0,1]$, a CFF $\kappa$ converges to a fixed-point under repeated applications of $\mc B$.

\begin{proof}
Let $\kappa^0 \in \mathcal{K}$, and define $\kappa^{n+1} := \mathcal{B}(\kappa^n)$ for all $n \ge 0$.  
Monotonicity implies (using an element-wise inequality),
\[
  \kappa^0 \;\le\; \kappa^1 \;\le\; \kappa^2 \;\le\;\cdots
\]
Boundedness on a closed interval ensures there is a supremum $\kappa^\infty := \sup_{n}\kappa^n \in \mathcal{K}$ (where this is an element-wise supremum which exists in the set $\mc K$ of CFFs).  
For each $n$, $\kappa^n \le \kappa^\infty$ implies $\mathcal{B}(\kappa^n) \le \mathcal{B}(\kappa^\infty)$ by monotonicity, so $\kappa^\infty$ is an upper bound of $\{\kappa^{n+1}\}$.  
Hence $\kappa^\infty \le \mathcal{B}(\kappa^\infty)$.  
By minimality of $\kappa^\infty$ as a supremum, $\mathcal{B}(\kappa^\infty) \le \kappa^\infty$.  
Thus $\mathcal{B}(\kappa^\infty) = \kappa^\infty$, and $\kappa^\infty$ is a fixed point of $\mathcal{B}$.
\end{proof}

\subsection{Additional Notes} The Bellman Operator is convergent, however there is not necessarily a unique solution. We can have situations where, if $\kappa$ is initialized to non-zero values, then these values never go to zero in parts of the state-space that are disconnected to a goal state because the continuation function $f_2$ evaluates to one over these states (and thus, the values will not converge to zero). The computed $\kappa$ will be a numerical solution but it will not representing the true feasibility of the goal under the policy. Therefore enforcing a zero-initialization means that $\kappa$ accurately reports genuine goal-feasibility because it progressively quantifies the true probability over each step of feasibility iteration.

\newpage
\section{Sublimation theorem}\label{appx:sublimation}

For compactness, we will prove this assuming that goal and constraint functions are not functions of actions, as including actions will give us the same result. Assume $\kappa$ is initialized to $\widetilde{\kappa}_{\pi}^{d_0}(\bs,z,x) = 0 = \widetilde{\kappa}_{\pi,\bs}^{d_0}(\bs)$ and the constraint function is separable $f_c(\bs,z,x)=f_c^{\bs}(\bs)f_c^{z}(z)f_c^{x}(x)$. Assume $f_\dg^{\bs}(\bs) = \max_{z,x}f_\dg(\bs,z,x)$, and recall $f_1 = f_{\dg}f_c$, $f_2 = (1-f_{\dg})f_c$.  We will start with the full OKBE and reduce it down to the sublimated OKBE, which will introduce an inequality between $\widetilde{\kappa}_{\pi}$ and $\widetilde{\kappa}_{\pi,\bs}$:

\begin{align}
    \widetilde{\kappa}_{\pi}^{d_2}(\bs,z,x)&=\max_a\left[f_1(\bs,z,x)+f_2(\bs,z,x)\sum_{\bs',z',x'}\widetilde{\kappa}_{\pi}^{d_0}(\bs',z',x')P(\bs',z',x'|\bs,z,x,a)\right],\\
    &=\max_a\left[f_1(\bs,z,x)+f_2(\bs,z,x)\sum_{\bs',z',x'}\widetilde{\kappa}_{\pi}^{d_0,\bs}(\bs')P(\bs',z',x'|\bs,z,x,a)\right],\\
    &=\max_a\left[f_1(\bs,z,x)+f_2(\bs,z,x)\sum_{\alpha_{\bs},\alpha_z,\bs',z',x'}\widetilde{\kappa}_{\pi,\bs}^{d_0}(\bs')P(\bs'|\bs,\alpha_{\bs})P(z'|z,\alpha_{z})F(\alpha_{\bs},\alpha_{z}|x,a)P(x'|x,a)\right],\\
    &=\max_a\left[f_1(\bs,z,x)+f_2(\bs,z,x)\sum_{\bs',\alpha_{\bs}}\widetilde{\kappa}_{\pi,\bs}^{d_0}(\bs')P(\bs'|\bs,\alpha_{\bs})\sum_{\alpha_z,z',x'}P(z'|z,\alpha_{z})F(\alpha_{\bs},\alpha_{z}|x,a)P(x'|x,a)\right],\\
    &=\max_a\left[f_1(\bs,z,x)+f_2(\bs,z,x)\sum_{\bs',\alpha_{\bs}}\widetilde{\kappa}_{\pi,\bs}^{d_0}(\bs')P(\bs'|\bs,\alpha_{\bs})F(\alpha_{\bs}|x,a)\right],\\
    &\leq\max_a\Bigg[\max_{z,x}[f_\dg(\bs,z,x)]f_c(\bs,z,x)+(1-\max_{z,x}[f_\dg(\bs,z,x)])f_c(\bs,z,x)\sum_{\bs',\alpha_{\bs}}\widetilde{\kappa}_{\pi,\bs}^{d_0}(\bs')P(\bs'|\bs,\alpha_{\bs})F(\alpha_{\bs}|x,a)\Bigg],\\
    &= \max_a\Bigg[f_\dg^{\bs}(\bs)f_c(\bs,z,x) +(1-f_\dg^{\bs}(\bs))f_c(\bs,z,x)\sum_{\bs',\alpha_{\bs}}\widetilde{\kappa}_{\pi,\bs}^{d_0}(\bs')P(\bs'|\bs,\alpha_{\bs})F(\alpha_{\bs}|x,a)\Bigg],\\
    &= \max_a\Bigg[f_\dg^{\bs}(\bs)f_c^{\bs}(\bs)f_c^{z}(z)f_c^{x}(x)+(1-f_\dg^{\bs}(\bs))f_c^{\bs}(\bs)f_c^{z}(z)f_c^{x}(x)\sum_{\bs',\alpha_{\bs}}\widetilde{\kappa}_{\pi,\bs}^{d_0}(\bs')P(\bs'|\bs,\alpha_{\bs})F(\alpha_{\bs}|x,a)\Bigg],\\
    &\leq \max_a\left[f_\dg^{\bs}(\bs)f_c^{\bs}(\bs)+(1-f_\dg^{\bs}(\bs))f_c^{\bs}(\bs)\sum_{\bs',\alpha_{\bs}}\widetilde{\kappa}_{\pi,\bs}^{d_0}(\bs')P(\bs'|\bs,\alpha_{\bs})F(\alpha_{\bs}|x,a)\right],\\
    &= \max_a\left[f_1^{\bs}(\bs)+f_2^{\bs}(\bs)\sum_{\bs',\alpha_{\bs}}\widetilde{\kappa}_{\pi,\bs}^{d_0}(\bs')P(\bs'|\bs,\alpha_{\bs})F(\alpha_{\bs}|x,a)\right],\\
    &\leq \max_{\alpha_{\bs}}\left[ f_1^{\bs}(\bs)+f_2^{\bs}(\bs)\sum_{\bs'}\widetilde{\kappa}_{\pi,\bs}^{d_0}(\bs')P(\bs'|\bs,\alpha_{\bs})\right],\\
    &=\widetilde{\kappa}_{\pi,\bs}^{d_2}(\bs).
\end{align}
This completes one step. For each step $d$ we complete the same steps while substituting $\widetilde{\kappa}_{\pi,\bs}^{d}(\bs)$ into the top. Thus, the this inequality holds between $\widetilde{\kappa}_{\pi}^{d}(\bs,z,x)$ and $\widetilde{\kappa}_{\pi,\bs}^{d}(\bs)$ until convergence of $\kappa$ as $d\rightarrow \infty$, resulting in:
\begin{align}
    \widetilde{\kappa}_{\pi}(\bs,z,x) \leq \widetilde{\kappa}_{\pi,\bs}(\bs),
\end{align}
which concludes the proof. $\hfill \qed$

\subsection{Proof Commentary}
Note that this inequality does not require $z$, but it is meant to show how it can apply to any sub-space (here $\Sigma$, which technically is a Cartesian product-space of many two-state state-spaces for each bit) of the full HL space $\Sigma \times \mc Z$, which of course also generalizes to the entire HL space.

In general, the sublimation theorem says something intuitive: If we have an initial vector $\bos = (z_1,z_2,...,z_n,x)$, and a set of accepting vectors $\mc S^*$ (vectors which accept with non-zero probability evaluated with $f_1$), then in order for there to be a feasible policy for the full problem, there must be a feasible policy that can produce a path from each element of $\bos$ (e.g. $z_k$), to the projection $\text{proj}_{\mc Z_k}(\mc S^*)$ of $\mc S^*$ onto a corresponding state-space $\mc Z_k \in \mathbf{Z}$ in the Cartesian product-space. The sublimated feasibility function gives us this information about element-wise feasible paths to the projected solution sets. It is therefore necessary for a true solution $\kappa^{*}(\bos)$ that each $\mc Z_k \in \mathscr{Z}$ has a feasible sublimated solution $\kappa^{*}_{z_k}(z_i)$ from a given element $z_i\in \bos$. However, this is only a necessary condition, because even if there are feasible solutions for each state-space, that does not imply a feasible solution on a (sub-) product-space.
\newpage
\section{Glossary of Important Identities}

Throughout this paper and appendix we have discussed a number of relationships between various versions of these functions: $\kappa$, $\eta$, $\chi$, $\rho$, $\xi$. We list the relationships below.

\begin{align}
    &\kappa(x,t_f) = \sum_{\tau_f=0}^{t_f} \eta(x_f,\tau_f|x),\\
    &\kappa(x) = \lim_{t_f\rightarrow \infty}\kappa(x,t_f),\\
    &\bar{\kappa}(x) = 1-\kappa(x),\\
    &\bar{\kappa}(x,t_f) = 1-\kappa(x,t_f),\\
    &\kappa(x) = \sum_{\xp}\sum_{\tp}\eta^+(\xp,\tp|x),\\
    &\bar{\kappa}(x) = \sum_{\xm}\sum_{\tm}\eta^-(\xm,\tm|x),\\
    &\chi^+(\xp|x) = \sum_{\tp}\eta^+(\xp,\tp|x),\\
    &\chi^-(\xm|x) = \sum_{\tm}\eta^-(\xm,\tm|x),\\
    &\chi_{\pi_o}^{**}(x_f|x)=\sum_{t_f}\eta_{\pi_o}^{**}(x_f,t_f|x)\\ 
    &\chi_{\pi_o}^{**}(x_f|x)=\chi_{\pi_o}^{+}(x_f|x)+\chi_{\pi}^{-}(x_f|x),\\
    &\eta_{\pi_o}^{**}(x_f,t_f|x)=\eta_{\pi_o}^{+}(x_f,t_f|x)+\eta_{\pi_o}^{-}(x_f,t_f|x),\\ 
    &\sum_{x_f}\sum_{t_f}\eta_{\pi_o}^{**}(x_f,t_f|x) =1,\\
    &\sum_{x_f}\chi_{\pi_o}^{**}(x_f|x) = 1,\\
    &\eta_{\mpol}(x_{\mpol},t_\mpol|x)=\sum_{x_{f_1}}\sum_{\mathclap{t_{f_1}}}\eta_{o_2}(x_{\mpol},t_{\mpol}-t_{f_1}|x_{f_1})\eta_{o_1}(x_{f_1},t_{f_1}|x),\\
    &\chi_{\mpol}(x_{\mpol}|x)=\sum_{x_{f_1}}\chi_{o_2}(x_{\mpol}|x_{f_1})\chi_{o_1}(x_{f_1}|x),\\
    &\bar{\kappa}_{\bos}(\bos,t_f) = \prod_k \bar{\kappa}_{s_k}(s_k,t_f),\\
    &\kappa_{\bos}(\bos,t_f) = 1-\bar{\kappa}_{\bos}(\bos,t_f),\\
    &\xi(t_f|s) = \sum_{s_f}\eta(s_f,t_f|s),\\
    &\xi_{\bos}(t_f|\bos) = \kappa_{\bos}(\bos,t_f)-\kappa_{\bos}(\bos,t_f-1),\\
    &\xi_{\bos}(t_f|\bos) = \bar\kappa_{\bos}(\bos,t_f-1)-\bar\kappa_{\bos}(\bos,t_f),\\
    &\rho_{s}^{\ell}(s_f|s_i,t_f) = P_{\alpha_{\ell}}^{t_f}[i,f],\\
    &\rho_{\pi}^{\ell}(x_f|x_i,t_f) = P_{\pi}^{t_f}[i,f],\\
    &\eta_{\pi}(s_f,t_f|s) = \rho_{s}(s_f|s,t_f)\xi_{\pi}(t_f|s),\\
    &\eta_{\pi,\ell}(\bz_f,x_f,t_f|\bz,x) = \xi_{\bos,\pi}^{\ell}(t_f|\bz,x)\rho_{\pi}^{\ell}(x_f|x,t_f)\prod_{k}\rho_{k}^{\ell}(z_{k,f}|z_k,t_f),\\
    &\kappa_{sub,\bs}^*(\bs)\leq \kappa^*(\bs,z,x).
\end{align}

\newpage
\section{Feasibility Iteration}
From the previous subsection, we can see that feasibility iteration for the OKBEs is propagating the (success/failure) typed absorption probabilities of a policy's absorbing Markov chain (which are the termination probabilities of an option), and we are optimizing the policy for its absorbing Markov chain's statistics (maximizing cumulative feasibility and minimizing expected time).

The feasibility iteration algorithm is as follows: 
\begin{algorithm2e}[h]
\DontPrintSemicolon
\SetKw{return}{return}
\SetKwRepeat{Do}{do}{while}
\SetKwData{conflict}{conflict}
\SetKwData{safe}{safe}
\SetKwData{sat}{sat}
\SetKwData{unsafe}{unsafe}
\SetKwData{unknown}{unknown}
\SetKwData{true}{true}
\SetKwInOut{Input}{input}
\SetKwInOut{Output}{output}
\SetKwFor{Loop}{Loop}{}{}
\SetKw{KwNot}{not}
\begin{small}
	\Input{$(P_x,f_\dg,f_c)$: Dynamics $P_x$, goal function $f_{\dg}$, constraint function $f_c$}
	\Output{$(\kappa,\pi,\eta)$: Cumulative feasibility function $\kappa$, Policy $\pi$, State-time option kernel $\eta$}
    define $f_1 = f_{\dg}f_c,\quad f_2 = (1-f_{\dg})f_{c}$\;
    define $nx$ as the number of states in $P_x$\;
    define $nt$ as an arbitrary time-horizon (this can be extended dynamically if exceeded, not shown below)\;
    initialize $\eta_{\dg}^+,\eta^{-}_{\dg} \gets zeros([nx,nx*const.])$~~\# \text{Const. is a pre-allocated size for time, $\eta$ may need to be adaptively expanded if it's too small.}\; 
    $\kappa(x) \gets zeros(nx)~~\# \text{CFF Initialization}$\;
    $\pi(x) \gets zeros(nx),~~\# \text{Policy Initialization}$\;
    $\eta_{\dg}^+(t_f,x_j|x_i) \gets zeros(nx,nx,nt),~~\# \text{STFF initialization}$\;
    $\eta^{-}_{\dg}(t_f,x_j|x_i) \gets zeros(nx,nx,nt),~~\#
    \text{STIF initialization}$\;
    initialize $\nu \gets zeros(nx)$ as an expected time-to-go function for simplifying Eq. \ref{eq:time-min-pol} by avoiding wasteful computations.\;
	\While{$\eta_{rel}\neq \eta_{old}$}{
        $\eta_{old} \gets \texttt{copy}(\eta_{rel})$\;
    	\For{$x_i \in \mc X$}{
    	$(max\_\kappa,\mc A^*_{x_i}) \gets \argmax_{a \in \mc A}\left[f_1(x_i,a) + f_2(x_i,a)\expec_{x'\sim P_x(\cdot|x_i,\pi(x_i))}\kappa(x')\right]$\;
    	$\kappa(x_i) \gets max\_\kappa$\;
        $\pi(x_i) \gets a^{**}_{x_i}\gets \argmin_{a \in \mc A^*_{x_i}}\left[f_2(x_i,a)\expec_{x'\sim P_x(\cdot|x_i,\pi(x_i))}\nu(x')\right]$\;
        $\nu(x_i) \gets 1 + f_2(x_i,\pi(x_i))\expec_{x'\sim P_{x}(\cdot|x_i,\pi(x_i))}\nu(x')$\;
        $\eta_{\dg}^+(t_f = 0,x_j|x_i) \gets f_1(x_i,a^{**}_{x_i})\delta_{ij}, ~~ \forall x_i \in \mc X$\;
        $\eta_{\dg}^+(t_f = [1:\texttt{end}],:|x_i) \gets f_2(x,a^{**}_{x})\expec_{x'\sim P_x(\cdot|x_i,\pi(x_i))}\eta_{\dg}^+([0:\texttt{end}-1],:|x')$\;
        $\eta_{\dg}^-(t_f = 0,x_j|x_i) \gets \mathbbm{1}_\kappa(x_i)(1-f_c(x_i,a_{x}^{\pi})) \delta_{ij}+\bar{\mathbbm{1}}_\kappa(x_i)\delta_{ij}, ~~ \forall x_i \in \mc X$\;
        $\eta_{\dg}^-(t_f = [1:\texttt{end}],:|x_i) \gets f_2(x_i,a^{**}_{x})\expec_{x'\sim P_x(\cdot|x,\pi(x_i))}\eta_{\dg}^-([0:\texttt{end}-1],:|x')$\;
        
    	}
	}
    $\eta \gets \texttt{Combine}(\eta_{\dg}^+, \eta^{-}_{\dg})$\;
    \texttt{return~}$\kappa$,$\pi$,$\eta$\;
\end{small}
\caption{Stationary Feasibility Iteration}
\label{alg:stationary_feasibility_iteration}
\end{algorithm2e}

\newpage
\section{Option Sequence Breadth First Search}\label{appx:forward_sampling}

\begin{algorithm2e}
\DontPrintSemicolon
\SetKw{return}{return}
\SetKwRepeat{Do}{do}{while}
\SetKwData{conflict}{conflict}
\SetKwData{safe}{safe}
\SetKwData{sat}{sat}
\SetKwData{unsafe}{unsafe}
\SetKwData{unknown}{unknown}
\SetKwData{true}{true}
\SetKwInOut{Input}{input}
\SetKwInOut{Output}{output}
\SetKwFor{Loop}{Loop}{}{}
\SetKw{KwNot}{not}
\begin{small}
	\Input{$\widehat{\eta}_{c}, \mc H_{c} = \{\eta_{e_1,g_1},\eta_{e_1,g_2},...\},\mc P_{z} = \{\pfun_w,...,\pfun_z\},\mc P_{\bz}=\{P_w,...,P_z\},P_x, P_{\bs}$, $\Pi_{e,g}=\{\pi_{e_1,g_1},\pi_{e_1,g_2},...,\pi_{e_\ell,g_{k}}\}$, $\mc O = \{o_{e_1,g_1},o_{e_1,g_2},...,o_{e_\ell,g_{k}} \}$, max\_horizon $M$, $\bar{f}_{1}$}
	\Output{tree}
    let $\mathbf{st}_{init} \gets (\bs_{init},\bz_{init}, x_{init}, e_{init}, t_0)$\;
    let $\kappa_{init} \gets \bar{f}_{1}(\mathbf{st}_{init})$\;
	let $root\leftarrow Node(\mathbf{st}_{init}, \mpol=(), \kappa_{\mpol}=\kappa_{init}, leaf\leftarrow False, parent \leftarrow none)$ be an initial node\;
    define $\pfun_z(\bz_f|\bz,t_d) := \prod_{\pfun_k \in \mc P}\pfun_k(\cdot|\cdot,t_d)$\;
	$tree \leftarrow initialize\_tree(root)$\;
	$Queue.push(root)$\;
	\While{$not\_empty(Queue)$}{
	    $node \leftarrow Queue.pop()$\;
        $({\bs},\bz,x,t) \gets node.{\mathbf{st}}$\;
        $e\leftarrow \zeta(\bz,{\bs})$\;
	    \For{$\pi_{e,g}$ in $\Pi_e\subseteq \Pi$}{ \If{$(\kappa_{\pi_{e,g}}(x,t)> 0)$}
	    {
	            $(\ba,x',t') \leftarrow \widehat{\eta}_c(\ba,x',t_f|x,\pi)$\;
	            $\bz' \leftarrow \pfun_z(\cdot|\bz,t_f)$\;
	            $x'' \leftarrow P_x(x''|x',\pi(x'))$\;
	            $\bz'' \leftarrow \texttt{one\_step\_internal\_update}(\bz',\ba,\mc P_{\bz})$\;
	            ${\bs}''\leftarrow P_{\bs}({\bs}''|{\bs},\ba)$\;
	            $t_f' \leftarrow t_f+1$\;
                $\mathbf{st}'' \gets ({\bs}'',\bz'', x'', t_f')$\;
                $cur\_feas \gets (node.\kappa_{\mpol})\times\bar{f}_{2}(\bs,\bz,x,t)+ \bar{f}_{1}(\bs'',\bz'',x'',t_f') $\;
    	        \uIf{$len(node.\mpol)<M$}{
        	       $new\_node \leftarrow Node(\mathbf{st}'', \kappa_{\mpol}\gets cur\_feas, leaf \leftarrow \texttt{False},parent\leftarrow node)$\;
        	       $tree \leftarrow add\_to\_tree(tree,new\_node)$\;
        	       $Queue.push(new\_node)$\;
    	        }
    	        \Else{
    	            $new\_node \leftarrow Node(\mathbf{s}'', leaf \leftarrow \texttt{True},parent\leftarrow node)$\;
    	           $tree \leftarrow add\_to\_tree(tree,new\_node)$\;
        	        
    	        }
	        }
	    }
	    
	}
        $\mc P_{f\_max} \gets \texttt{get\_feasibility\_maximizing\_plans}(tree.leaves)$\;
        $\mc P_{f\_max\_t\_min} \gets \texttt{get\_time\_minimizing\_plans}(\mc P_{f\_max})$\;
        $\texttt{Return}(\mc P_{f\_max\_t\_min})$\;
\end{small}
\caption{Breadth$\_$First$\_$Plan$\_$Search}
\label{alg:BFS}
\end{algorithm2e}


\end{document}